\documentclass[final,3p,times]{elsarticle}

\usepackage{amssymb}
\usepackage{amsthm}
\usepackage{commath}
\usepackage{amsfonts}
\usepackage{graphicx}
\usepackage{epstopdf}
\usepackage{algorithmic}
\usepackage{algorithm}
\usepackage{hyperref}
\usepackage{subfigure}
\newcommand{\bx}{\mathbf{x}}
\newcommand{\mathf}{\mathbf{f}}
\newcommand{\by}{\mathbf{y}}

\newcommand{\txtin}{\text{in}}
\newcommand{\txtout}{\text{out}}
\newcommand{\train}{\text{train}}
\newcommand{\validate}{\text{validate}}
\newcommand{\se}{\text{se}}
\newcommand{\reconx}{\tilde{\mathbf{x}}}
\newcommand{\dindex}[2]{{(#1,#2)}}
\newcommand{\hatK}{\widehat{K}}
\newcommand{\netN}{\mathcal{N}}
\newcommand{\trainD}{\mathcal{D}_{\mathrm{train}}}
\newcommand{\validS}{\mathcal{S}_{\mathrm{validate}}}
\newcommand{\hatt}{\hat{t}}
\newcommand{\validE}{\mathcal{E}_{\mathrm{validate}}}
\DeclareMathOperator*{\argmin}{arg\,min}
\DeclareMathOperator*{\argmax}{arg\,max}
\newtheorem{theorem}{Theorem}
\newtheorem{lemma}[theorem]{Lemma}
\newtheorem{proposition}[theorem]{Proposition}
\newdefinition{definition}{Definition}
\newdefinition{rmk}{Remark}
\usepackage{lineno}

\journal{}

\begin{document}

\begin{frontmatter}



\title{Deep neural network based adaptive learning for switched systems}

\author[1,2,3]{Junjie He}
\ead{hejj1@shanghaitech.edu.cn}

\author[1,2,3]{Zhihang Xu}
\ead{xuzhh@shanghaitech.edu.cn}

\author[1]{Qifeng Liao\corref{cor1}}
\ead{liaoqf@shanghaitech.edu.cn}
\cortext[cor1]{Corresponding author}

\address[1]{{School of Information Science and Technology, ShanghaiTech University}, {Shanghai}, {China}}
\address[2]{{Shanghai Institute of Microsystem and Information Technology, Chinese Academy of Sciences}, {Shanghai}, {China}}
\address[3]{{University of Chinese Academy of Sciences}, {Beijing}, {China}}


\begin{abstract}
In this paper, we present a deep neural network based adaptive learning (DNN-AL) approach for switched systems. Currently, deep neural network based methods are actively developed for learning governing equations in unknown dynamic systems, but their efficiency can degenerate for switching systems, where structural changes exist at discrete time instants. In this new DNN-AL strategy, observed datasets are adaptively decomposed into subsets, such that no structural changes within each subset. During the adaptive procedures, DNNs are hierarchically constructed, and unknown switching time instants are gradually identified. Especially, network parameters at previous iteration steps are reused to initialize networks for the later iteration steps, which gives efficient training procedures for the DNNs. For the DNNs obtained through our DNN-AL, bounds of the prediction error are established. Numerical studies are conducted to demonstrate the efficiency of DNN-AL.  
\end{abstract}




\end{frontmatter}


\section{Introduction}
\label{sec:intro}
Many problems arising in computational science and engineering are described by dynamical systems with high complexity. 
In many practical applications of these systems, measurement data typically can be collected, but mechanisms or governing equations may not be given a priori, which causes difficulties in understanding the underlying physical phenomena and predicting future dynamics. How to discover or learn the governing equations then becomes a crucial problem. For this purpose, there has been a rapid development in system identification methods during the past few decades. Early methods to learn unknown dynamical systems include the eigen-system realization algorithm  \cite{juang1985eigensystem} and the observer/Kalman filter identification  \cite{juang1993identification}. To seek exact expressions of governing equations, sparse identification of nonlinear dynamics is developed in \cite{brunton2016discovering}, that for noisy and corrupted data is studied in \cite{schaeffer2017sparse,zhang2018robust,tran2017exact}, and extracting dynamics with limited data is studied in \cite{schaeffer2018extracting}. 
In addition, symbolic regression for learning nonlinear dynamical systems is proposed in \cite{schmidt2009distilling}, and equation-free modeling is developed in \cite{kevrekidis2003equation}. Recently, variants of deep neural networks (DNNs) have been interpreted and analyzed with dynamical systems  \cite{chen2018neural,lu2018beyond,weinan2017proposal}. For instance, residual networks (ResNets) can be considered as dynamical systems with forward Euler discretization \cite{he2016deep,qin2019data}. Furthermore, ResNet based methods are proposed to approximate the governing equations for autonomous systems \cite{qin2019data}, non-autonomous systems \cite{qin2021data} and reduced systems \cite{fu2020learning}. 
More specifically, the ResNet block used in the work \cite{qin2019data} is considered as a one-step exact integrator without temporal error, and the proposed networks consisting of multiple ResNet blocks are applied to approximate the evolution operator without requiring time derivatives, so as to avoid additional numerical errors. A general framework for recovering missing dynamical systems is presented in \cite{shixiaojiang21}.

In this work, we focus on dynamical systems with structural changes at discrete time instants, which are referred to as switched systems. 
Switched systems have wide applications, which include epidemiology \cite{keeling2001seasonally}, legged locomotion \cite{holmes2006dynamics}, cascading failures on the electrical grid \cite{dobson2007complex}, and the design of cyber-physical systems \cite{sanfelice2016analysis}. 
For example, in the process of robotic manipulation, interactions with the environment naturally give switching time instants  \cite{van2000introduction,cortes2008discontinuous,liberzon2003switching}. Recently, a hybrid-sparse identification approach has been proposed to identify hybrid systems in \cite{mangan2019model}. In switched systems, discontinuities exist in the governing equations, such that constructing their global approximations becomes challenging. In addition, the number of switching time instants and their locations are not given a priori, which causes significant difficulties in learning the underlying systems.  

To efficiently learn switched systems, we propose a deep neural network based adaptive learning (DNN-AL) approach. In our approach, neural network models to approximate underlying governing equations are hierarchically constructed, and the deep ResNets developed in \cite{qin2019data} are applied to construct local models. 
As the switching time instants are not given, we gradually decompose global DNN models into local models, so that the switching time instants can be identified. That is, for a given adaptivity iteration step, the overall time interval are decomposed into small local intervals, and the observed dataset is decomposed associated with them. Then, with the decomposed datasets, local DNNs are trained, and their validation errors are computed. For the next iteration step, the local interval with the largest validation error is decomposed. Especially, local DNNs for the newly decomposed intervals are initialized with the previous network parameters, which results in an efficient initialization strategy. The adaptive procedure stops when the maximum validation error is small enough. To summarize, the main contributions of this work are three-fold: first, we propose a DNN-AL approach to learn governing equations in switched systems with unknown switching time instants; second, our hierarchical initialization approach results in efficient training procedures for local DNNs; third, for the models constructed by our DNN-AL, our analysis gives the bound of the prediction errors. 

The rest of the paper is organized as follows. In Section \ref{sec:setup}, we present the general problem setup. Section \ref{sec:learn-autonomous-dnn} reviews ResNet based learning methods for  autonomous systems. In Section \ref{sec:method}, our main DNN-AL algorithm is presented, and its error analysis is conducted. Numerical studies are discussed in Section \ref{sec:experiments}. Finally, Section \ref{sec:conclusions} concludes the paper.

\section{Problem setup and preliminary}
\label{sec:setup}

Let $I = \{1,\ldots,K\}$ denote a finite index set, and for each $k\in I$, $\mathf^{(k)}(\cdot):\mathbb{R}^d\rightarrow \mathbb{R}^d$ denotes a function, which is referred to as the governing equation. 
Given a time interval $(0,T_{\max}]$, 
a switching signal function is denoted by $\sigma(t):(0,T_{\max}] \to I$. The switching signal function 
is typically assumed to be a piecewise constant function with a finite number of discontinuities, which are referred to as switching times \cite{liberzon2003switching}. In this paper, the switching signal function is defined as \begin{equation}
\label{eq:switch-signal}
    \sigma(t) = \begin{cases}
   1,\quad T_0:=0< t\leq T_1,\\
   2,\quad T_1 < t \leq T_2,\\
   \cdots\\
   K,\quad T_{K-1} < t\leq T_K:=T_{\max},
\end{cases}
\end{equation}
where $\{T_k\}_{k=1}^{K-1}$ are the $K-1$ discontinuities (or switching times). 
The dynamical system with time-dependent switching \eqref{eq:switch-signal} considered in this work is written as 
\begin{equation}
     \label{eq:swithed-system}
     \frac{\mathrm{d}}{\mathrm{d}t} \bx(t) = \mathf^{(\sigma(t))}(\bx(t)),\quad \bx(0) = \bx_0,
\end{equation} where $\bx(t)\in \mathbb{R}^d$
is a state vector of the system state variables, $\bx_0$ is an initial state and $\sigma(t)$ specifies the active governing function for $t\in (0,T_{\max}]$.

This work focuses on the situation where the switched system \eqref{eq:swithed-system} is unknown, i.e., the governing functions, the discontinuity locations, 
and the number of discontinuity points are not given a priori. Our aim is to construct its numerical model using observed measurement data. 
The observed data are assumed to be available as a collection of several trajectories, where the number of collected trajectories is denoted by $N$, and the corresponding initial states are denoted by $\{\bx_0^{(n)}\}^N_{n=1}$. The time interval is discretized with $J$ uniform grids, the time lag between any two adjacent grid points is denoted by $\Delta:=T_{\max}/(J-1)$, and each grid point is denoted by $t_{j}:=(j-1)\Delta$ for $j=1,\ldots,J$. 
For an initial state $\bx_0^{(n)}$, the state vector at time $t_{j}$ is written as  $\bx(t_{j};\bx_0^{(n)})$, where $n=1,\ldots,N$ and $j=1,\ldots,J$. 
For each $n=1,\ldots,N$, the observation of the trajectory is denoted by
\begin{equation}
    \label{eq:each-trajectory}
    \mathbf{Y}^{(n)} = \{ \by_j^{(n)} := \bx(t_{j};\bx_0^{(n)}) + \epsilon^{(n)}_j \}_{j=1}^{J}\,,
\end{equation}
where the measurement noises $\epsilon^{(n)}_j$ are assumed to be independent and identically distributed random variables. 
We reorganize the overall observed trajectories $\{\mathbf{Y}^{(n)}\}_{n=1}^N$ as 
\begin{equation}
    \begin{aligned}
        \mathcal{D}:=&\{
        (\by_j^{(n)}, \by_{j+1}^{(n)}): j =1,\ldots,J-1,\quad  n=1,\ldots,N\}.
    \end{aligned}
    \label{eq:data-pairs}
\end{equation}

The goal of this work is to develop a numerical model to capture the inherent evolution of the system \eqref{eq:swithed-system} with the limited observed measurement data \eqref{eq:each-trajectory}. 
Precisely speaking, 
we train local models with the data \eqref{eq:each-trajectory}, and for an arbitrary initial state $\bx_0$, we use the models to make predictions $\widehat{\by}$ to approximate the true solutions $\bx$ 
as follows 
\[\widehat{\by}(t_j;\bx_0)\approx \bx(t_j;\bx_0),\quad j=1,\ldots,J.\]

\section{Review of learning autonomous systems via deep Residual networks}
\label{sec:learn-autonomous-dnn}  
In this section, we review neural network based methods in learning autonomous systems following the presentation in work \cite{qin2019data}. 
Consider an autonomous system 
\begin{equation}
\label{eq:autonomous-system}
    \frac{\mathrm{d} \bx(t)}{\mathrm{d} t} = \mathf(\bx(t)),\quad \bx(0) = \bx_0,
\end{equation} where $\mathf(\cdot):\mathbb{R}^d\rightarrow \mathbb{R}^d$ is a Lipschitz continuous governing equation and $\bx_0\in\mathbb{R}^d$ is an initial state. 
Given the state $\bx(t_0)$ at $t_0 \geq 0$ and the  governing equation $\mathf$, for $t>0$, the flow of the system \eqref{eq:autonomous-system} is represented by  
\[\Phi_t(\bx(t_0)) := \bx(t_0 + t)=\bx(t_0)+\int_0^t \mathf(\bx(s+t_0)) \mathrm{d} s.\] 
The exact $\Delta$-lag flow map of the autonomous system \eqref{eq:autonomous-system} is defined as  
\begin{equation}
\begin{aligned}
\Phi_{\Delta}(\mathbf{x}(t_0))&:=
      \mathbf{x}(t_{0}+\Delta)  \\ 
      & =\mathbf{x}(t_0) + \int_0^\Delta \mathbf{f}(\mathbf{x}(s+t_0))\mathrm{d}s\\
      &= \mathbf{x}(t_0) +  \Delta\cdot \mathbf{f}(\mathbf{x}(t_0+\gamma))\\
      &= \bx(t_0) + \Delta\cdot \mathbf{f}(\Phi_\gamma (\bx(t_0))),
      \quad \gamma \in [0,\Delta]\,,
  \end{aligned}
  \label{eq:flow_map}
\end{equation}
and $\phi_{\Delta}(\bx(t_0);\mathf) := \Delta\cdot \mathbf{f}(\Phi_\gamma (\bx(t_0)))$ is the \textit{effective increment}.
When the governing equation $\mathf$ 
and the time lag $\Delta$ are fixed, 
the effective increment $\phi_{\Delta}$ 
is uniquely determined by 
the input state $\bx(t_0)$. 
\par 
In the following, we generically denote the input and the output of the 
$\Delta$-lag flow map \eqref{eq:flow_map} as $\by_\txtin$ and $\by_\txtout$ (i.e., let $\by_\txtin=\mathbf{x}(t_0)$ and $\by_\txtout=\mathbf{x}(t_{0}+\Delta)$ for \eqref{eq:flow_map}). 
An efficient residual network (ResNet) is proposed to model the
$\Delta$-lag flow map in \cite{qin2019data} (see \cite{he2016deep} for the original ResNet). 
A feedforward neural network composing $L\in \mathbb{N}^+$ fully connected layers is defined as 
\[
\mathcal{F}(\bx;\Theta):=\mathbf{w}_L\cdots h(\mathbf{w}_2(h(\mathbf{w}_1\bx+\mathbf{b}_1))+\mathbf{b}_2) +\mathbf{b}_L,
\] where $h$ is the non-linear activation function, $\mathbf{w}_l$ and $\mathbf{b}_l$ are the weight and the bias in the $l$-th layer respectively, 
and $\Theta:=\left\{\mathbf{w}_l,\mathbf{b}_l \right\}_{l=1}^{L}$ is the associated parameter set of the network. 
Denoting the identity operator as $\mathcal{I}:\mathbb{R}^d \to\mathbb{R}^d$, 
a ResNet block is defined as
\begin{equation}
\mathcal{R}(\by_\txtin;\Theta):=\by_\txtin + \mathcal{F}(\by_\txtin; \Theta), \label{eq:res}
\end{equation}
which can also be rewritten as $\mathcal{R}(\by_\txtin;\Theta):=( \mathcal{I}+ \mathcal{F}(\cdot; \Theta))(\by_\txtin)$.  
The ResNet block \eqref{eq:res} can be considered as a one-step ``exact" integrator for solving autonomous systems \cite{qin2019data}. 
Here, we choose the following element-wise hyperbolic tangent (tanh) as the activation function, that is
\[
\text{tanh}(x) = \frac{\mathrm{e}^{2x}-1}{\mathrm{e}^{2x}+1}
.\]  
\par
Following \cite{qin2019data}, a ResNet block is stacked $M\in \mathbb{N}^+$ times to construct a deep neural network. 
Based on the structure of the recurrent neural network (see  \cite[ch.~10]{goodfellow2016deep}),
a deep ResNet is defined as 
\begin{equation}
    \label{eq:RT-ResNet}
    \left\{ 
    \begin{aligned}
    \widehat{\by}_0 &= \by_\txtin,\\
    \widehat{\by}_{m+1} &= \mathcal{R}(\widehat{\by}_m; \Theta),\quad m =0,\ldots,M-1,\\
    \widehat{\by}_\txtout &= \widehat{\by}_{M},
    \end{aligned}
    \right.  
\end{equation}
where the output gives an approximation of  
the output of the $\Delta$-lag flow map, i.e., $\widehat{\by}_\txtout\approx \by_\txtout$. 
For the convenience of notation, the deep ResNet is denoted by $\mathcal{N}$, i.e., $\widehat{\by}_{\txtout}=\netN(\by_{\txtin};\Theta)$.

\par
To train the deep ResNet, i.e., to find optimal values of  $\Theta$ in \eqref{eq:RT-ResNet}, the following square error loss function is considered  
\begin{equation}
   \begin{aligned}
   \ell_{\se}(\by_{\txtin},\by_{\txtout};\Theta) &:=
   \Big\|\netN(\by_{\text{in}};\Theta) - \by_{\text{out}}
   \Big\|^2,
   \end{aligned}
   \label{eq:square-loss}
\end{equation} 
where $\|\cdot \|$ denotes the standard Euclidean norm.  
Given a training dataset 
$\trainD= \{(\by^{(s)}_\txtin, \by^{(s)}_\txtout),s=1,\ldots,N_{\train}\}$ with size $N_{\train}$,
 the training loss function of the deep neural network $\mathcal{N}$ is 
\begin{equation}
    \begin{aligned}
        \mathcal{L}_{\train}(\Theta) &= \frac{1}{N_{\train}} \sum_{(\by_\txtin,\by_\txtout)\in \trainD}
        \ell_{\se}(\by_\txtin,\by_\txtout;\Theta). 
    \end{aligned}
    \label{eq:train-loss}
\end{equation}
Based on \eqref{eq:train-loss}, we choose the optimal parameters $\Theta^*$
\begin{equation}
        \Theta^* = \argmin_{\Theta} (\mathcal{L}_{\train}(\Theta)).
    \label{eq:train-REsNet-autonomous}
\end{equation} 
As the above optimization problem is highly non-convex and the amount of data is large, 
the stochastic gradient descent (SGD) method \cite{bottou2018optimization} is applied, which is a common choice for training deep neural networks with backpropagation \cite{lecun1989backpropagation,krizhevsky2012imagenet,he2016deep}. 
The training procedure is summarized as follows. 
First,  
the training dataset $\mathcal{D}_{\train}$ is  divided into $N_{\text{batch}}$ mini-batches.
Each mini-batch is denoted by $M_{r}$ with size  $|M_{r}|$ for $r=1,\ldots,N_{\text{batch}}$, and the sizes of different mini-batches are similar. 
Then for each mini-batch $M_{r}$, the expectation of the square error loss \eqref{eq:square-loss} and the corresponding gradient are estimated, and the parameters are updated as
\[
\Theta=\Theta - \tau \nabla_{\Theta}\frac{1}{|M_{r}|}\left(\sum_{(\by_\txtin,\by_\txtout)\in M_{r}} \Big\|\mathcal{N}(\by_{\text{in}};\Theta) - \by_{\text{out}}\Big\|^2  \right),
\]
where $\tau$ is a given learning rate. 
For each epoch of the optimization procedure, 
the above updating step is repeated $N_{\text{batch}}$ times. 
At the end of each epoch, a pre-set learning rate scheduler is applied to decrease the learning rate $\tau$, and the mini-batch data are randomly reshuffled. The training procedure is summarized in Algorithm \ref{alg:trining-ResNet}, where $N_E$ is the number of epochs. 

\begin{algorithm}[!ht]
    \caption{Training a deep ResNet}
    \label{alg:trining-ResNet}
    \begin{algorithmic}[1]
    \REQUIRE{Initial ResNet 
    $\mathcal{N}(\cdot;\Theta_0)$,
    and training dataset $\trainD=\left\{(\by_{\txtin}^{(s)},\by_\txtout^{(s)})\right\}_{s=1}^{N_{\train}}$.}
    \STATE{Set an initial learning rate $\tau$, and initialize a learning rate scheduler.}
    \STATE{Set $\Theta=\Theta_0$.}
    \FOR{training epoch $=1,\ldots, N_E$}
    \FOR{$r =1,\ldots, N_{\text{batch}}$}
    \STATE{Obtain the mini batch $M_{r}$ containing $|M_{r}|$ data pairs.}
    \STATE{Update the parameters $\Theta$ by the stochastic gradient descent: $\Theta=\Theta - \tau \nabla_{\Theta}\frac{1}{|M_{r}|}\left (\sum_{(\by_\txtin,\by_\txtout)\in M_{r}} \Big\|\mathcal{N}(\by_{\text{in}};\Theta) - \by_{\text{out}}\Big\|^2  \right)  $.}
    \ENDFOR
    \STATE{Decrease the learning rate $\tau$ using the learning rate scheduler.}
    \ENDFOR
    \STATE{Let $\Theta_*=\Theta$.}
    \ENSURE{Trained RestNet $\mathcal{N}(\cdot;\Theta_*)$.}
    \end{algorithmic}
\end{algorithm}

\section{Method and analysis}
\label{sec:method}
Our goal is to construct efficient neural network approximations for the $\Delta$-lag flow map associated with the switched system \eqref{eq:swithed-system}, which is written as
\begin{equation}
    \label{eq:flow-swithced}
    \begin{aligned}
    \Phi_{\Delta}^{\mathrm{switched}}(\bx(t_0))&:=
        \bx(t_0+\Delta)\\
        &=\bx(t_0)+
        \begin{cases}
             \int_0^\Delta \mathf^{(1)}(\bx(t_0+s))\dif s,\quad 0=T_0<t_0\le t_0+\Delta \le T_1\\
             \cdots\\
             \int_0^\Delta \mathf^{(i)}(\bx(t_0+s))\dif s,\quad T_{i-1}<t_0\le t_0+\Delta \le T_i\\
             \cdots\\
             \int_0^\Delta\mathf^{(K)}(\bx(t_0+s))\dif s,\quad T_{K-1}<t_0\le t_0+\Delta \le T_K,\\
        \end{cases}
    \end{aligned}
\end{equation}
where $T_1,\ldots,T_K$ are the switching time instants introduced in \eqref{eq:switch-signal}, $\mathf^{(\cdot)}(\cdot)$ refers to the governing equation associated with the switching signal (see \eqref{eq:swithed-system}), and $\phi_{\Delta}^{(i)}(\bx(t_0)):=\int_0^\Delta \mathf^{(i)}(\bx(t_0+s))\dif s$ denotes the effective increment corresponding to  $\mathf^{(i)}$.  
For this purpose, we propose a deep neural network based adaptive learning (DNN-AL) approach. In DNN-AL, training datasets of the $\Delta$-lag flow map are adaptively decomposed, and local neural network approximations are constructed using the decomposed datasets, so that the switching time instants can be identified, and the local networks can approximate the $\Delta$-lag flow map well. 
During the adaptive procedure, the networks at the current iteration step are reused in the next iteration step, which results in efficient initial guesses for training deep ResNets. Details of our DNN-AL approach are presented in Section \ref{subsec:DNN-AL}, and error bounds of the predicated states are analyzed in Section \ref{subsec:theo}.

\subsection{Adaptive learning for switched systems}
\label{subsec:DNN-AL}
To begin with, the adaptivity iteration step is denoted by $\hatK$, and the time interval $(0,T_{\max}]$ is partitioned into $\hatK$ sub-intervals at the step $\hatK$ (details are given in the following). 
The double index $\dindex{\hatK}{i}$ is used to denote the $i$-th sub-interval at adaptivity iteration step $\hatK$. 
Endpoints of the sub-intervals are denoted by $\{\hatt^{(\hatK,i)}\}_{i=0}^{\hatK}$, where $0=\hatt^\dindex{\hatK}{0} <\cdots<\hatt^\dindex{\hatK}{\hatK}=T_{\max}$. 
Here, the validation data is the last trajectory of the $N$ collected orbits \eqref{eq:each-trajectory} at the time between 0 and $T_{\max}$, $\mathbf{Y}^{(N)}$, and let $\by_\validate^{(j)}=\by_{j}^{(N)},$ for $j=1,\ldots,J$, where $J$ is the number of discrete states at each trajectory. 
Details of DNN-AL to approximate switched systems are presented as follows.

For the first step, i.e.\ $\hatK=1$,  
let $\hatt^{(1,0)}=0$, $\hatt^\dindex{1}{1}=T_{\max}$, and partition the dataset \eqref{eq:data-pairs} into a training dataset $\trainD^\dindex{1}{1}$  and a validation dataset $\validS^\dindex{1}{1}$, which are defined as 
\begin{equation}
\label{eq:initial-dataset}
\begin{aligned}
\trainD^\dindex{1}{1}&:=\left\{(\by_j^{(n)}, \by_{j+1}^{(n)}):\quad n=1,\ldots,N-1,\quad j=1,\ldots, J-1\right\}, \\
\validS^\dindex{1}{1}&:=\left\{(\by_{\validate}^{(j)},\by_{\validate}^{(j+1)}):\quad j=1,\ldots,J-1\right\}.
\end{aligned}
\end{equation}
A deep neural network (see \eqref{eq:RT-ResNet}) with (untrained) parameters is initialized,
and it is denoted by $\netN^\dindex{1}{1}_0(\cdot;\Theta^\dindex{1}{1}_0)$, where $\Theta^\dindex{1}{1}_0$ are the parameters. We then train a deep ResNet 
using Algorithm \ref{alg:trining-ResNet} with the initial network $\netN^\dindex{1}{1}_0$ and the training dataset $\trainD^\dindex{1}{1}$,
and denote the trained deep neural network by  $\netN^\dindex{1}{1}(\cdot; \Theta^\dindex{1}{1})$.  
To assess the accuracy of the trained network,
the following validation error is considered,
\[
\validE^\dindex{1}{1} = \frac{1}{\left|\validS^\dindex{1}{1}\right|}\sum_{(\by_{\txtin},\by_{\txtout})\in\validS^\dindex{1}{1}}\ell_{\se}(\by_{\txtin},\by_{\txtout};\Theta^{(1,1)})\,,
\] 
where $\ell_{\se}$ is the square error function defined in \eqref{eq:square-loss}.
If the $\validE^\dindex{1}{1}$ is larger than a given tolerance $tol$, the interval $(\hatt^\dindex{1}{0},\hatt^\dindex{1}{1}]$ is equally divided into two parts, i.e., $(\hatt^\dindex{1}{0},\hatt^\dindex{1}{1}]=(\hatt^\dindex{1}{0},t^{(1)}]\cup (t^{(1)},\hatt^\dindex{1}{1}]$, where 
$t^{(1)}= (\hatt^\dindex{1}{0}+\hatt^\dindex{1}{1})/2$,
and the datasets are divided and associated with these two intervals. After that, the adaptivity iteration step is updated, i.e.\ $\hatK=2$, and the new deep neural networks are trained based on the divided datasets. 

In general, after $\hatK$ adaptivity iteration steps, the time domain $(0,T_{\max}]$ is partitioned into $\hatK$ intervals  
$(\hatt^\dindex{\hatK}{i-1},\hatt^\dindex{\hatK}{i}]$ for $i=1,\ldots,\hatK$, where 
$0=\hatt^\dindex{\hatK}{0}<\cdots<\hatt^\dindex{\hatK}{\hatK}=T_{\max}$, and 
the trained neural networks $\netN^\dindex{\hatK}{i}$, the associated training sets $\trainD^\dindex{\hatK}{i}$, the validation sets $\validS^\dindex{\hatK}{i}$ and  the validation errors $\validE^\dindex{\hatK}{i}$ are obtained.
The procedures for the next iteration step 
are as follows.
First, the index of the local deep neural network
with the maximum validation error is identified as, 
\begin{eqnarray} \label{eq:max_error}
i^{*}=\argmax_{i=1,\ldots,\hatK} \validE^\dindex{\hatK}{i} \,.
\end{eqnarray}
When the maximum validation error is not unique, 
$i^*$ in \eqref{eq:max_error} is set to an arbitrary 
index with the maximum value. 
Next, the interval associated with the maximum error $(\hatt^\dindex{\hatK}{i^*-1},\,\hatt^\dindex{\hatK}{i^*}]$ is equally divided into two intervals. Denoting 
    $t^{(\hatK)}= (\hatt^\dindex{\hatK}{i^*-1} + \hatt^\dindex{\hatK}{i^*})/2$, 
the time domain are partitioned by $0=\hatt^\dindex{\hatK+1}{0}<\cdots<\hatt^\dindex{\hatK+1}{\hatK+1}=T_{\max}$ with
\begin{equation}
\label{eq:update-time-instants}
\hatt^\dindex{\hatK+1}{i}=
\begin{cases} 
\hatt^\dindex{\hatK}{i},\quad &0\leq i< i^*\\
t^{(\hatK)},\quad &i=i^*\\
\hatt^\dindex{\hatK}{i-1},\quad & \hatK+1\geq i>i^*.
\end{cases}
\end{equation}

From the iteration step $\hatK$
to the step $\hatK+1$, the time intervals, 
datasets, local deep neural networks and validation errors are the same, except those associated with the 
index $i^*$. 
So, for $i=1,\ldots,i^*-1$, we define 
\begin{equation}
    \label{eq:left-data}
    \begin{aligned}
    &\trainD^\dindex{\hatK+1}{i}:= \trainD^\dindex{\hatK}{i}, \,\validS^\dindex{\hatK+1}{i}:=\validS^\dindex{\hatK}{i},\\
    &\netN^\dindex{\hatK+1}{i}(\cdot;\Theta^\dindex{\hatK+1}{i}):= \netN^\dindex{\hatK}{i}(\cdot;\Theta^\dindex{\hatK}{i}),\\  
    &\validE^\dindex{\hatK+1}{i}:=\validE^\dindex{\hatK}{i},
    \end{aligned}
\end{equation}
and for $i=i^*+2,\ldots,\hatK+1$,
\begin{equation}
    \label{eq:right-data}
    \begin{aligned}
    &\trainD^\dindex{\hatK+1}{i} :=\trainD^{(\hatK,i-1)},\, \validS^{(\hatK+1,i)}:=\validS^{(\hatK,i-1)},\\
    &\mathcal{N}^{(\hatK+1,i)}(\cdot;\Theta^{(\hatK+1,i)}):=\mathcal{N}^{(\hatK,i-1)}(\cdot;\Theta^{(\hatK,i-1)}), \\ 
    &\validE^\dindex{\hatK+1}{i}:=\validE^\dindex{\hatK}{i-1}.
    \end{aligned}
\end{equation}

Next, new local deep ResNets are constructed for $(\hatt^\dindex{\hatK+1}{i^*-1},\hatt^\dindex{\hatK+1}{i^*}]$ and $(\hatt^\dindex{\hatK+1}{i^*},\hatt^\dindex{\hatK+1}{i^*+1}]$. For the interval $(\hatt^\dindex{\hatK+1}{i^*-1},\hatt^\dindex{\hatK+1}{i^*}]$, a training dataset $\trainD^\dindex{\hatK+1}{i^*}$ and a validation dataset $\validS^\dindex{\hatK+1}{i^*}$ are 
constructed as 
\begin{equation}
    \label{eq:sub-dataset}
    \begin{aligned}
    \trainD^\dindex{\hatK+1}{i^*}&:=\left\{(\by^{(n)}_j,\by^{(n)}_{j+1}): n=1,\ldots,N-1,\, t_j\in (\hatt^\dindex{\hatK+1}{i^*-1}, \hatt^\dindex{\hatK+1}{i^*}]\right\},\\
    \validS^\dindex{\hatK+1}{i^*}&:=\left\{(\by_{\validate}^{(j)},\by_{\validate}^{(j+1)}):t_j\in (\hatt^\dindex{\hatK+1}{i^*-1}, \hatt^\dindex{\hatK+1}{i^*}]\right\}.
    \end{aligned}
\end{equation}
Initial deep ResNets for these two intervals are set to the trained neural network $\netN^\dindex{\hatK}{i^*}$,
i.e., $\netN_0^\dindex{\hatK+1}{i^*}:=\netN^\dindex{\hatK}{i^*}$ and
    $\netN_0^\dindex{\hatK+1}{i^*+1}:=\netN^\dindex{\hatK}{i^*}$. 
Then the trained neural network $\netN^\dindex{\hatK+1}{i^*}$ is obtained using Algorithm \ref{alg:trining-ResNet} with inputs $\netN^\dindex{\hatK+1}{i^*}_0$ and  $\trainD^\dindex{\hatK+1}{i^*}$. 
The validation error for $\netN^\dindex{\hatK+1}{i^*}$ is computed as 
\begin{equation}
\begin{aligned}
    \validE^\dindex{\hatK+1}{i^*} = \frac{1}{\left |\validS^\dindex{\hatK+1}{i^*}\right |}
    \sum_{(\by_{\txtin},\by_{\txtout})\in \validS^\dindex{\hatK+1}{i^*}} \ell_{\se}(\by_{\txtin},\by_{\txtout};\Theta^\dindex{\hatK+1}{i^*})\,,
    \end{aligned}
    \label{eq:valid-error}
\end{equation}
where $\ell_{\se}$ is defined in \eqref{eq:square-loss}. 
Similarly, for the interval $(\hatt^\dindex{\hatK+1}{i^*},\, \hatt^\dindex{\hatK+1}{i^*+1}]$, the datasets $\trainD^\dindex{\hatK+1}{i^*+1}$ and  $\validS^\dindex{\hatK+1}{i^*+1}$ are constructed, the neural network $\netN^\dindex{\hatK+1}{i^*+1}$ is trained, and the validation error $\validE^\dindex{\hatK+1}{i^*+1}$ is computed. After that, 
the iteration step is updated $\hatK=\hatK+1$, and the above procedures are repeated until the maximum validation error is smaller than a given tolerance. 

\begin{figure}
    \centering
    \subfigure[Training.]{
    \label{fig:train-phase}
     \includegraphics[scale=0.42]{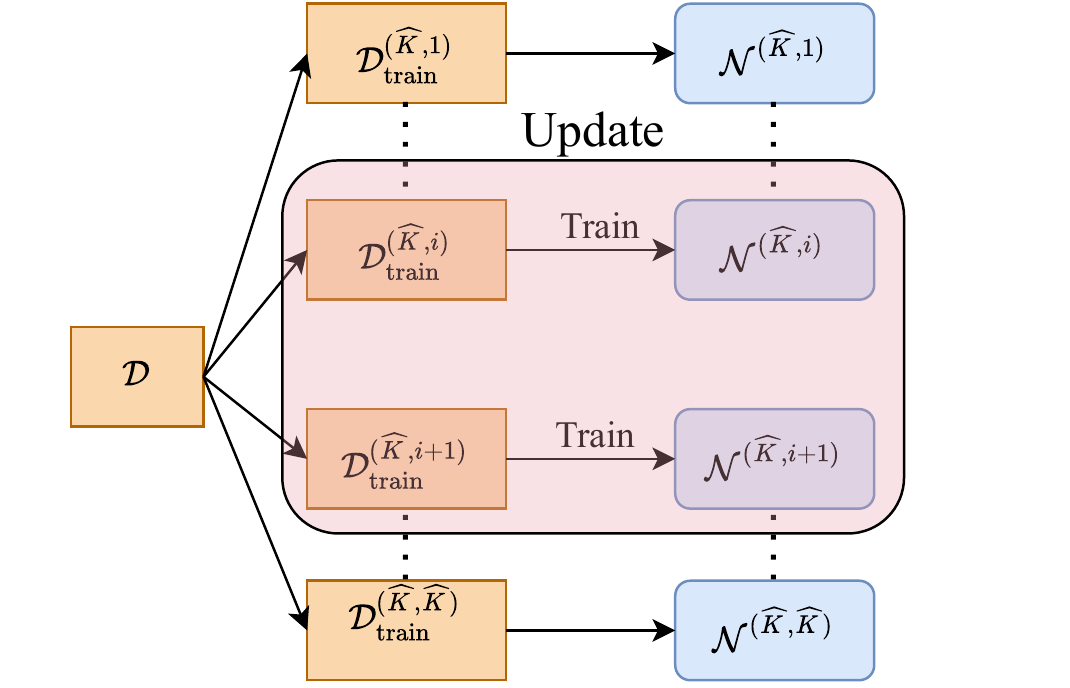}
    }
    \subfigure[Validation.]{\label{fig:validation-phase}\includegraphics[scale=0.38]{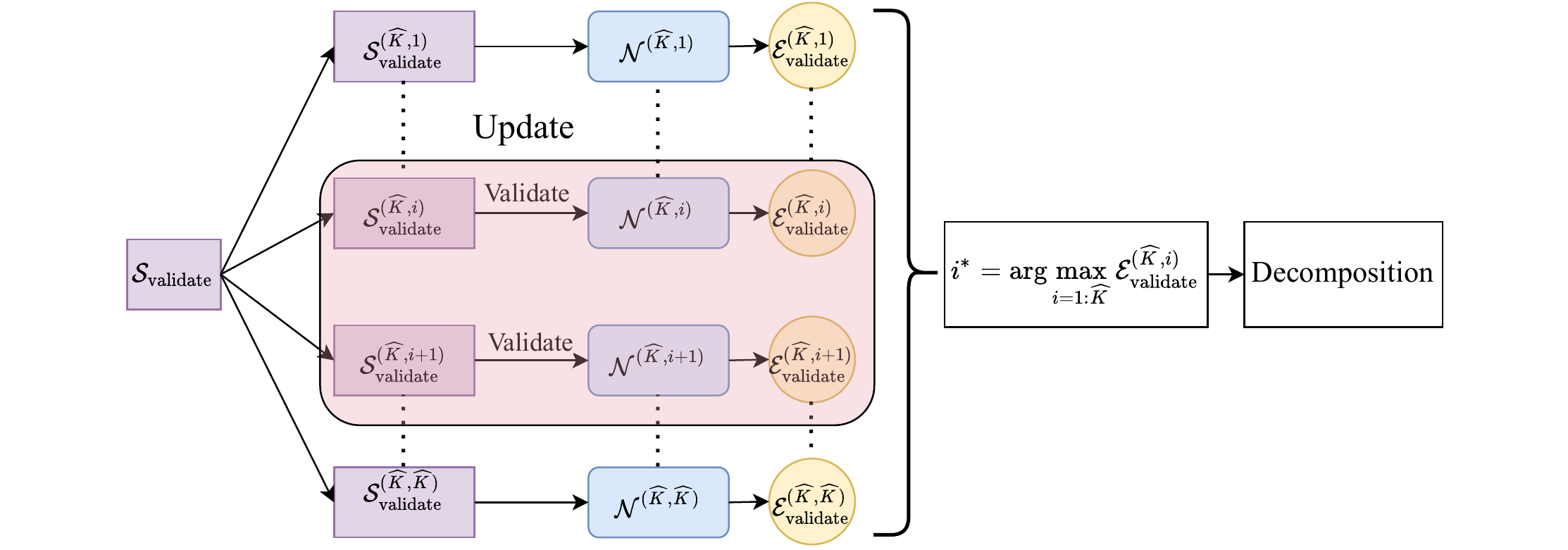}}
    \subfigure[Network architecture.]{\label{fig:ResNet-used}\includegraphics[scale=0.9]{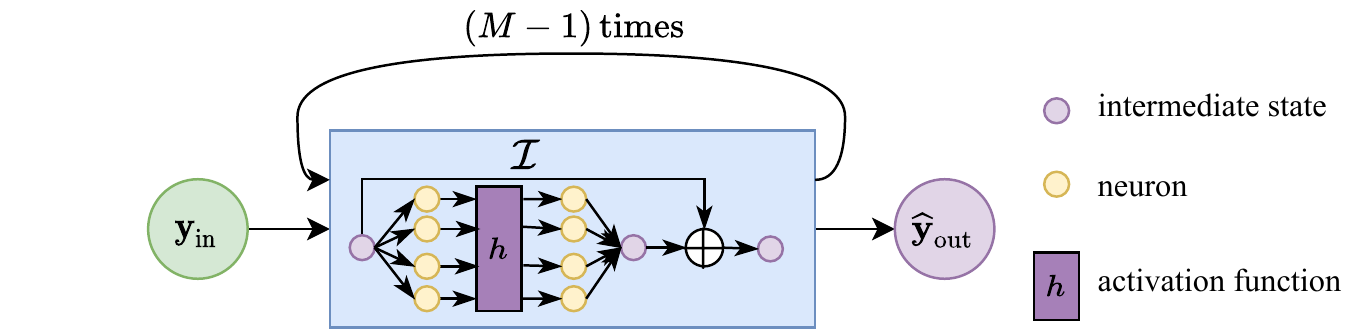}}
    \caption{DNN-AL for switched systems:  (a) 
    Two new training datasets and the  corresponding neural networks are obtained at each adaptivity iteration; 
    (b) The validation errors  are computed, and the interval with the largest error  is identified and decomposed; 
    (c) Neural networks with recurrent structures.
    }
    \label{fig:DNN-AL-train-valid}
\end{figure}

Our DNN-AL approach is summarized in Algorithm \ref{alg:main-alg}, where $tol$ is the given tolerance. 
The major ingredients at each iteration step of DNN-AL include: (1) training deep neural networks to learn flow maps using datasets associated with time intervals, 
(2) decomposing the interval with the largest validation error. 
\figurename{\ref{fig:DNN-AL-train-valid}} illustrates the simple schematic of DNN-AL. \figurename{\ref{fig:train-phase}} and \figurename{\ref{fig:validation-phase}} show the flow charts of the training procedure and the validation procedure respectively.  The network architecture used in the algorithm is illustrated in \figurename{\ref{fig:ResNet-used}}. 

\begin{algorithm}[!htp]
\caption{Adaptive learning for unknown switched systems}
\label{alg:main-alg}

\begin{algorithmic}[1]
    \REQUIRE{Dataset $\mathcal{D}=\left\{
        (\by_j^{(n)}, \by_{j+1}^{(n)}): j =1,\ldots,J-1,\quad  n=1,\ldots,N\right\}$.}
    \STATE{Initialize $\hatK=1$.}
    \STATE{Partition $\mathcal{D}$ into the training dataset $\trainD^\dindex{1}{1}$ and the validation dataset $\validS^\dindex{1}{1}$ (see \eqref{eq:initial-dataset}).} 
    \STATE{Initialize a deep ResNet  $\mathcal{N}_0^\dindex{1}{1}$  (see network structure in  \eqref{eq:RT-ResNet}).}
    \STATE{Obtain the trained deep ResNet $\mathcal{N}^\dindex{1}{1}$ using Algorithm \ref{alg:trining-ResNet} with
    the initial network $\mathcal{N}_0^\dindex{1}{1}$ and the data set  $\trainD^\dindex{1}{1}$.
    }
    \STATE{Compute the validation error $\validE^\dindex{1}{1}$ for $\mathcal{N}^\dindex{1}{1}$ with $\validS^\dindex{1}{1}$ (see \eqref{eq:valid-error}).}
    \WHILE{$\max\left( \{\validE^\dindex{\hatK}{i}  \}_{i=1}^{\hatK}\right)\geq tol$}
    \STATE{Find the index of the maximum error: $i^*= \argmax_{i=1:\hatK} \validE^\dindex{\hatK}{i}$.}
    \STATE{Determine the time instant $t^{(\hatK)}:= \left(\hatt^\dindex{\hatK}{i^*-1}+\hatt^\dindex{\hatK}{i^*}\right)/2$.}
    \STATE{Divide the interval  $\left(\hatt^\dindex{\hatK}{i^*-1},\hatt^\dindex{\hatK}{i^*}\right]$ into $\left(\hatt^\dindex{\hatK}{i^*-1},t^{(\hatK)}\right]$ and $\left(t^{(\hatK)},\hatt^\dindex{\hatK}{i^*}\right]$.}
    \STATE{Update time instants (see \eqref{eq:update-time-instants}).}
    \STATE{Update datasets, neural networks and others for unchanged sub-intervals (see \eqref{eq:left-data} and \eqref{eq:right-data}).}
\STATE{Obtain  the training and the validation sets $\trainD^\dindex{\hatK+1}{i^*}$,  $\trainD^\dindex{\hatK+1}{i^*+1}$, $\validS^\dindex{\hatK+1}{i^*}$ and $\validS^\dindex{\hatK+1}{i^*+1}$ (see \eqref{eq:sub-dataset}).}
    \STATE{Initialize $\netN_0^\dindex{\hatK+1}{i^*}:=\netN^\dindex{\hatK}{i^*}$ and
    $\netN_0^\dindex{\hatK+1}{i^*+1}:=\netN^\dindex{\hatK}{i^*}$}.
    \STATE{Obtain  $\netN^\dindex{\hatK+1}{i^*}$ using Algorithm \ref{alg:trining-ResNet} with $\netN_0^\dindex{\hatK+1}{i^*}$ and $\trainD^\dindex{\hatK+1}{i^*}$.}
    \STATE{Obtain  $\mathcal{N}^\dindex{\hatK+1}{i^*+1}$ using Algorithm \ref{alg:trining-ResNet} with $\netN_0^\dindex{\hatK+1}{i^*+1}$ and $\trainD^\dindex{\hatK+1}{i^*+1}$.}
    \STATE{Compute the validation errors $\validE^\dindex{\hatK+1}{i^*},\,\validE^\dindex{\hatK+1}{i^*+1}$ for $\netN^\dindex{\hatK+1}{i^*}$ and $\netN^\dindex{\hatK+1}{i^*+1}$ with  $\validS^\dindex{\hatK+1}{i^*}$ and $\validS^\dindex{\hatK+1}{i^*+1}$ (see \eqref{eq:valid-error}).}
    \STATE{Let $\hatK=\hatK+1$.}
    \ENDWHILE
\ENSURE{Trained DNN models $\netN^\dindex{\hatK}{i}$ for $i=1,\ldots,\hatK$, identified time points $\left\{t^{(i)}\right\}_{i=1}^{\hatK-1}$ and the endpoints of sub-intervals $\left\{\hatt^\dindex{\hatK}{i}\right\}_{i=0}^{\hatK}$.}
\end{algorithmic}
\end{algorithm}

\subsection{Theoretical properties}
\label{subsec:theo}
In this section, we present a theoretical analysis of our proposed algorithm. 
The analysis provides an error bound between the solution of switched systems \eqref{eq:swithed-system} and the predictions of neural networks given by Algorithm \ref{alg:main-alg}. Our analysis processes through the following three steps. First, an auxiliary system is introduced, and a bound of the error between the solution of the auxiliary system and the original system \eqref{eq:swithed-system} is given. Second, a bound of the error between solution states of the auxiliary system and predictions by neural networks is provided. 
Finally, based on results obtained in the first and the second steps, the bound of the error between the solution of the switched system \eqref{eq:swithed-system} and the predictions of neural networks in Algorithm \ref{alg:main-alg} is given. 

To simplify the presentation in this section, we assume that the switched system has only one switching time instant, i.e., 
\begin{equation}
\label{eq:2seg-signal}
    \sigma(t)=
\begin{cases}
1,\, t\in (T_0=0,T_1]\\
2,\, t\in(T_1,T_2=T_{\max}].
\end{cases}
\end{equation}
We note that the following analysis can be extended to switched systems with multiple switching time instants. 
Based on the time instants obtained in Algorithm \ref{alg:main-alg}, an auxiliary system is defined as follows. 

\begin{definition}[Auxiliary system]
\label{def:auxiliary-system}
Given the time instants $\{\hat{t}^{(\widehat{K},i)}\}_{i=0}^{\widehat{K}}$ obtained using Algorithm \ref{alg:main-alg}, denoting 
\[
    \breve{t}=\underset{t\in \{\hat{t}^{(\widehat{K},i)}\}_{i=0}^{\widehat{K}}}{\argmin} |t - T_1|,
\]
an auxiliary system is defined as 
\begin{equation}
    \label{eq:auxiliary-system}
    \frac{\mathrm{d}\reconx(t)}{\mathrm{d}t}=\mathf^{(\tilde{\sigma}(t))}(\reconx(t)),\quad \reconx(0)=\bx_0,
\end{equation}
where $\mathf^{(1)}$ and $\mathf^{(2)}:\mathbb{R}^d\rightarrow \mathbb{R}^d$ are introduced in \eqref{eq:swithed-system},
$\bx_0$ is an initial state 
and $\tilde{\sigma}(t)$ is the signal function defined as 
\[\tilde{\sigma}(t)=\begin{cases}
1,\quad t\in (0, \breve{t}]\\
2,\quad t\in (\breve{t}, T_2].
\end{cases}\]
\end{definition}

Here, we give a bound for the error between the solution $\bx(t)$ of the original system \eqref{eq:swithed-system} and the solution $\reconx(t)$ of the auxiliary system \eqref{eq:auxiliary-system}. 
\begin{proposition}
\label{proposition:reconstruct-system}
Assume that the functions $\mathf^{(k)}(\bx)$ for $k=1,2$ in \eqref{eq:swithed-system} and \eqref{eq:auxiliary-system} are Lipschitz continuous 
with Lipschitz constants $L_k$ for $\bx\in\mathbb{R}^d$,
and there exist constants $\mu,\eta >0$ such that $\max_{\bx\in \mathbb{R}^d}\|\mathf^{(1)}(\bx)-\mathf^{(2)}(\bx)\|\leq \mu$ and $|\breve{t}-T_1|\leq \eta$, where $\breve{t}$ is introduced in  Definition \ref{def:auxiliary-system}. Without loss of generality, we assume that $\breve{t}\geq T_1$. The difference between the solution $\bx(t)$ of the original switched system 
\eqref{eq:swithed-system}
and the solution $\reconx(t)$ of the auxiliary system \eqref{eq:auxiliary-system}
is bounded as  
\[\|\bx(t)-\reconx(t)\|\leq \begin{cases}
0,\quad &t\in (0, T_1]\\
\mu(t-T_1)\exp(\max(L_1,L_2)(t-T_1)),\quad &t\in(T_1, \breve{t}]\\
\mu\eta\exp(L_2(t-\breve{t})+\max(L_1,L_2)\eta),\quad &t\in(\breve{t}, T_2].
\end{cases}\]
\end{proposition}
\begin{proof}
First, for any $t\in (0,T_1]$, given the initial condition $\reconx(0)=\bx(0)=\bx_0$, 
\[
\begin{aligned}
    \|\bx(t)-\reconx(t)\| &=\left \|\bx(0)-\reconx(0)+\int_0^{t} \mathf^{(1)}(\bx(s))-\mathf^{(1)}(\reconx(s))\dif s \right\| = 0.    
\end{aligned}
\]

Second, for any $t\in (T_1, \breve{t}]$, 
\[
\begin{aligned}
    \|\bx(t)-\reconx(t)&\|=\left\|\bx(T_1)-\reconx(T_1) + \int_{T_1}^{t} \mathf^{(\sigma(s))}(\bx(s)) - \mathf^{(\tilde{\sigma}(s))}(\reconx(s))\dif s\right\| \\
    &\leq \|\bx(T_1)-\reconx(T_1)\| + 
    \int_{T_1}^{t} \|\mathf^{(\sigma(s))}(\bx(s)) - \mathf^{(\tilde{\sigma}(s))}(\reconx(s))\|\dif s\\
    &\leq 0 +  \int_{T_1}^{t} \|\mathf^{(\sigma(s))}(\bx(s)) - \mathf^{(\tilde{\sigma}(s))}(\bx(s))\|\dif s\\
    &\quad +\int_{T_1}^{t} \|\mathf^{(\tilde{\sigma}(s))}(\bx(s)) - \mathf^{(\tilde{\sigma}(s))}(\reconx(s))\|\dif s \\
    &\leq \mu(t-T_1) +  \int_{T_1}^{t} L_{\tilde{\sigma}(s)}\|\bx(s) - \reconx(s)\|\dif s \\
    &\leq \mu(t-T_1) +  \int_{T_1}^{t} \max(L_1,L_2)\|\bx(s) - \reconx(s)\|\dif s.
\end{aligned}
\]
By using Gronwall-Bellman inequality, $\|\bx(t)-\reconx(t)\|\leq \mu(t-T_1)\exp(\max(L_1,L_2)(t-T_1))$.

Lastly, for $t\in ( \breve{t}, T_2]$,
\[
\begin{aligned}
    \|\bx(t)-\reconx(t)\|&=
    \|\bx(\breve{t})-\reconx(\breve{t})+\int_{\breve{t}}^{t} \mathf^{(2)}(\bx(s)) - \mathf^{(2)}(\reconx(s))\dif s\|\\
    &\leq \|\bx(\breve{t})-\reconx(\breve{t})\|+\int_{\breve{t}}^{t} \|\mathf^{(2)}(\bx(s)) - \mathf^{(2)}(\reconx(s))\|\dif s\\
    &\leq \|\bx(\breve{t})-\reconx(\breve{t})\|+\int_{\breve{t}}^{t} L_2\|\bx(s) - \reconx(s)\|\dif s\\
    &\leq \exp(L_2(t-\breve{t}))\|\bx(\breve{t})-\reconx(\breve{t})\|(\text{Using Gronwall-Bellman inequality})\\
    &\leq \mu(\breve{t}-T_1)\exp(L_2(t-\breve{t})) \exp(\max(L_1,L_2)(\breve{t}-T_1))\\
    &\leq \mu\eta\exp(L_2(t-\breve{t})) \exp(\max(L_1,L_2)\eta)\\
    &=\mu\eta\exp(L_2(t-\breve{t})+\max(L_1,L_2)\eta).
\end{aligned}
\]
\end{proof}
Similarly, under the same assumptions of Proposition \ref{proposition:reconstruct-system}, when $\breve{t}<T_1$, 
the differences between the states of the switched system and the auxiliary system are concluded
\[\|\bx(t)-\reconx(t)\|\leq \begin{cases}
0,\quad &t\in (0, \breve{t}]\\
\mu(t-\breve{t})\exp(\max(L_1,L_2)(t-\breve{t})),\quad &t\in(\breve{t}, T_1]\\
\mu\eta\exp(L_2(t-T_1)+\max(L_1,L_2)\eta),\quad &t\in(T_1, T_2].
\end{cases}\]

\par 
For the system involving $\mathf^{(1)}$ and $\mathf^{(2)}$, the flow map of the system can be shown it is locally Lipschitz.
\begin{lemma}
	\label{lemma:flow-map}
For $k=1,2$, the $\Delta$-lag flow map associated with function $\mathf^{(k)}$ is defined as 
\begin{equation}
    \label{eq:flow-map-with-function}
    \Phi_{\Delta}^{(k)}(\bx_0):=\bx_0 + \int_0^\Delta \mathf^{(k)}(\bx(t)) \dif t,
\end{equation}
where $\bx(0)= \bx_0\in\mathbb{R}^d$ is a given initial state. If the assumptions in Proposition \ref{proposition:reconstruct-system} hold, 
for any given initial conditions $\bx,\by\in \mathbb{R}^d$, we have
 \[
 \|\Phi_{\Delta}^{(k)}(\bx) - \Phi_{\Delta}^{(k)}(\by)\|\leq \exp(L_k \Delta)\|\bx-\by\|.
 \]
\end{lemma}
\begin{proof}
The proof directly follows the continuity of the dynamical system with respect to the initial condition \cite[p.~43]{teschl2012ordinary}.
\end{proof}
Since neural networks are employed, the universal approximation property is recalled.
\begin{lemma}(see \cite{pinkus1999approximation})
	\label{lemma:universal-approximation}
	For any compact set $D\subset \mathbb{R}^d$, any continuous function $F: \mathbb{R}^d\rightarrow \mathbb{R}^d$ and any positive real number $\varepsilon$, 
	there exists a single-hidden-layer neural network $\mathcal{N}(\cdot, \Theta)$ with the parameters $\Theta$ such that 
	\[\max_{\by\in D}|F(\by) - \mathcal{N}(\by;\Theta)|\leq \varepsilon,\] if and only if the activation functions are continuous and are not polynomials.
\end{lemma}
\begin{proof}
The proof is given in the \cite{pinkus1999approximation}.
\end{proof}
As in Algorithm \ref{alg:main-alg}, the neural networks to approximate the flow maps \eqref{eq:flow-map-with-function} are denoted by $\{\netN^\dindex{\hatK}{i}\}_{i=1}^{\widehat{K}}$.
Letting $\mathcal{C}_i$ be the convex hull of the training dataset $\trainD^\dindex{\hatK}{i}$, and
denoting 
\[
i' = \argmin_{i\in \{0,\ldots,\widehat{K}\}} |\hatt^\dindex{\hatK}{i} - \breve{t}|,
\]
it is assumed that the learned neural networks $\{\netN^\dindex{\hatK}{i}\}_{i=1}^{\hatK}$ have sufficient accuracy, i.e., for a sufficiently small number $\varepsilon\geq 0$,  
\begin{equation}
\label{eq:neural-network-flow}
	\begin{cases}
	    \max_{\by_\txtin\in \mathcal{C}_i}
	\|\netN^\dindex{\hatK}{i}(\by_\txtin;\Theta^\dindex{\hatK}{i})-\Phi_{\Delta}^{(1)}(\by_\txtin)\| \le \varepsilon\,,\quad i=1,\ldots,i',\\
	\max_{\by_\txtin\in \mathcal{C}_i}
	\|\netN^\dindex{\hatK}{i}(\by_\txtin;\Theta^\dindex{\hatK}{i})-\Phi_{\Delta}^{(2)}(\by_\txtin)\| \le \varepsilon\,,\quad i=i'+1,\ldots,\widehat{K}.\\
	\end{cases}
\end{equation}
We utilize the trained networks to predict the states of the unknown switched system \eqref{eq:swithed-system}. 
For a given initial condition $\widehat{\by}(0)$, the prediction at time $t_{j+1}$ is generated as 
\begin{equation}
    \label{eq:nn-predictions}
    \widehat{\by}(t_{j+1}) = \begin{cases}
    \netN^\dindex{\hatK}{1}(\widehat{\by}(t_j);\Theta^\dindex{\hatK}{1}),\quad &t_j\in (\hat{t}^\dindex{\hatK}{0}=0,\hat{t}^\dindex{\hatK}{1}]\\
    \ldots\\
    \netN^\dindex{\hatK}{i}(\widehat{\by}(t_j);\Theta^\dindex{\hatK}{i}),\quad &t_j\in (\hat{t}^\dindex{\hatK}{i-1},\hat{t}^\dindex{\hatK}{i}]\\
    \ldots \\
    \netN^\dindex{\hatK}{\hatK}(\widehat{\by}(t_j);\Theta^\dindex{\hatK}{\hatK}),\quad &t_j\in (\hat{t}^\dindex{\hatK}{\hatK-1},\hat{t}^\dindex{\hatK}{\hatK}=T_{\max}],
    \end{cases} 
\end{equation}
where $j=1,\ldots,J-1$.

In this section, the initial state for neural networks predictions \eqref{eq:nn-predictions}, the initial condition of the switched system \eqref{eq:swithed-system} and the auxiliary system  \eqref{eq:auxiliary-system} are considered to be the same, i.e., $\widehat{\by}(0)=\bx(0)=\reconx(0)$. 
\begin{proposition}
\label{proposition:neural-network-approximation}
Suppose the assumptions in Lemma \ref{lemma:flow-map} hold.  
For $\widehat{\by}(t_{j-1})\in \mathcal{C}_i$ (where $i=1,\ldots,\widehat{K}$, and $\mathcal{C}_i$ is the convex hull of the training set $\trainD^\dindex{\hatK}{i}$) and $\widehat{\by}(t_j)=\netN^\dindex{\hatK}{i}(\widehat{\by}(t_{j-1});\Theta^\dindex{\hatK}{i})$ with $j=1,\ldots,J$, assume that $\widehat{\by}(t_j)\in \mathcal{C}_i$
and the neural networks are sufficiently accurate such that \eqref{eq:neural-network-flow} holds. 
Then the difference between the prediction using neural networks $\widehat{\by}(t_j)$ and the solution of the auxiliary system $\reconx(t_j)$ is bounded by
	\[\|\widehat{\by}(t_j)-\reconx(t_j)\|\leq \begin{cases}
	\frac{1-\exp(L_1 t_j)}{1-\exp(L_1\Delta)}\varepsilon,\quad &t_j\in(t_1=0,\breve{t}] \\
	\frac{1-\exp(L_2 (t_j-\breve{t}))}{1-\exp(L_2\Delta)}\varepsilon \\
	\quad +\exp(L_2(t_j-\breve{t}))\frac{1-\exp(L_1 \breve{t})}{1-\exp(L_1\Delta)}\varepsilon,
	\quad &t_j\in(\breve{t},T_2] .
	\end{cases}\]
\end{proposition}
\begin{proof}

When $t_j\in(t_1=0,\breve{t}]$ and $\hat{t}^{(\widehat{K},i-1)}<t_j<\hat{t}^{(\widehat{K},i)}$, 
we have
\[
\begin{aligned}
    \|\widehat{\by}(t_j)-\reconx(t_j)\|&=\|\netN^\dindex{\hatK}{i} (\widehat{\by}(t_{j-1});\Theta^\dindex{\hatK}{i})-\Phi_{\Delta}^{(1)}(\reconx(t_{j-1}))\|\\
    &\leq  \|\netN^\dindex{\hatK}{i} (\widehat{\by}(t_{j-1});\Theta^\dindex{\hatK}{i})-\Phi_{\Delta}^{(1)}(\widehat{\by}(t_{j-1}))\|\\
    &\quad +\|\Phi_{\Delta}^{(1)}(\widehat{\by}(t_{j-1}))-\Phi_{\Delta}^{(1)}(\reconx(t_{j-1}))\| \\
    &\leq \varepsilon + \exp(L_1 \Delta)\|\widehat{\by}(t_{j-1})-\reconx(t_{j-1})\| \\
    &\quad \ldots \\
    &\leq \frac{1-\exp(L_1 t_j)}{1-\exp(L_1\Delta)}\varepsilon + \exp(L_1 t_j)\|\widehat{\by}(0)-\reconx(0)\|\\
    &=\frac{1-\exp(L_1 t_j)}{1-\exp(L_1\Delta)}\varepsilon,
\end{aligned}
\]
where the last equation is obtained as $\widehat{\by}(0)=\reconx(0)$.
Similarly, for $t_j\in(\breve{t},T_2]$, the error bound satisfies
\[
\begin{aligned}
    \|\widehat{\by}(t_j)-\reconx(t_j)\|&\leq \varepsilon + \exp(L_2 \Delta)\|\widehat{\by}(t_{j-1})-\reconx(t_{j-1})\| \\
    &\quad \ldots\\
    &\leq \frac{1-\exp(L_2 (t_j-\breve{t}))}{1-\exp(L_2\Delta)}\varepsilon + \exp(L_2(t_j-\breve{t}))\|\widehat{\by}(\breve{t})-\reconx(\breve{t})\|\\
    &\leq \frac{1-\exp(L_2 (t_j-\breve{t}))}{1-\exp(L_2\Delta)}\varepsilon \\
    &\quad + \exp(L_2(t_j-\breve{t}))\frac{1-\exp(L_1 \breve{t})}{1-\exp(L_1\Delta)}\varepsilon.
\end{aligned}
\]
\end{proof}
The bounds in Proposition \ref{proposition:reconstruct-system} and Proposition \ref{proposition:neural-network-approximation} result in
the following error bounds for the neural networks 
obtained using Algorithm \ref{alg:main-alg}. 
\begin{theorem}
\label{thm:nn-error-bound}
Supposing assumptions of Proposition \ref{proposition:reconstruct-system} and Proposition \ref{proposition:neural-network-approximation} hold, the error of the predictions using the neural networks \eqref{eq:nn-predictions} is bouned as
\[
\|\widehat{\by}(t_j) - \bx(t_j)\|\leq
\begin{cases}
    \frac{1-\exp(L_1 t_j)}{1-\exp(L_1\Delta)}\varepsilon, &t_j\in(0,T_1]  \\
\mu(t_j-T_1)\exp(\max(L_1,L_2)(t_j-T_1))\\
\quad +\frac{1-\exp(L_1 t_j)}{1-\exp(L_1\Delta)}\varepsilon, &t_j\in(T_1,\breve{t}]\\
\mu\eta\exp(L_2(t_j-\breve{t})+\max(L_1,L_2)\eta) \\
\quad + \frac{1-\exp(L_2 (t_j-\breve{t}))}{1-\exp(L_2\Delta)}\varepsilon\\
\quad + \exp(L_2(t_j-\breve{t}))\frac{1-\exp(L_1 \breve{t})}{1-\exp(L_1\Delta)}\varepsilon, &t_j\in(\breve{t},T_2],
\end{cases}
\]
where $\bx(t)$ is the accurate solution of the original system \eqref{eq:swithed-system} 
and $t_{j}:=(j-1)\Delta$ with  $\Delta:=T_{\max}/(J-1)$ for $j=1,\ldots,J$.
\end{theorem}
\begin{proof}
Using Proposition \ref{proposition:reconstruct-system}, we have 
\[
\begin{aligned}
   \|\reconx(t_j) - \bx(t_j)\|\leq 
\begin{cases}
0,\quad &t_j\in(0,T_1]\\
\mu(t_j-T_1)\exp(\max(L_1,L_2)(t_j-T_1)),\quad &t_j\in(T_1,\breve{t}]\\
\mu\eta\exp(L_2(t_j-\breve{t})+\max(L_1,L_2)\eta),\quad &t_j\in(\breve{t},T_2].
\end{cases}
\end{aligned}
\]
The bounds in Proposition \ref{proposition:neural-network-approximation} give 
\[
\begin{aligned}
    \|\widehat{\by}(t_j)-\reconx(t_j)\|\leq 
    \begin{cases}
	\frac{1-\exp(L_1 t_j)}{1-\exp(L_1\Delta)}\varepsilon,\quad  &t_j\in(0,T_1]\\
	\frac{1-\exp(L_1 t_j)}{1-\exp(L_1\Delta)}\varepsilon,
	\quad  &t_j\in(T_1,\breve{t}]\\
	\frac{1-\exp(L_2 (t_j-\breve{t}))}{1-\exp(L_2\Delta)}\varepsilon \\
	\quad +\exp(L_2(t_j-\breve{t}))\frac{1-\exp(L_1 \breve{t})}{1-\exp(L_1\Delta)}\varepsilon,
	\quad  &t_j\in(\breve{t},T_2].
	\end{cases}
\end{aligned}
\]
Finally, using the triangle inequality gives
\[
\begin{aligned}
    \|\widehat{\by}(t_j) - \bx(t_j)\| &\leq \|\widehat{\by}(t_j) - \reconx(t_j)\| + \|\reconx(t_j) - \bx(t_j)\|\\
    &\leq 
    \begin{cases}
	\frac{1-\exp(L_1 t_j)}{1-\exp(L_1\Delta)}\varepsilon + 0,\quad  &t_j\in(0,T_1] \\
	\mu(t_j-T_1)\exp(\max(L_1,L_2)(t_j-T_1))
	 \\
	\quad +\frac{1-\exp(L_1 t_j)}{1-\exp(L_1\Delta)}\varepsilon ,
	\quad &t_j\in(T_1,\breve{t}] \\
	\mu\eta\exp(L_2(t_j-\breve{t})+\max(L_1,L_2)\eta)\\ \quad +\frac{1-\exp(L_2 (t_j-\breve{t}))}{1-\exp(L_2\Delta)}\varepsilon\\ 
	\quad +\exp(L_2(t_j-\breve{t}))\frac{1-\exp(L_1 \breve{t})}{1-\exp(L_1\Delta)}\varepsilon 
	,
	\quad &t_j\in(\breve{t},T_2].
	\end{cases}
\end{aligned}
\]
\end{proof}
From Theorem \ref{thm:nn-error-bound},
it is clear that the prediction error of the neural networks obtained from Algorithm \ref{alg:main-alg} is bounded by the error of the neural network approximation for the $\Delta$-lag flow maps \eqref{eq:flow-map-with-function} and the error for the switching point \eqref{eq:switch-signal}.

\section{Numerical experiments}
\label{sec:experiments}
In this section, 
numerical experiments are conducted to illustrate the effectiveness of our DNN-AL approach presented in Algorithm \ref{alg:main-alg}. 
Four test problems are considered---the first and the second ones consider the forced damped oscillator problem with two and three subsystems respectively, the third one considers a forced damped pendulum problem, and the fourth test problem focuses on the heat equation. 
In this work, deep neural networks are implemented based on the open-source machine library Pytorch \cite{paszke2019pytorch}. 
To train deep neural networks (see Algorithm \ref{alg:trining-ResNet}), the size of each mini-batch is set to $100$, and the number of epochs is set to $100$. 
The SGD method used in this paper is AdamW \cite{loshchilov2017decoupled}, and the initial learning rate is set to 0.001. In addition, we implement the cosine annealing part of \cite{loshchilov2016sgdr} to tune the learning rate. To initialize the deep ResNet on line 3 of Algorithm \ref{alg:main-alg}, the Kaiming initialization \cite{he2015delving} is used, and all the biases are initially set to zeros. The tolerance $tol$ in Algorithm \ref{alg:main-alg} is set to $0.005$.  

\subsection{Forced damped oscillator with two subsystems}\label{sec:test1}
We start with the following forced damped oscillator equation 
\begin{equation}
   \left\{
      \begin{array}{ll}
         \frac{\dif }{\dif t}x_1 = x_2\,,\\
         \frac{\dif }{\dif t}x_2 = -k x_1 - \nu x_2 +f\,,
      \end{array}
    \right.
    \label{eq:oscillator}
\end{equation}
where $\nu$ is the damping term from the friction force, $f$ is the external force term, $k$ is the spring constant, and $t\in (0,40]$. 
Following the generic notation in \eqref{eq:switch-signal}--\eqref{eq:swithed-system}, the governing equations for this test problem are set to 
\[
\mathf^{(1)} = 
    \left ( \begin{array}{cc}
     x_2  \\
     -x_1 - 0.1 x_2 + 2
\end{array} \right),\,
    \mathf^{(2)}=
    \left ( \begin{array}{cc}
     x_2  \\
     - x_1 - 0.5 x_2 + 10
\end{array} \right),\]
and the time-dependent signal function is set to  
\[\sigma(t)=\left\{
\begin{array}{ll}
     1\, ,t\in (0,27.6]  \\
      2\,,t\in (27.6,40].
\end{array}\right.\]
The spatial domain considered is $D=[-3,3]^2$. The time lag is set to $\Delta=0.05$ and then $J=801$.  

To generate the observed datasets,  200 initial states are generated using the uniform distribution with the range $D$. Then the switched system is solved using the LSODA solver provided in SciPy \cite{petzold1983automatic,2020SciPy-NMeth} for each initial state, and 200 trajectories are collected. The datasets are constructed through the procedures in Section \ref{sec:setup}. For our DNN-AL approach, the neural network for this test problem consists of 10 ResNet blocks with 2 fully connected layers per block. The 2 fully connected layers in each block have 20 nodes and 2 nodes respectively. 

\begin{figure}[htp!]
   \centering
      \subfigure[Iteration step 1, $x_1(t)$]{
      \includegraphics[scale=0.34]{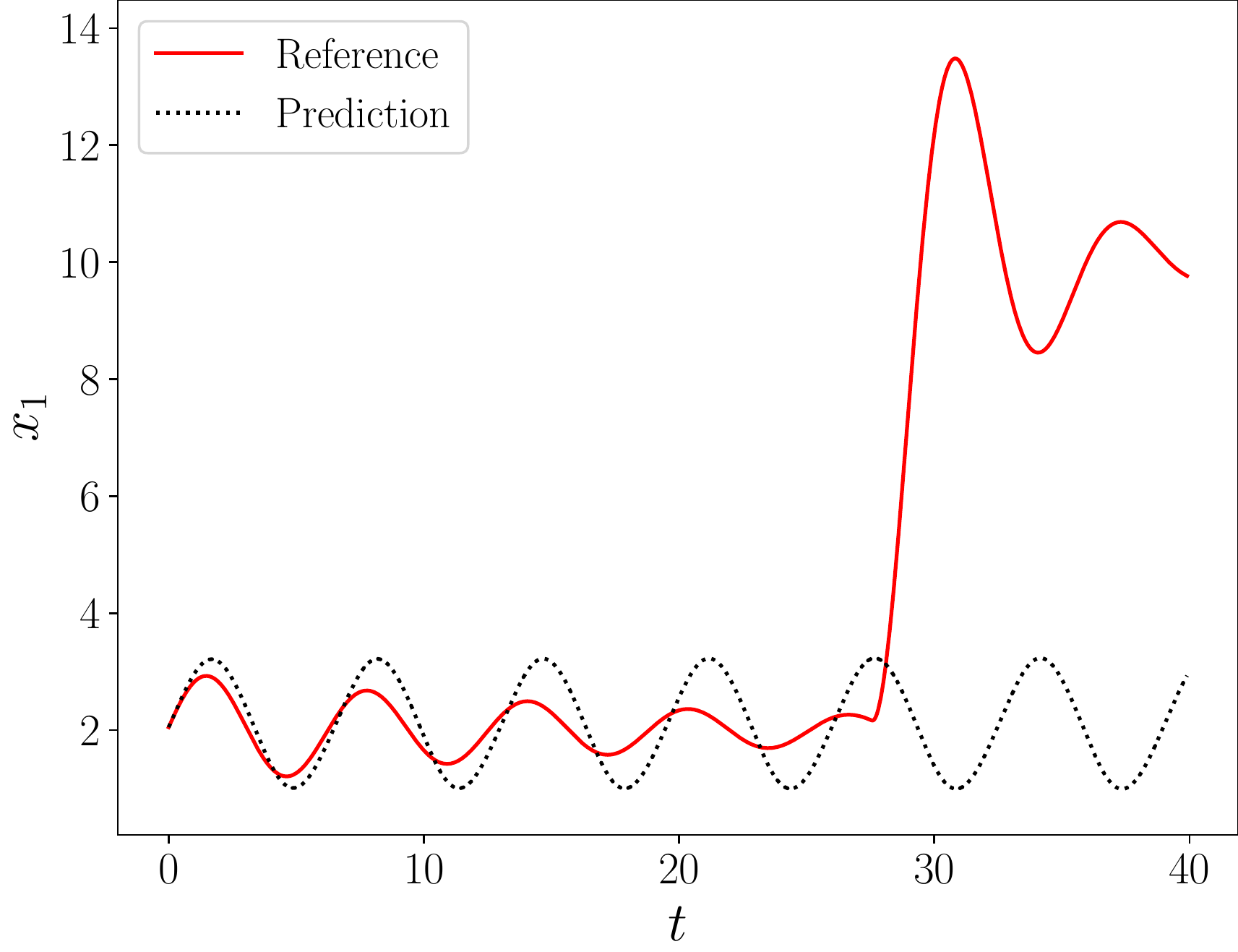}
      \label{fig:osci-x1-k1}
      }
   \subfigure[Iteration step 1, $x_2(t)$]{
      \includegraphics[scale=0.34]{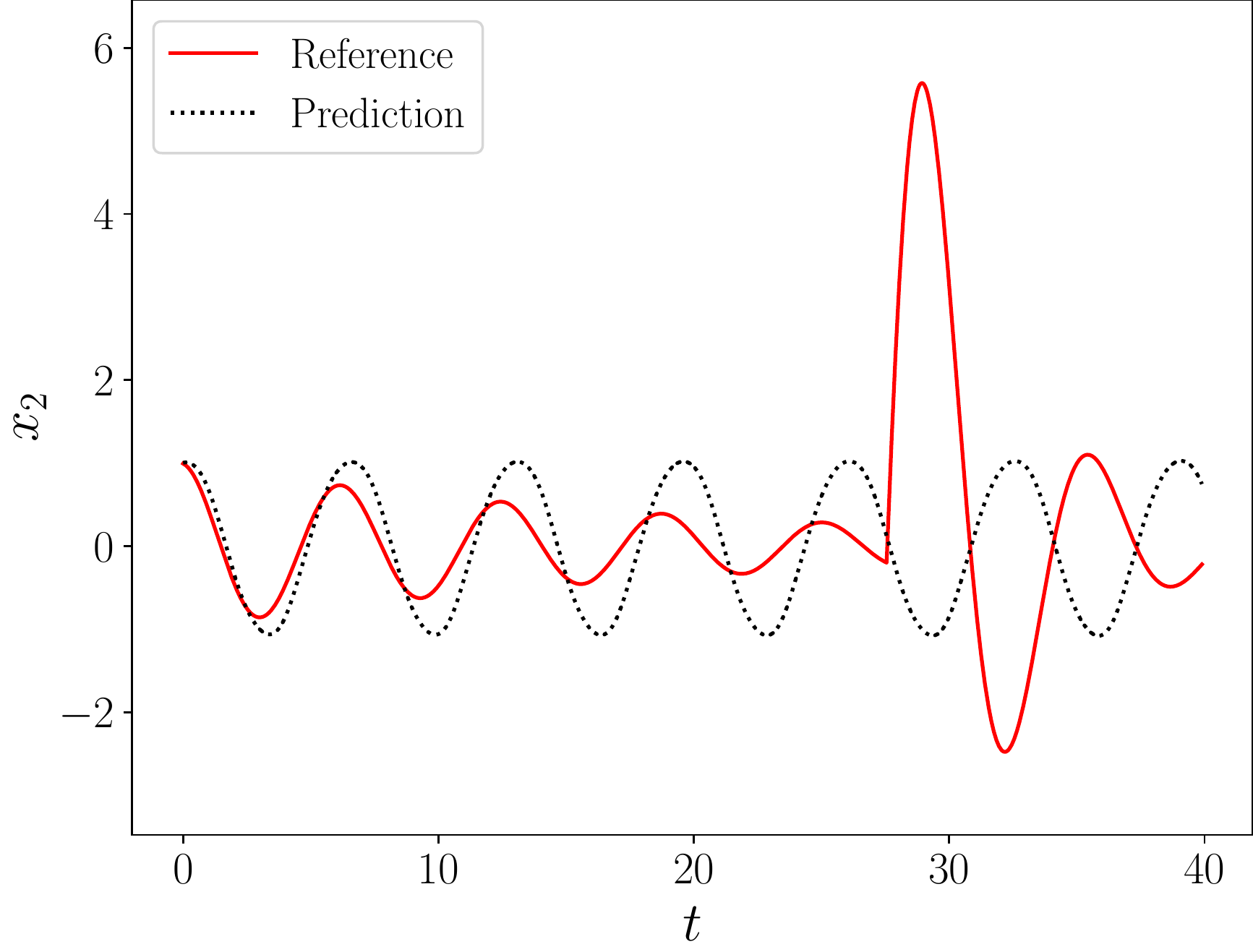}
      \label{fig:osci-x2-k1}
      }
   \subfigure[Iteration step 3, $x_1(t)$]{
      \includegraphics[scale=0.34]{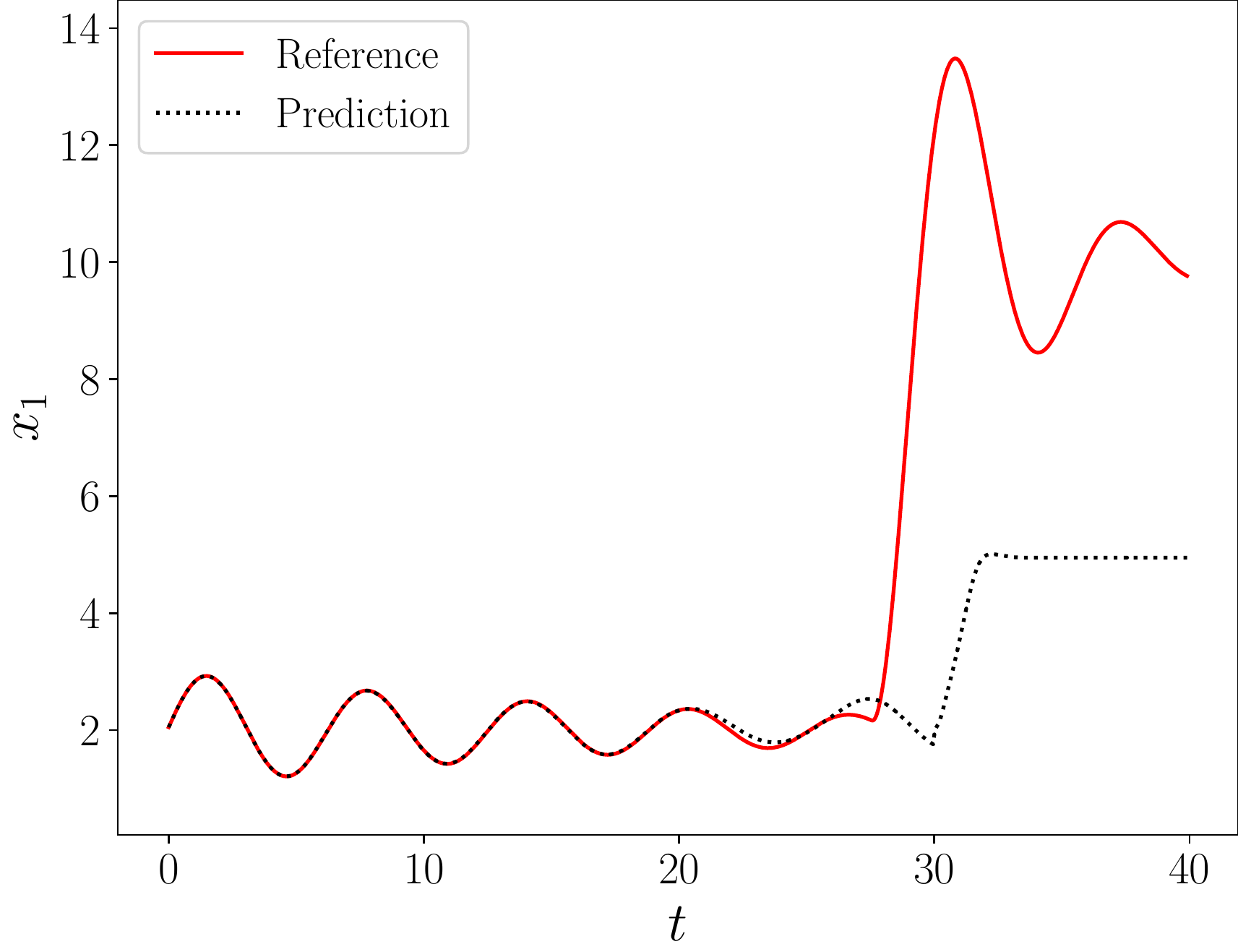}
      \label{fig:osci-x1-k3}
      }
   \subfigure[Iteration step 3, $x_2(t)$]{
      \includegraphics[scale=0.34]{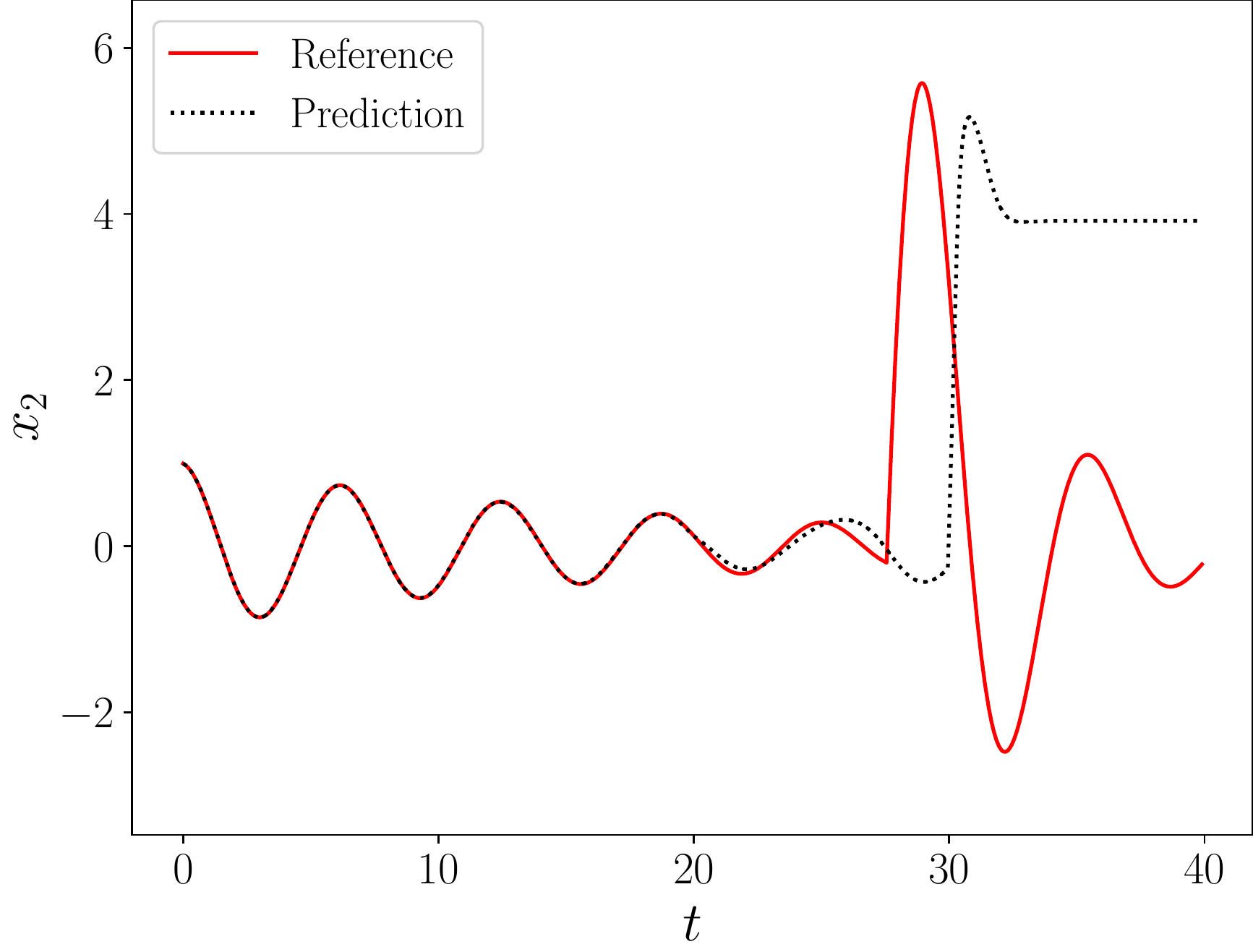}
      \label{fig:osci-x2-k3}
      }   
  \subfigure[Iteration step 7, $x_1(t)$]{
      \includegraphics[scale=0.34]{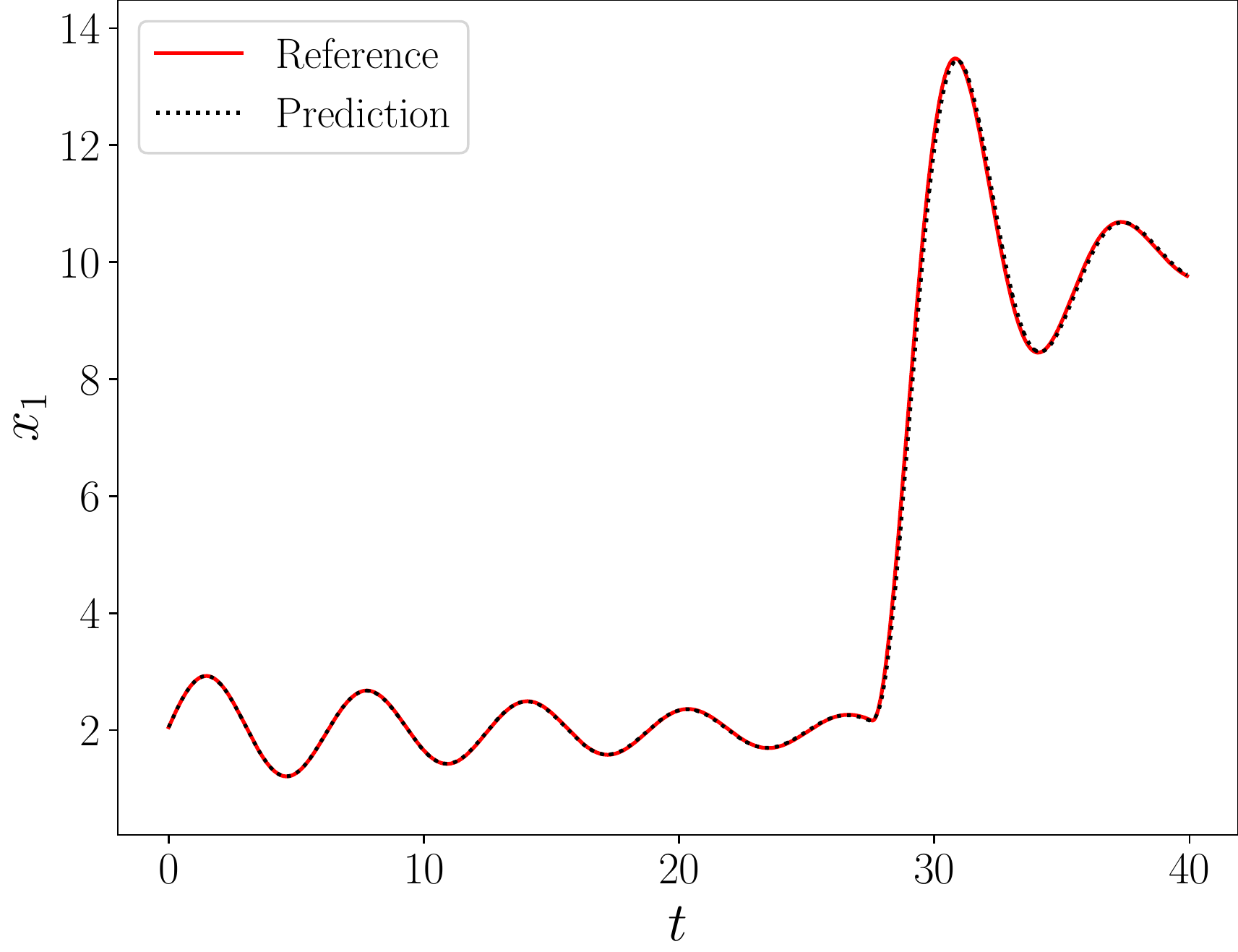}
      \label{fig:osci-x1-k7}
      }
  \subfigure[Iteration step 7, $x_2(t)$]{
      \includegraphics[scale=0.34]{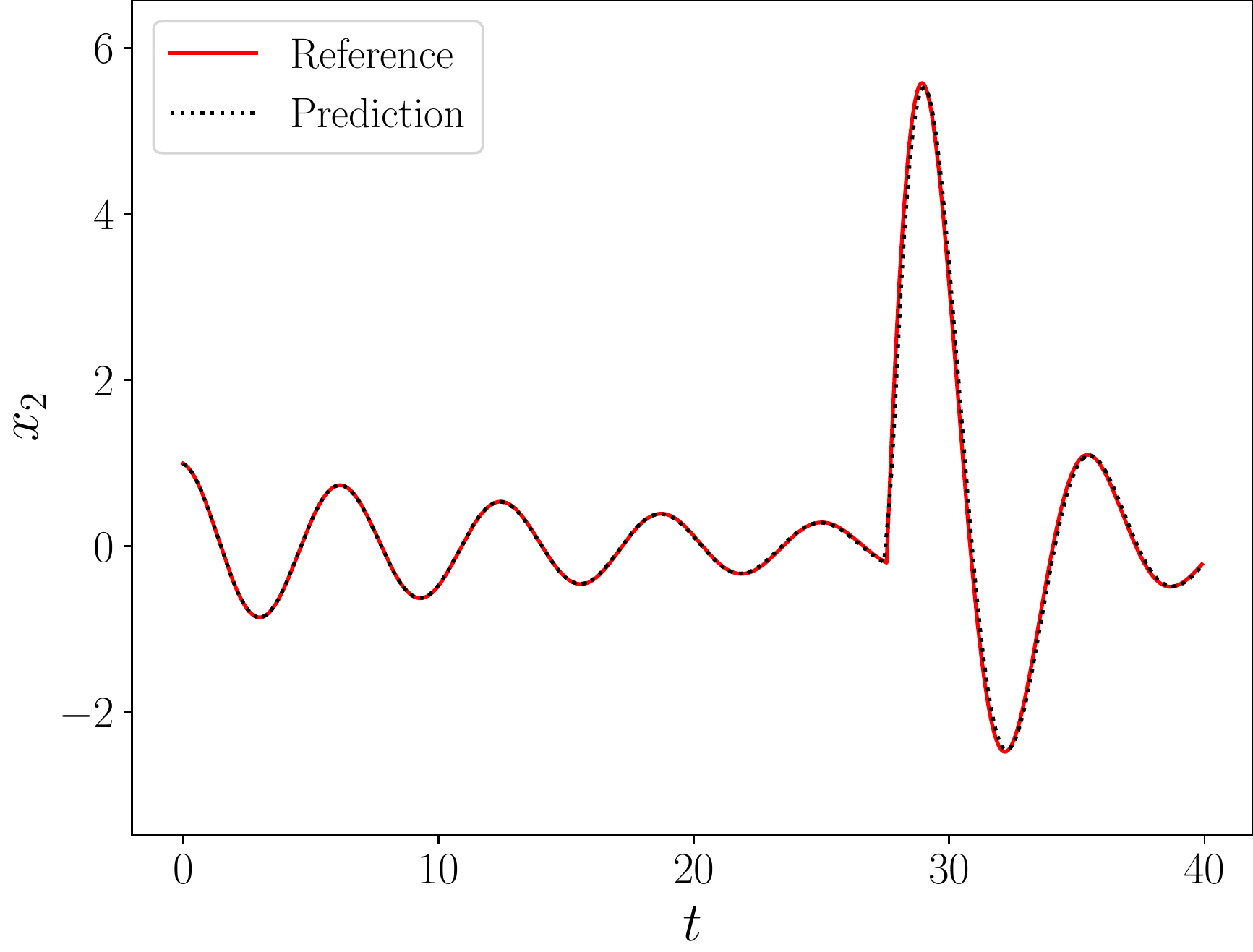}
      \label{fig:osci-x2-k7}
      }
   \caption{The neural network predictions and solutions of the LSODA solver, $\bx_0=[2,1]^T$, forced damped oscillator equation with two subsystems.}
   \label{fig:force-damp-oscillator-plots}
\end{figure}

To assess the effectiveness of our DNN-AL approach, the orbit with the initial state $\bx_0=[2,1]^T$ is considered, which is not included in our observed datasets. A reference result for this orbit is computed with the LSODA solver for comparison.  
\figurename{\ref{fig:force-damp-oscillator-plots}} shows the orbits obtained by DNN-AL and the LSODA solver. From \figurename{\ref{fig:force-damp-oscillator-plots}}(a) (for $x_1$) and \figurename{\ref{fig:force-damp-oscillator-plots}}(b) (for $x_2$), it can be seen that at adaptivity iteration step one of DNN-AL, i.e., just using a deep RestNet to approximate the $\Delta$-lag flow map of the whole system, the predictions of the neural networks are inaccurate, especially for the states at the time after the switching time instant $t=27.6$. \figurename{\ref{fig:force-damp-oscillator-plots}}(c) and \figurename{\ref{fig:force-damp-oscillator-plots}}(d) show that, at adaptivity iteration step three in DNN-AL, the predictions of the neural networks get closer to the results of the LSODA solver. From \figurename{\ref{fig:force-damp-oscillator-plots}}(e) and \figurename{\ref{fig:force-damp-oscillator-plots}}(f), it is clear that, after seven adaptivity iteration steps, predictions of the neural networks obtained by DNN-AL and the results of the LSODA solver are visually indistinguishable, which shows that DNN-AL effectively learns this switched system.  

\begin{figure}[htp!]
       \centering
       \includegraphics[scale=0.36]{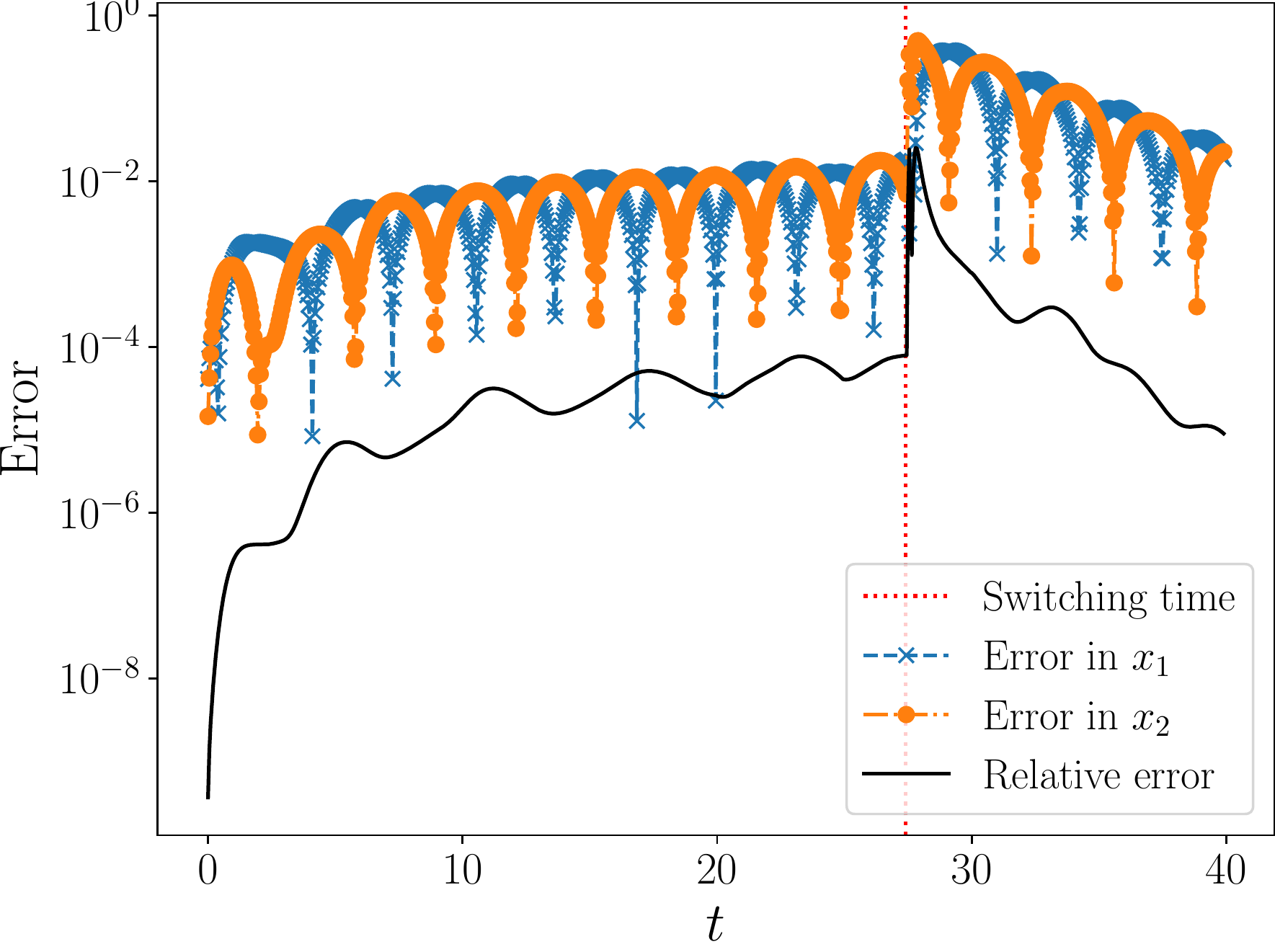}
       \caption{Errors of DNN-AL predictions, forced damped oscillator equation with two subsystems.}
       \label{fig:oscilator-2seg-loss}
\end{figure}

At time $t_j$ (for $j=1,\ldots,J$), let $\bx(t_j)=[x_1(t_j),x_2(t_j)]^T$ be the state obtained by the LSODA solver, and $\widehat{\by}(t_j)=[\widehat{y}_1(t_j),\widehat{y}_2(t_j)]^T$ be the prediction of the deep ResNets obtained by DNN-AL. The error for $x_1$ is given by $|x_1(t_j) - \widehat{y}_1(t_j)|$, and the error for $x_2$ is given by $|x_2(t_j) - \widehat{y}_2(t_j)|$. 
The relative error 
is defined as \begin{equation}
\label{eq:test-relative-err}
    \ell_{\mathrm{relative}}:=\frac{\|\bx(t_j) - \widehat{\by}(t_j)\|^2}{\|\bx(t_j)\|^2}.
\end{equation}
\figurename{\ref{fig:oscilator-2seg-loss}} shows the errors for this test problem where seven adaptivity iterations are conducted in DNN-AL. 
The approximate switching time instant is $\breve{t}=27.5$ (see Definition \ref{def:auxiliary-system}), which is very close to the true switching time instant $27.6$. It can be seen that the errors are very small in the time interval $(0,27.5]$, which are mainly caused by the neural network approximation error. There is a clear jump in the relative error (and the errors for $x_1$ and $x_2$) around the switching time instant ($t=2.76$). This is because the trained deep ResNets are built with the approximate switching time instant $\breve{t}=27.5$. However, the relative error is still bounded (clearly less than $10^{-1}$). These results are consistent with Theorem \ref{thm:nn-error-bound}.  

We next consider the situation of noisy data. The noise in \eqref{eq:each-trajectory} is set to $\epsilon^{(n)}_j:=\xi^{(n)}_j\bx(t_{j};\bx_0^{(n)})$, where $\xi^{(n)}_j$ is drawn from the uniform distribution over $[0,\,\eta_{\mathrm{noise}}]$ and $\eta_{\mathrm{noise}}$ is a given positive number. The test set is again set to the orbit with the initial state $\bx_0=[2,\,1]^T$. To quantify the accuracy of the neural networks obtained through DNN-AL,  
the mean square error (MSE) is considered, which is defined as  
\begin{equation}
    \label{eq:test-mse}
    \ell_{\mathrm{mse}}:=\frac{1}{J}\sum_{j=1}^J \|\bx(t_j) - \widehat{\by}(t_j)\|^2,
\end{equation} 
where $\bx(t_j)$ is the state obtained by the LSODA solver, and 
$\widehat{\by}(t_j)$ is the prediction of the neural networks.
The MSEs for this test problem are shown in Table \ref{tab:err-noise}, where different ranges of the noises are tested. In this table, the test MSE is defined in \eqref{eq:test-mse}, and the validation MSE is computed through replacing the test set for \eqref{eq:test-mse} by the validation set (see \eqref{eq:initial-dataset}). Table \ref{tab:err-noise} shows that the validation MSE is smaller than the test MSE, which is expected. 
It also can be seen that the test MSE increases approximately in proportion to the noise, which shows that the performance of the trained neural networks of DNN-AL is stable with respect to the noise.
\begin{table}[htp!]
    \centering
    \begin{tabular}{cccc}
    \hline
    $\eta_{\mathrm{noise}}$ & 2e-2  & 5e-2 & 1e-1\\
    \hline
   Validation MSE & 4.769e-03 &4.645e-03 &6.731e-03\\
   Test MSE &2.442e-02 &4.451e-02&1.061e-01\\
   \hline
    \end{tabular}
    \caption{Errors with different noisy levels, forced damped oscillator equation with two subsystems.}
    \label{tab:err-noise}
\end{table} 

\begin{figure}[htp!]
\centering  	
\subfigure[Iteration step 3, left]{
   \includegraphics[scale=0.33]{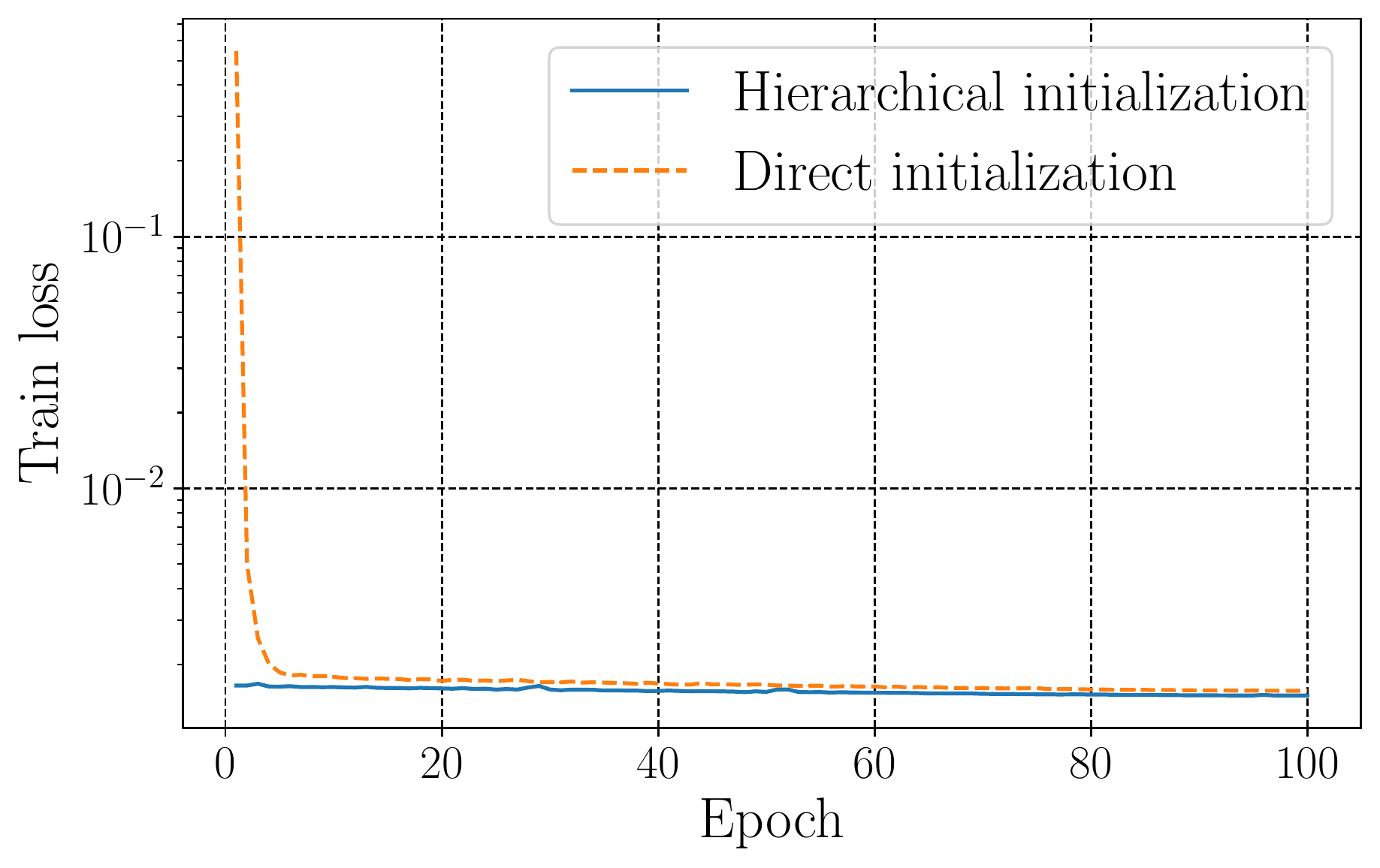}
   }
\subfigure[Iteration step 3, right]{
   \includegraphics[scale=0.33]{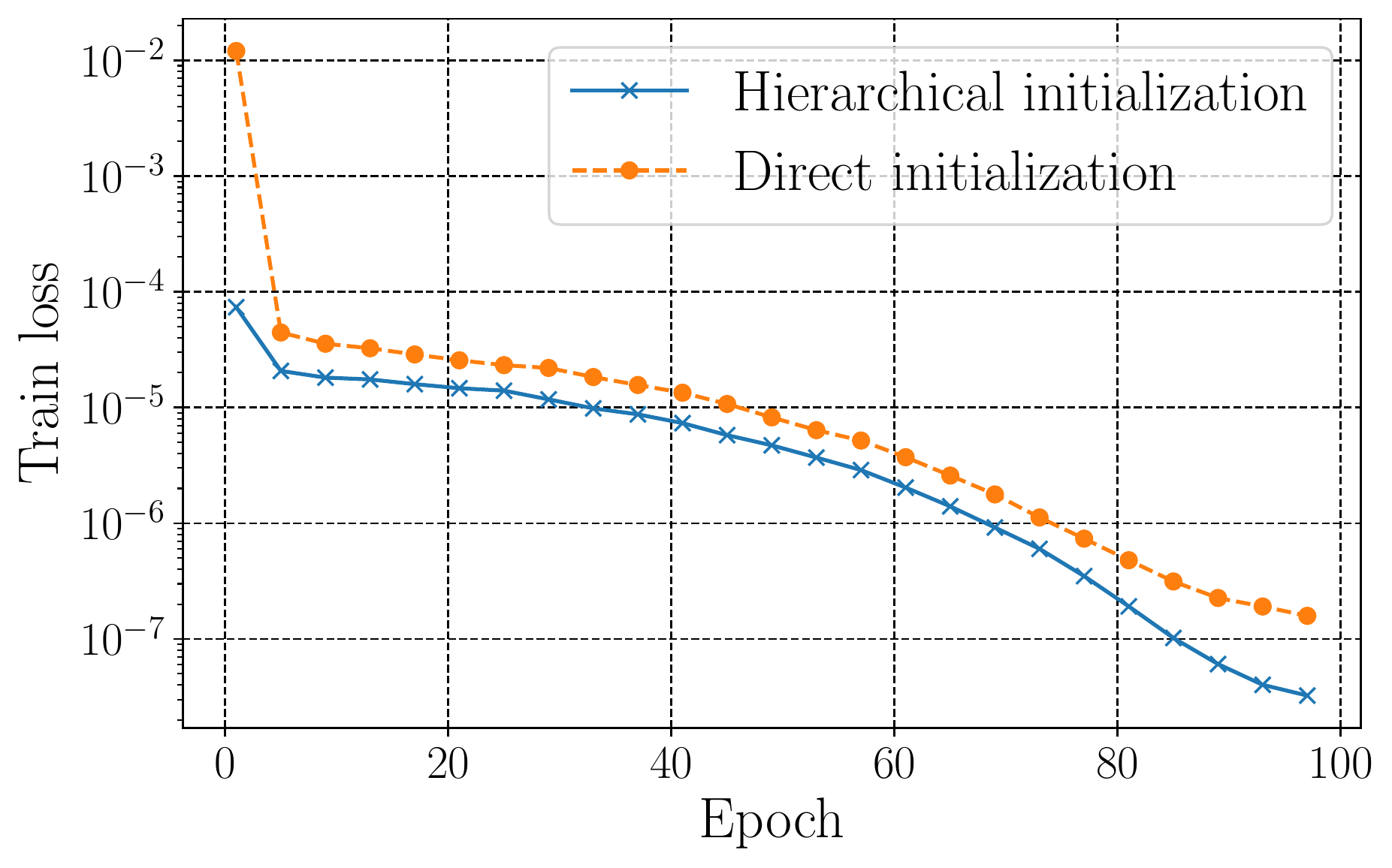}
   }

\subfigure[Iteration step 5, left]{
   \includegraphics[scale=0.33]{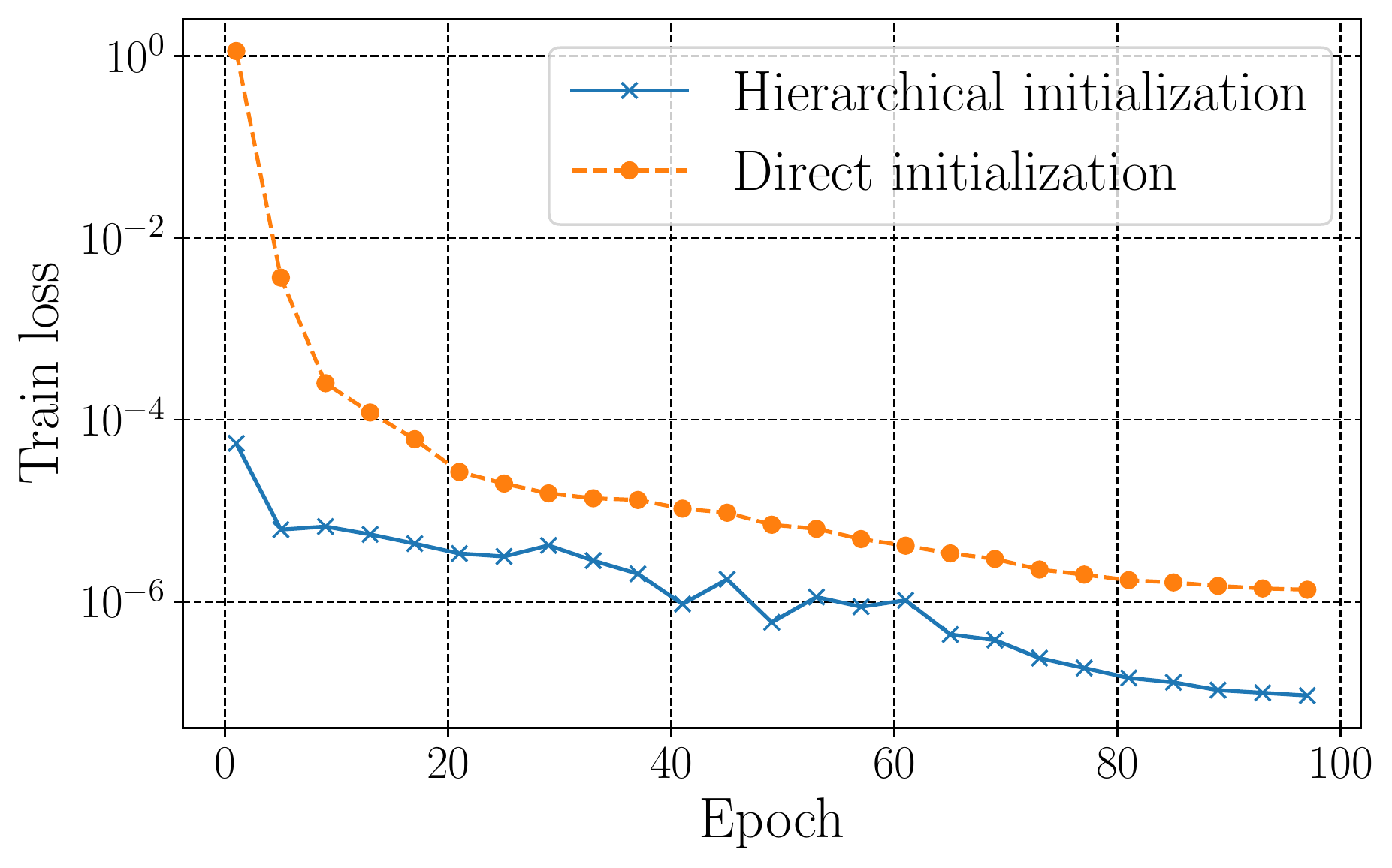}
   }
\subfigure[Iteration step 5, right]{
   \includegraphics[scale=0.33]{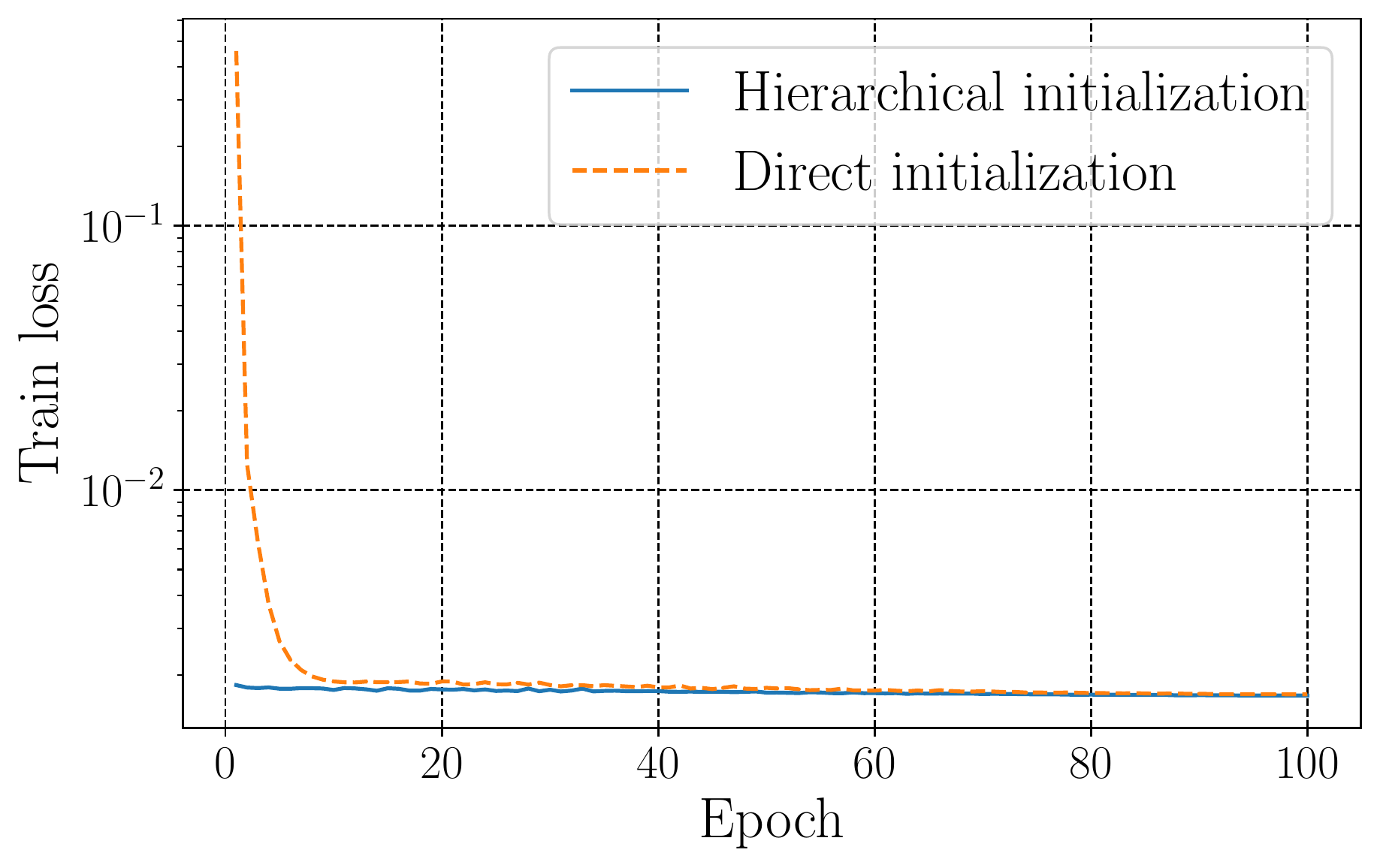}
   }
   
\subfigure[Iteration step 7, left]{
   \includegraphics[scale=0.33]{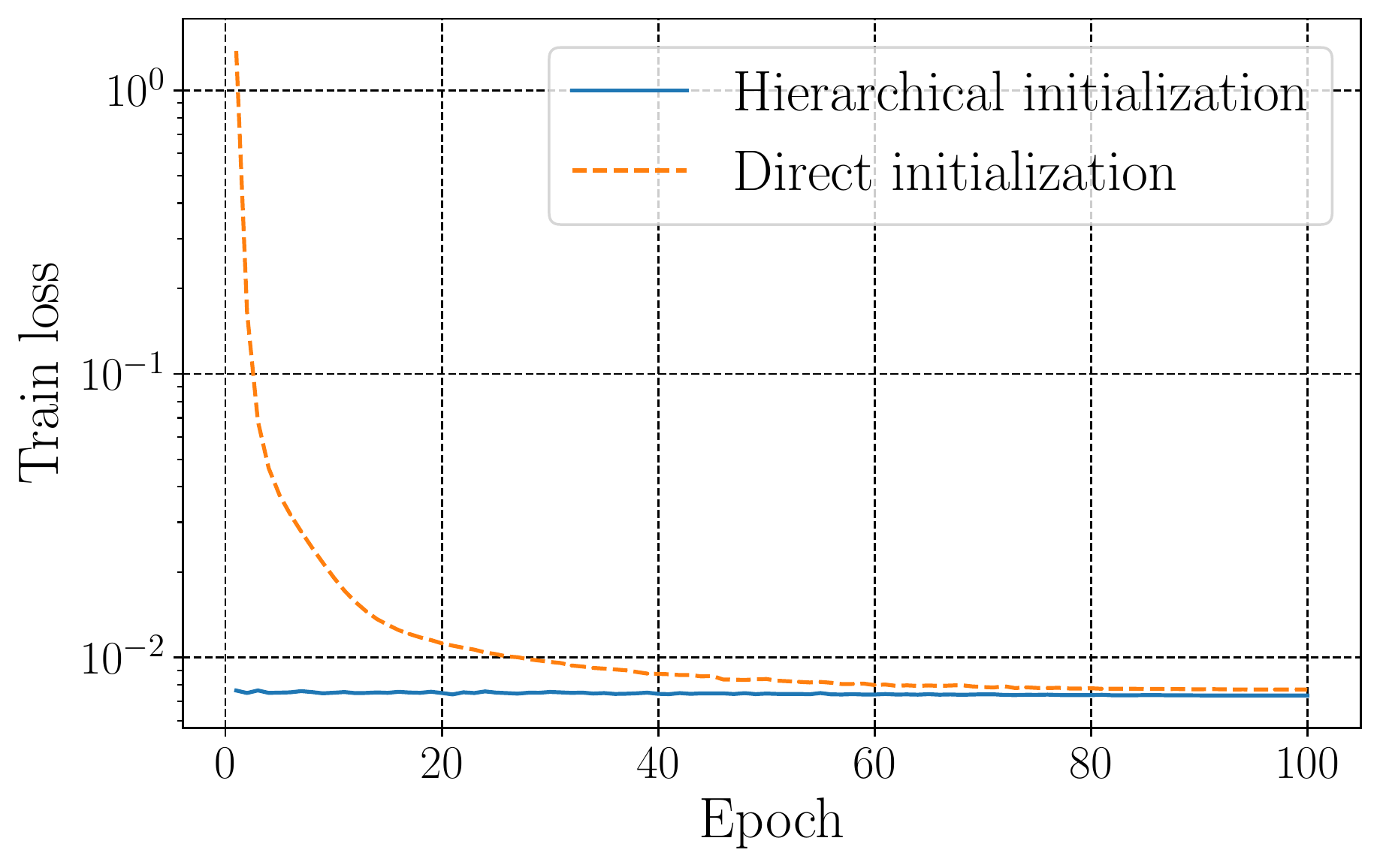}
   }
\subfigure[Iteration step 7, right]{
   \includegraphics[scale=0.33]{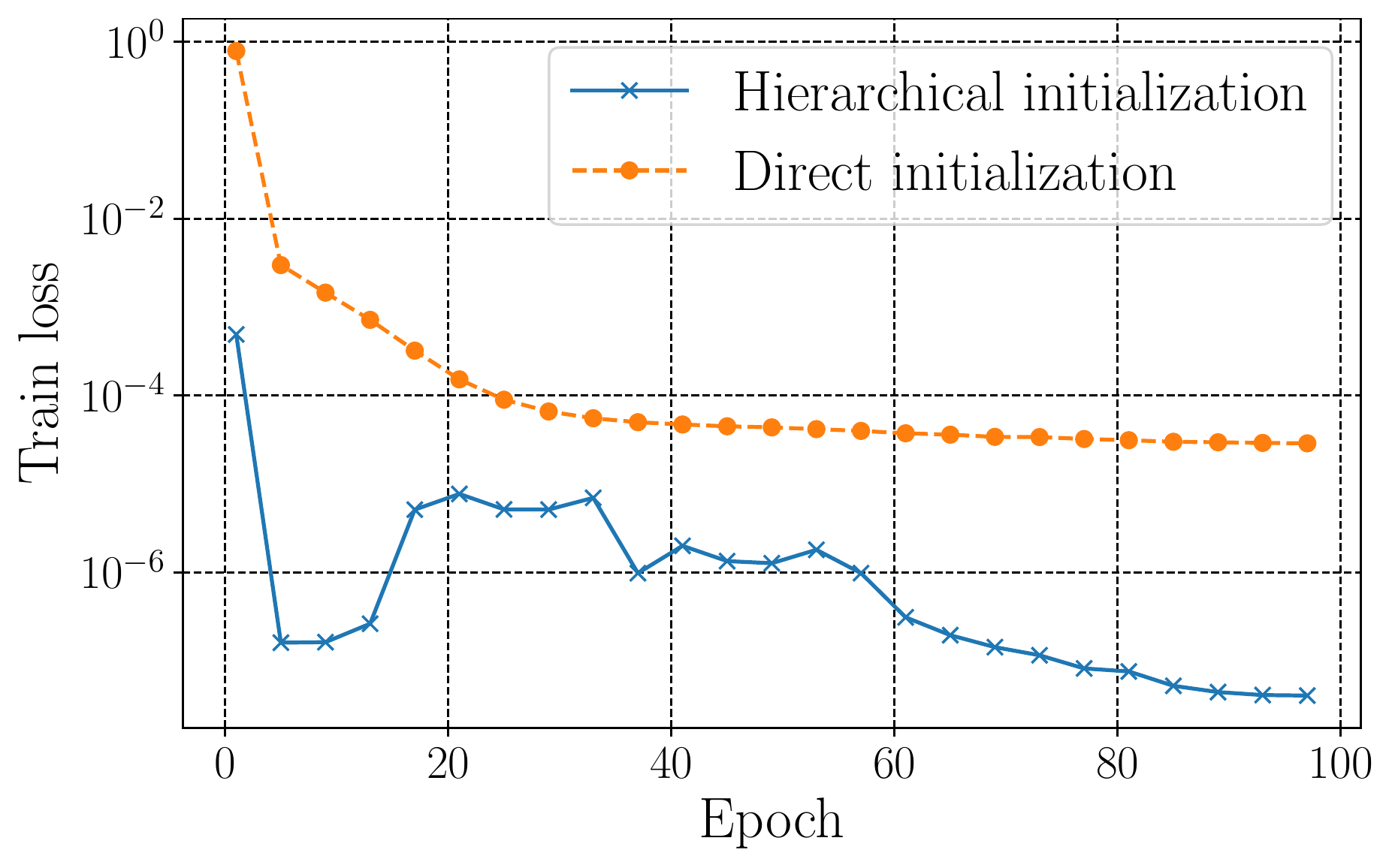}
   }

\caption{Training loss at different adaptivity iteration steps, forced damped oscillator equation with two subsystems.}
\label{fig:adaptive-train-loss}
\end{figure}

\begin{figure}[htp!]
\centering  	
\includegraphics[scale=0.35]{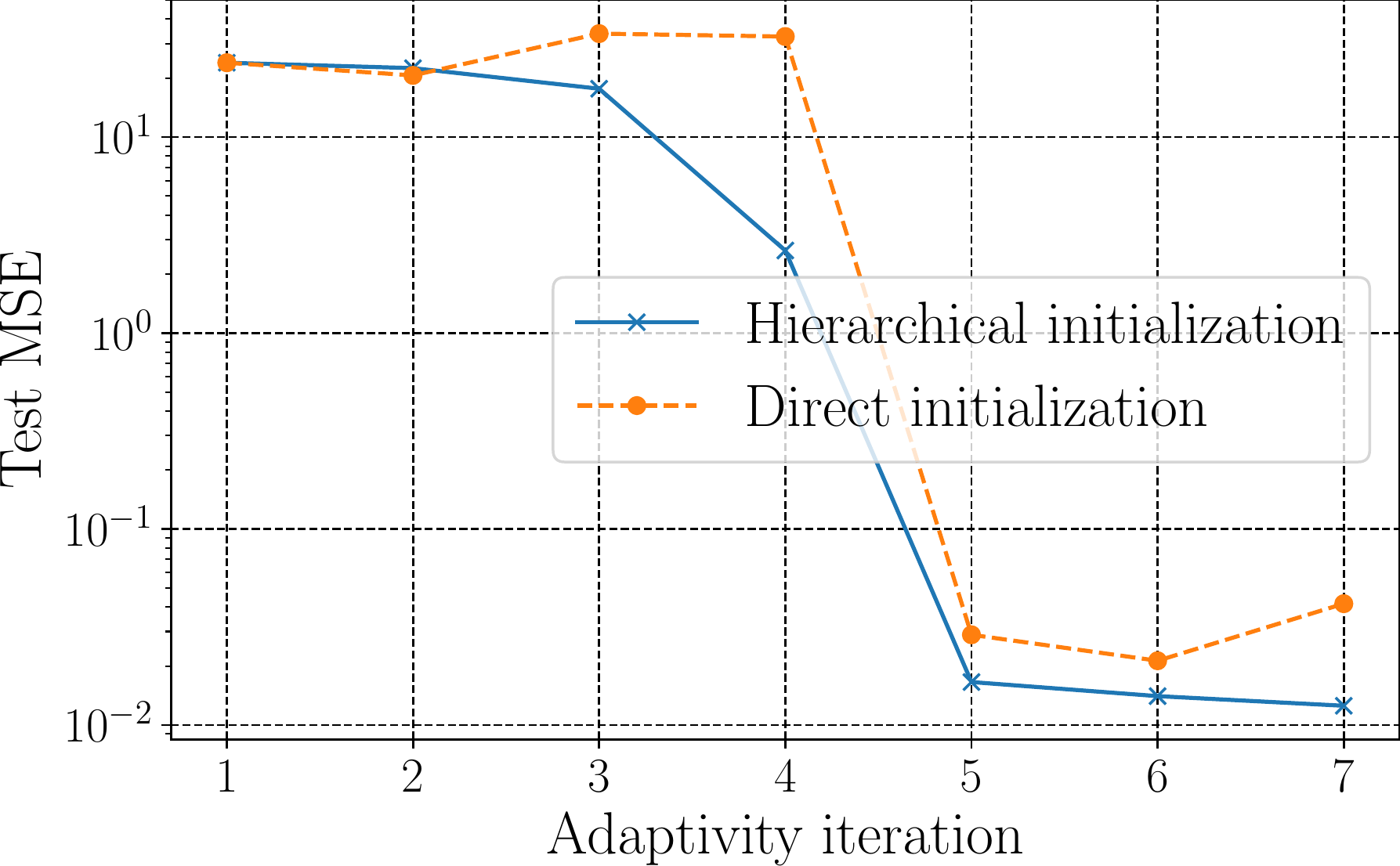}
\caption{Test errors at different adaptivity iteration steps, forced damped oscillator equation with two subsystems.}
\label{fig:adaptive-test-error}
\end{figure}

The efficiency of our hierarchical initialization strategy is investigated herein. As shown on line $13$ of Algorithm \ref{alg:main-alg}, the initial deep ResNets are hierarchically constructed using the trained ResNets at the previous adaptivity iteration steps. For comparison, a direct initialization strategy is tested, which is to use the standard Kaiming initialization \cite{he2015delving} to construct the initial ResNets at every adaptivity iteration step. 
As the identified interval with the largest validation error is decomposed into two parts at each adaptivity iteration step (line $9$ of Algorithm \ref{alg:main-alg}), two new deep ResNets need to be trained, which are referred to as the left and the right ones. 
\figurename{\ref{fig:adaptive-train-loss}} shows the values of the training loss \eqref{eq:train-loss} at adaptivity iteration step three, five and seven respectively. 
It can be seen that our hierarchical initialization strategy results in smaller training loss values than the direct initialization approach. 
\figurename{\ref{fig:adaptive-test-error}} shows the test MSEs (see \eqref{eq:test-mse}) associated with the two initialization strategies, where the initial state is $\bx_0=[2,\,1]^T$.  
It is clear that our hierarchical initialization results in smaller test errors after two adaptivity iteration steps.

\subsection{Forced damped oscillator with three subsystems}
Here, we consider the forced damped oscillator problem with three subsystems. That is, the equation \eqref{eq:oscillator} is considered again, and the signal function and the governing equations are set to
\[\sigma(t)=
\begin{cases}
1\,,\quad t\in (0,17.4]\\
2\,,\quad t\in (17.4,27.6]\\
3\,,\quad t\in (27.6,40],
\end{cases} 
\]
and 
\[
    \begin{aligned}
    \mathf^{(1)}=\left ( \begin{array}{cc}
     x_2  \\
     2-0.1x_2-x_1 
\end{array} \right),\mathf^{(2)}=\left ( \begin{array}{cc}
     x_2  \\
     4-0.2x_2-x_1 
\end{array} \right),\mathf^{(3)}=\left ( \begin{array}{cc}
     x_2  \\
     8-0.4x_2-x_1 
\end{array} \right).
    \end{aligned}
\] 
Other settings for this test problem are the same as those for the two-subsystem test problem in Section \ref{sec:test1}, and $200$ trajectories with different initial states are generated to construct the observed data. The neural network structure  and the settings for Algorithm \ref{alg:main-alg} are the same as those in Section \ref{sec:test1}.

\begin{figure}[htp!]
\centering
\subfigure[Iteration step 2, $x_1(t)$]{
   \includegraphics[scale=0.34]{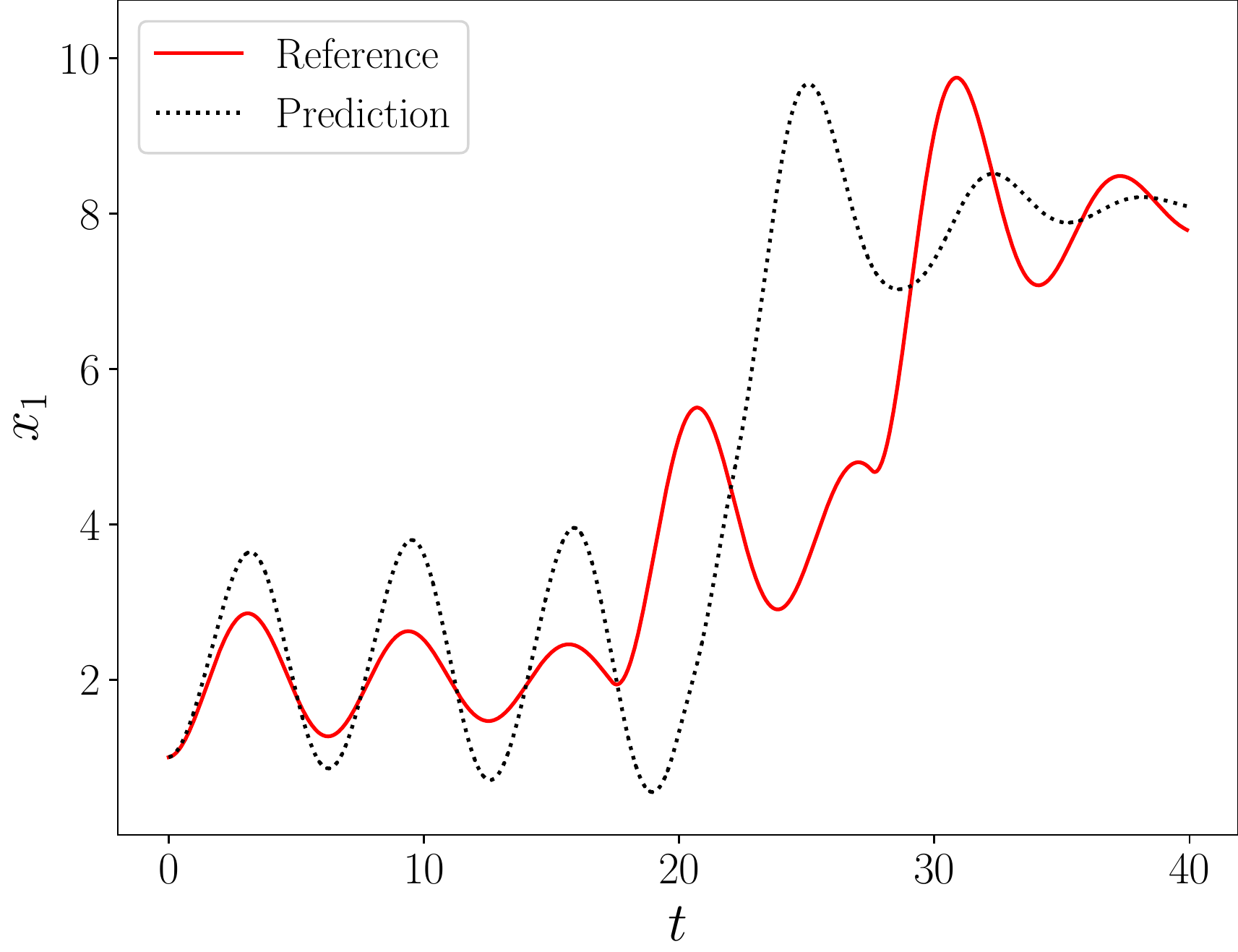}
   \label{fig:osci-3seg-x1-k2}
   }
\subfigure[Iteration step 2, $x_2(t)$]{
   \includegraphics[scale=0.34]{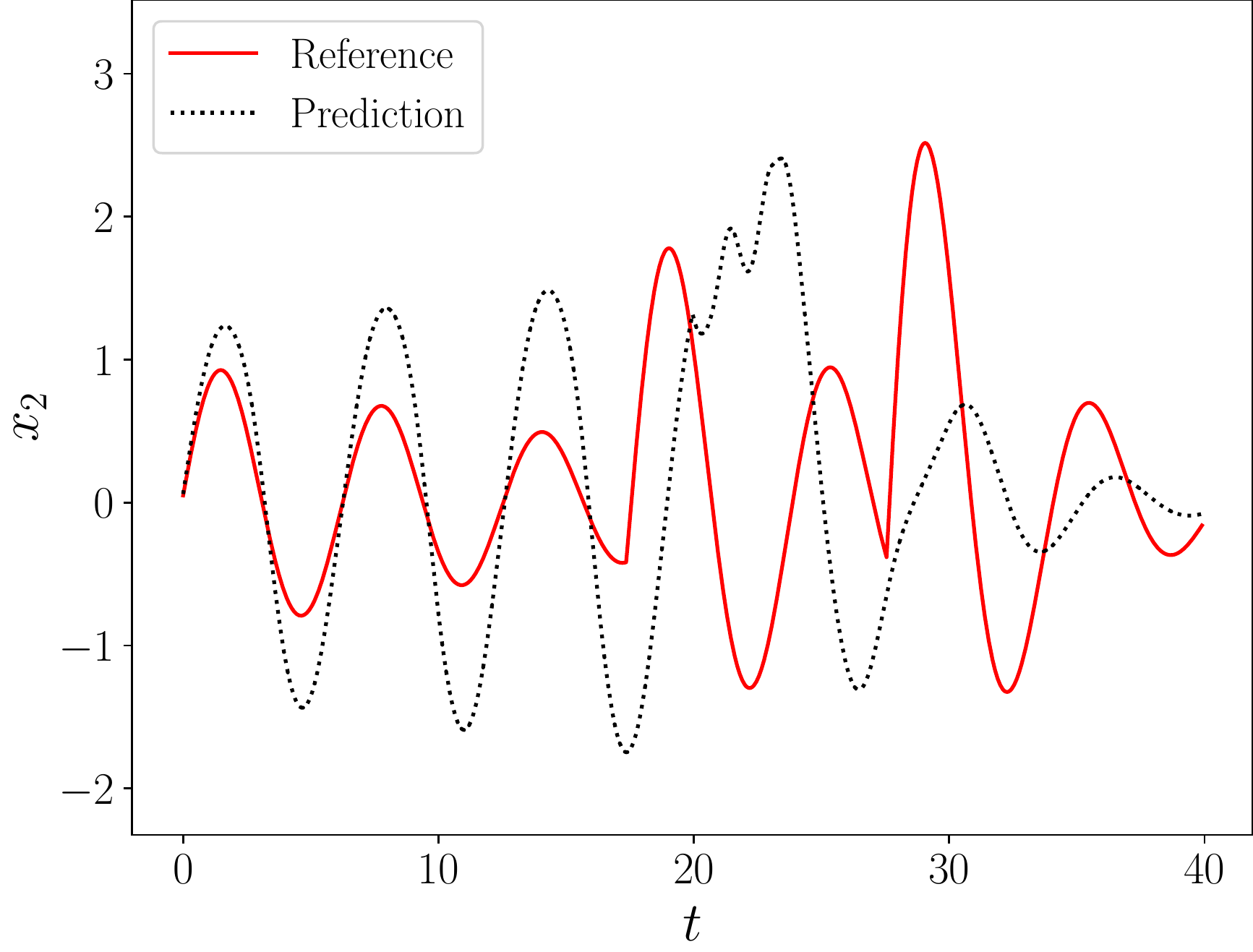}
   \label{fig:osci-3seg-x2-k2}
   }
\subfigure[Iteration step 6, $x_1(t)$]{
   \includegraphics[scale=0.34]{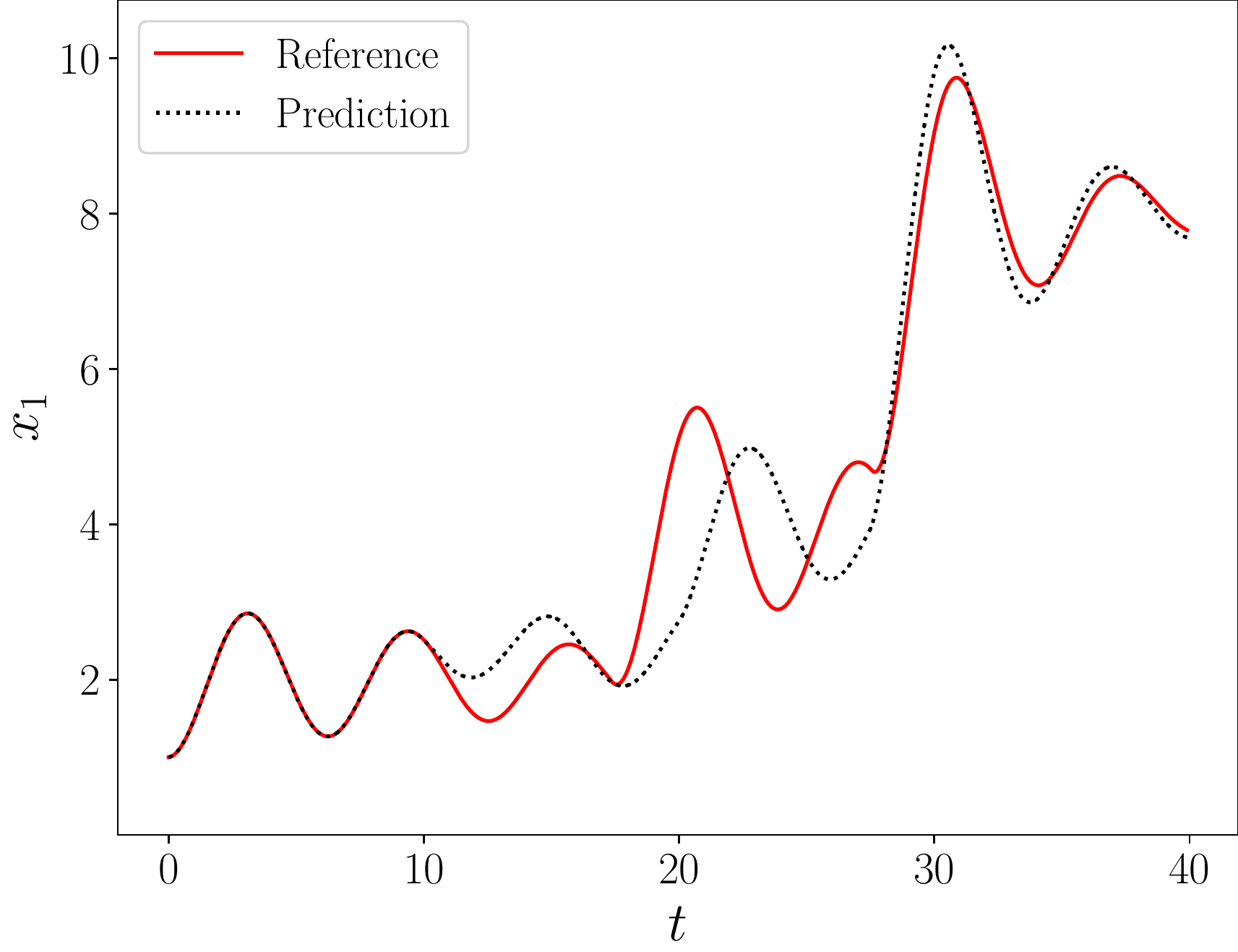}
   \label{fig:osci-3seg-x1-k6}
   }
\subfigure[Iteration step 6, $x_2(t)$]{
   \includegraphics[scale=0.34]{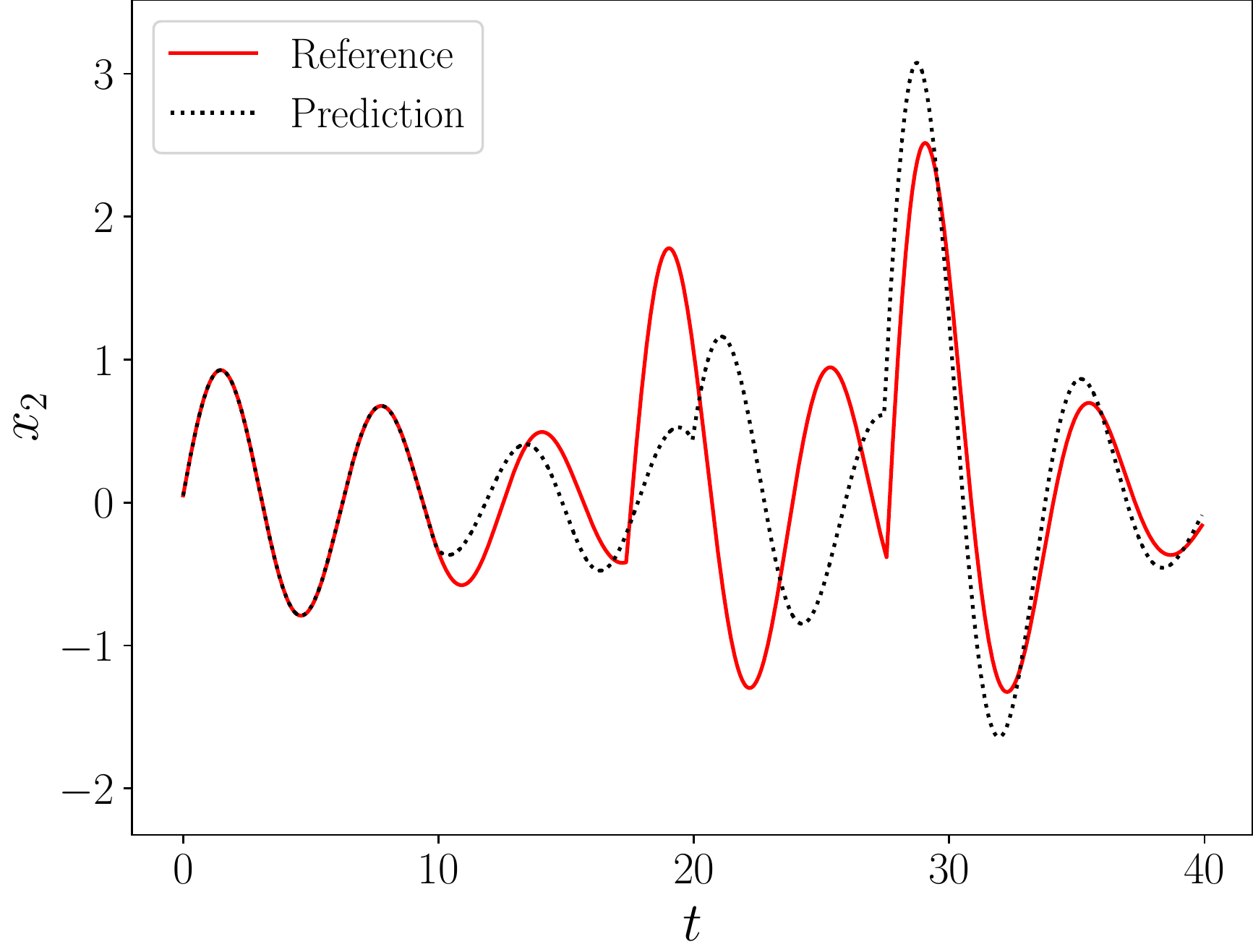}
   \label{fig:osci-3seg-x2-k6}
   }
\subfigure[Iteration step 8, $x_1(t)$]{
   \includegraphics[scale=0.34]{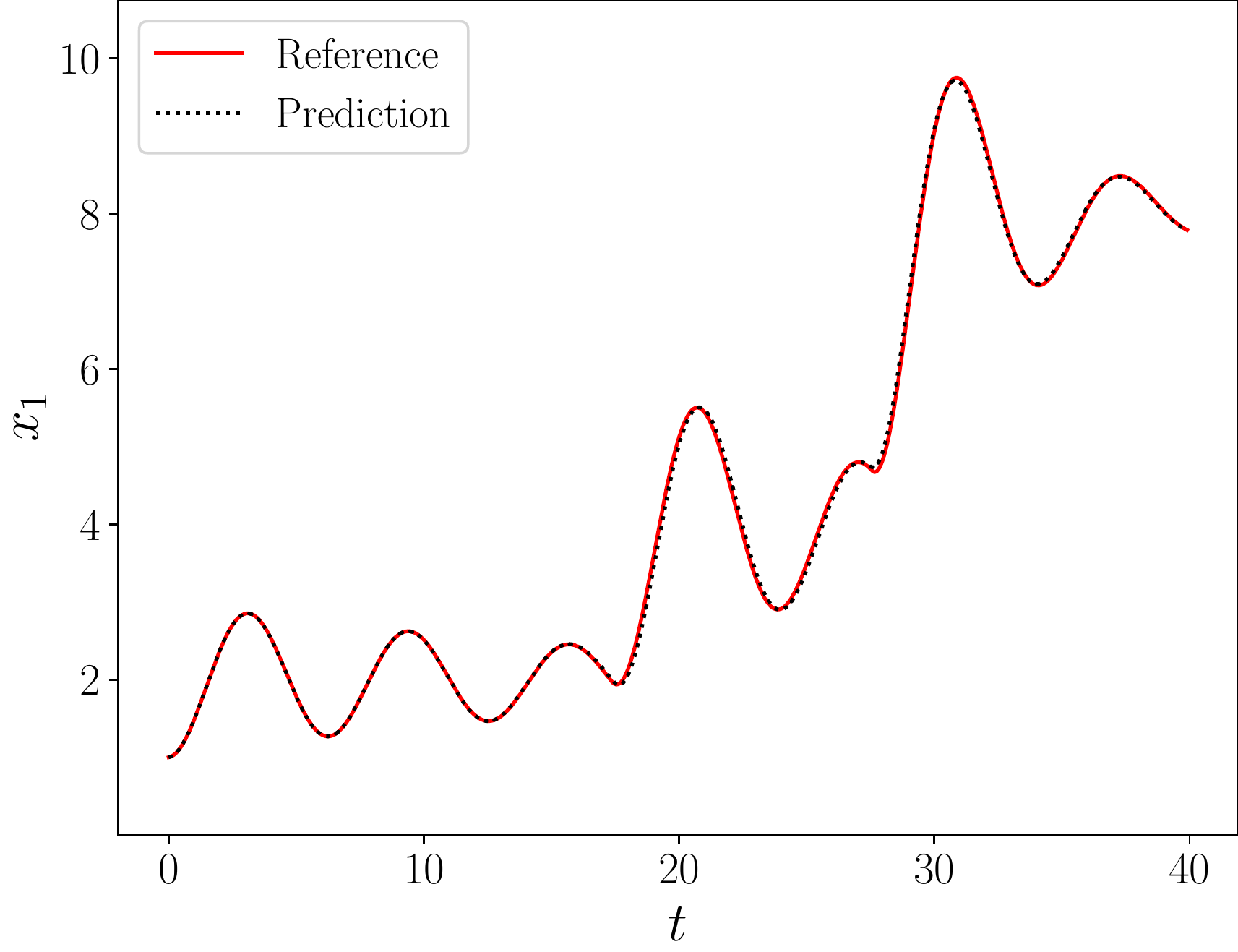}
   \label{fig:osci-3seg-x1-k8}
   }
\subfigure[Iteration step 8, $x_2(t)$]{
   \includegraphics[scale=0.34]{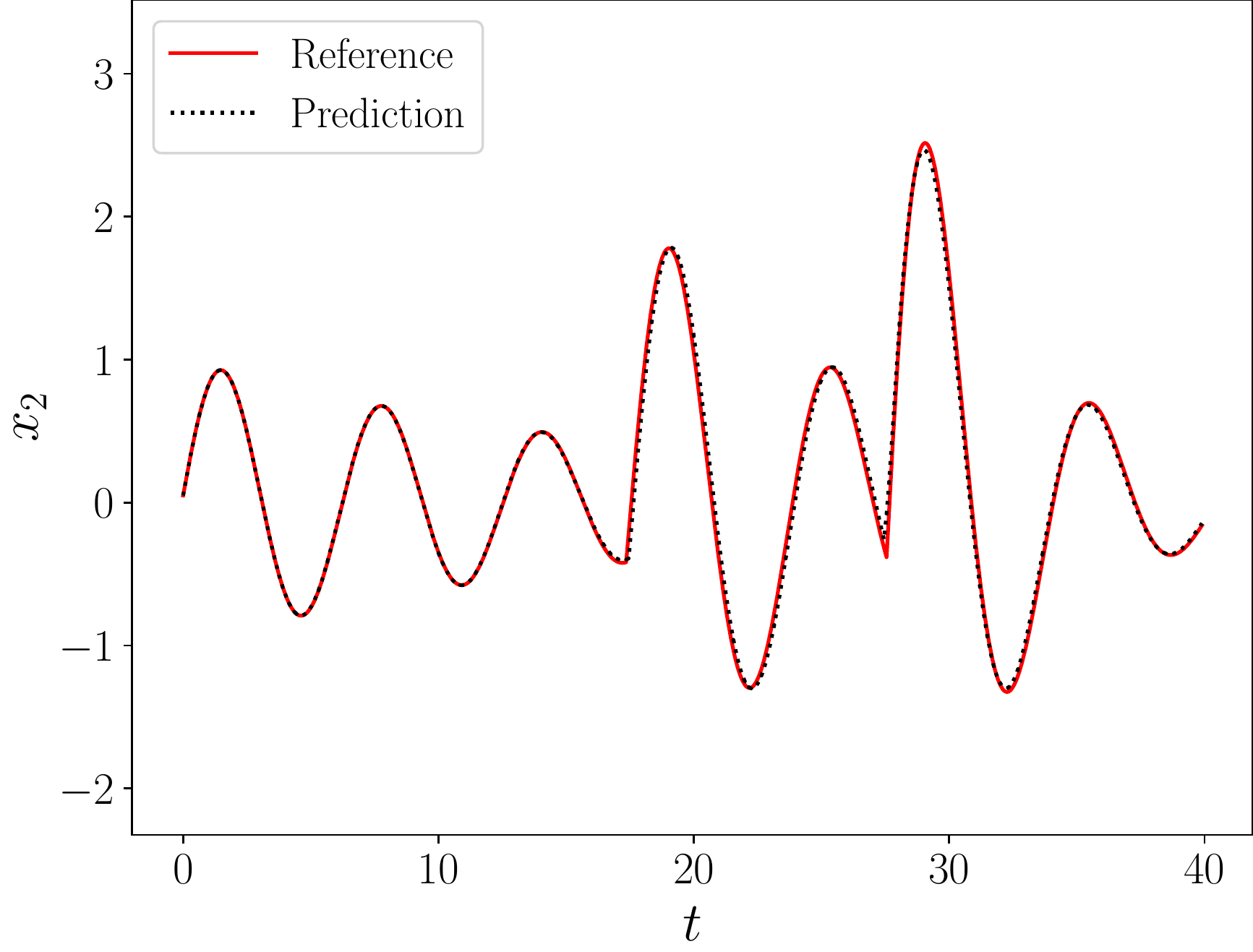}
   \label{fig:osci-3seg-x2-k8}
   }
\caption{The neural network predictions and solutions of the LSODA solver, $\bx_0=[1,0]^T$, forced damped oscillator equation with three subsystems.}
\label{fig:force-damp-oscillator-3seg-plots}
\end{figure}

To test the effectiveness of our DNN-AL approach, the orbit with the initial state $\bx_0=[1,0]^T$ is considered, which is not included in the observed dataset. \figurename{\ref{fig:force-damp-oscillator-3seg-plots}} shows the reference solution (generated by the LSODA solver) and the predictions of the trained deep ResNets obtained through Algorithm \ref{alg:main-alg}, where the trajectories at adaptivity iteration step two, six, and eight are presented. It is clear that, as the adaptivity iteration step increases, the predictions of the trained deep ResNets get closer to the reference results. 
Especially at adaptivity iteration step eight, the predictions of the trained deep ResNets and the reference results are almost indistinguishable.  

\subsection{Forced damped pendulum} 
The ordinary differential equation system for the forced damped pendulum considered is, 
\begin{equation}
  \left\{
  \begin{array}{ll}
  \frac{\mathrm{d}}{\mathrm{d}t} x_1=x_2,\\
  \frac{\mathrm{d}}{\mathrm{d}t} x_2=-a x_2 - g\sin x_1 + b , 
  \end{array}
  \right.
  \label{eq:pendulum}
  \end{equation} where $-a x_2$ represents a friction force, $g$ is a gravity constant in a constant gravitational field, and $ b$ refers to an external force. 
Following the generic notation in \eqref{eq:switch-signal}--\eqref{eq:swithed-system}, the two governing equations of the subsystems for this test problem are,
\[
\mathf^{(1)} =
  \left ( \begin{array}{cc}
     x_2  \\
     -0.15x_2 - 9.8\sin x_1
\end{array} \right),\,
\mathf^{(2)}=  
  \left ( \begin{array}{cc}
     x_2  \\
     -0.15 x_2 - 9.8\sin x_1 + 2
\end{array} \right),\]
and the corresponding signal function is set to \[\sigma(t) = \begin{cases}
   1\,,\quad t\in (0,15.2]\\
   2\,,\quad t\in (15.2,40].
\end{cases}\]

The spatial domain is set to  $D=[-\pi/2,\pi/2]\times [-\pi,\pi] $, and the time step is set to $\Delta=0.05$. Similar to Section \ref{sec:setup}, 200 initial states are generated through the uniform distribution with the range $D$, and trajectories associated with these initial states are obtained with the LSODA solver, which are used to construct the datasets (see Section \ref{sec:setup}). 
For this test problem, the neural network in DNN-AL consists of 10 ResNet blocks with 2 fully connected layers per block. For each block, the 2 fully connected layers have 20 nodes and 2 respectively. 

\begin{figure}[htp!]
   \centering
   \subfigure[Iteration step 1, $x_1(t)$]{
      \includegraphics[scale=0.34]{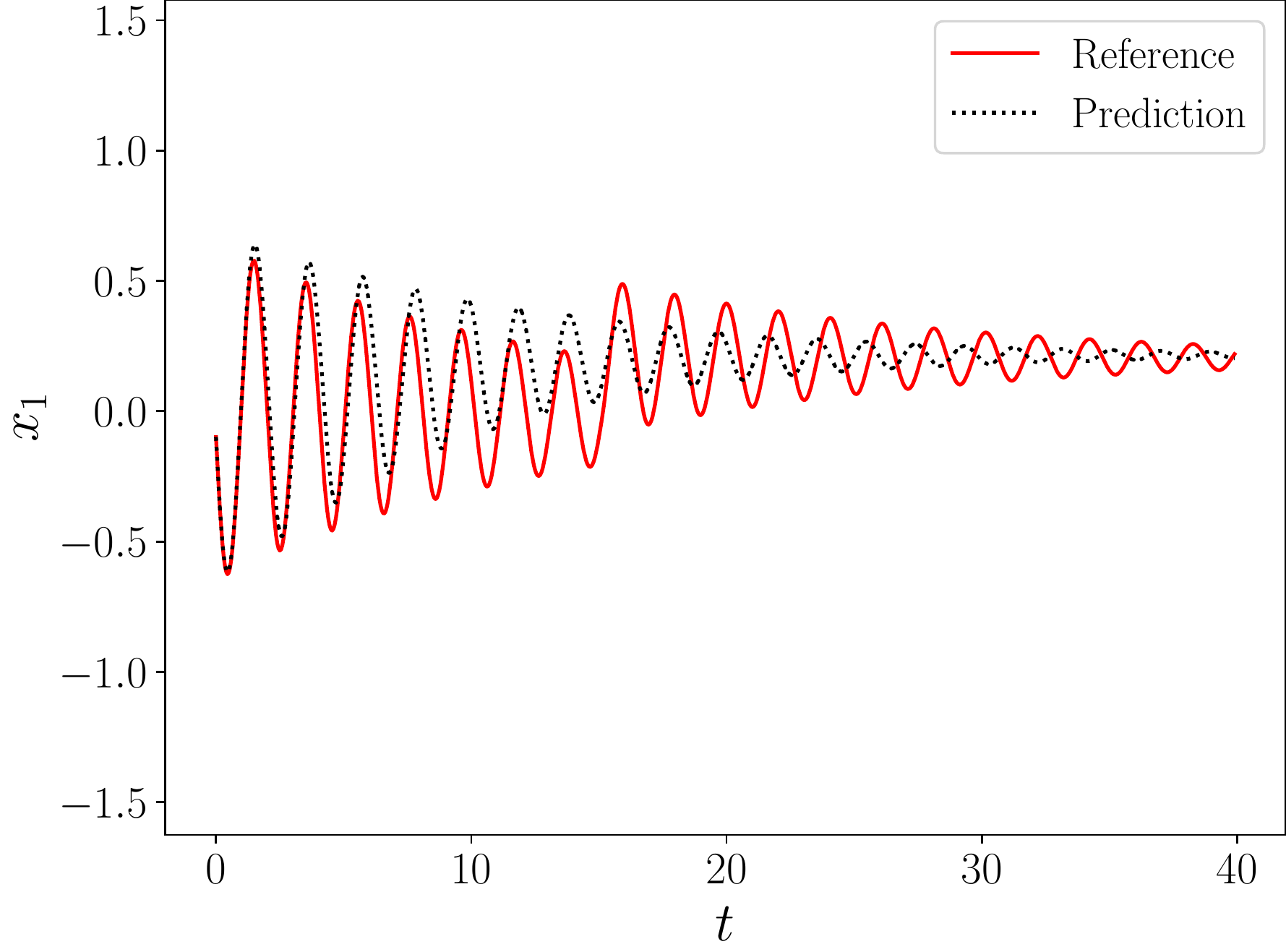}
      \label{fig:pendulum-x1-k1}
   }
   \subfigure[Iteration step 1, $x_2(t)$]{
      \includegraphics[scale=0.34]{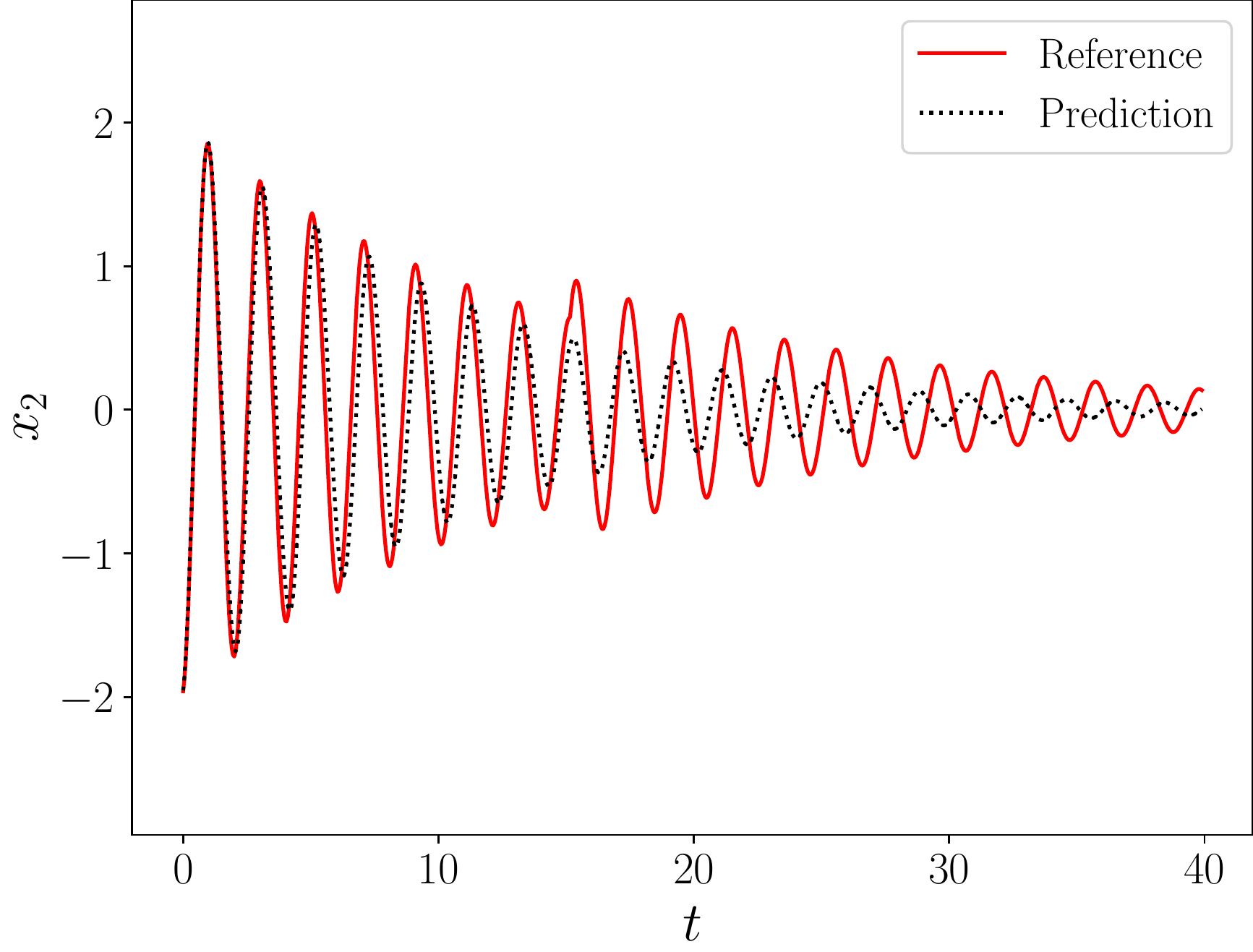}
      \label{fig:pendulum-x2-k1}
   }
    \subfigure[Iteration step 3, $x_1(t)$]{
      \includegraphics[scale=0.34]{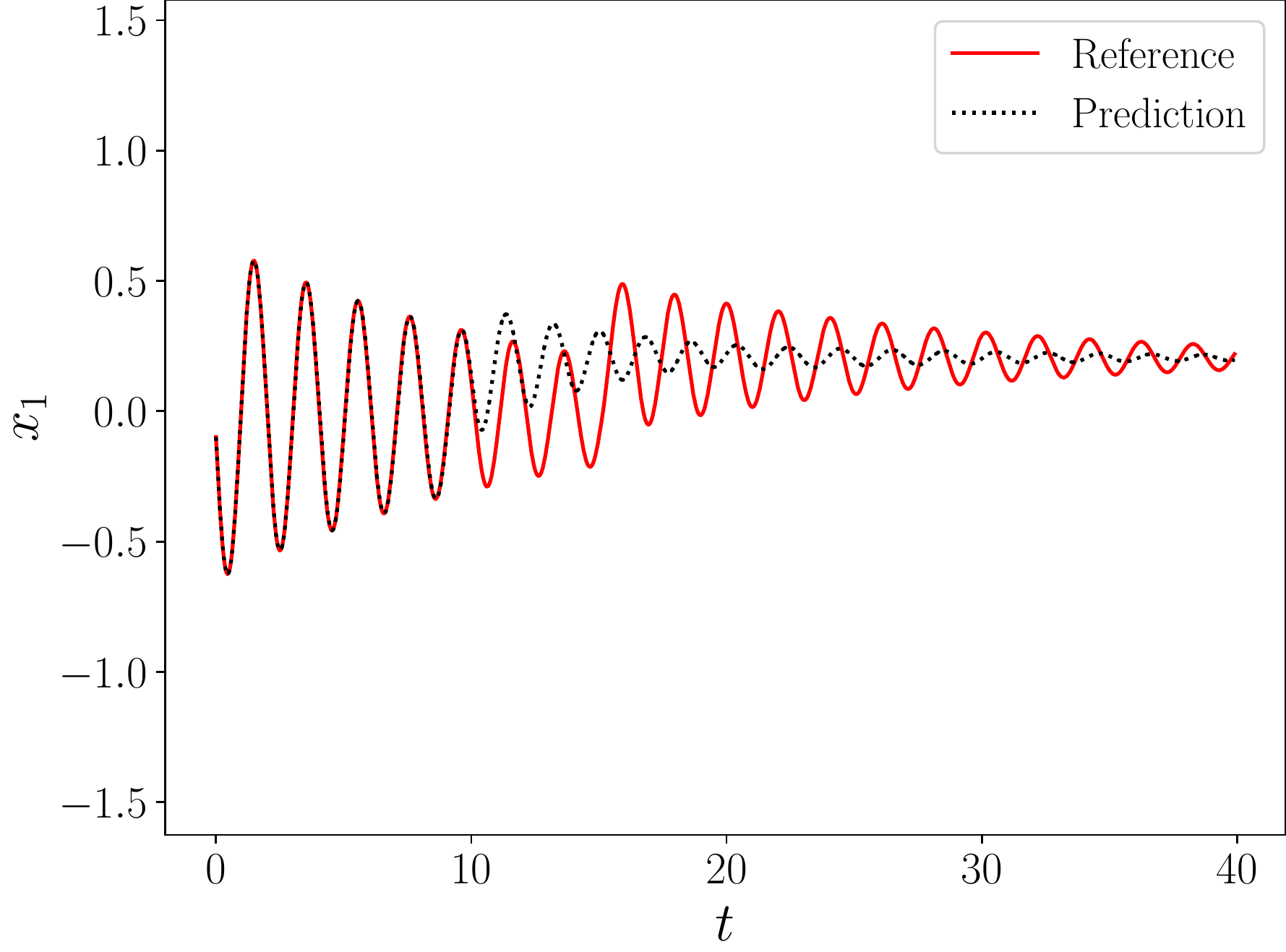}
  }
  \subfigure[Iteration step 3, $x_2(t)$]{
      \includegraphics[scale=0.34]{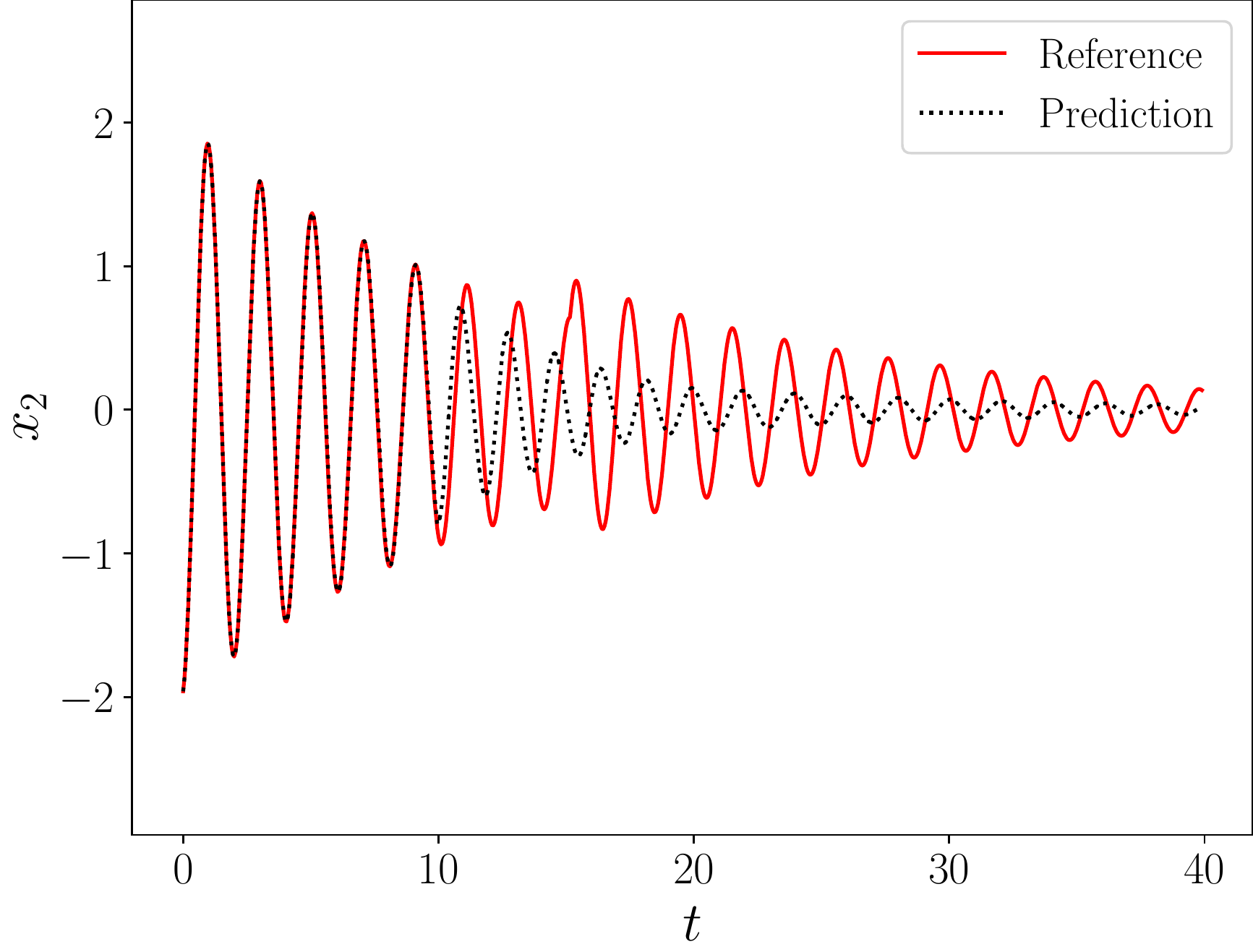}
  }
   \subfigure[Iteration step 5, $x_1(t)$]{
      \includegraphics[scale=0.34]{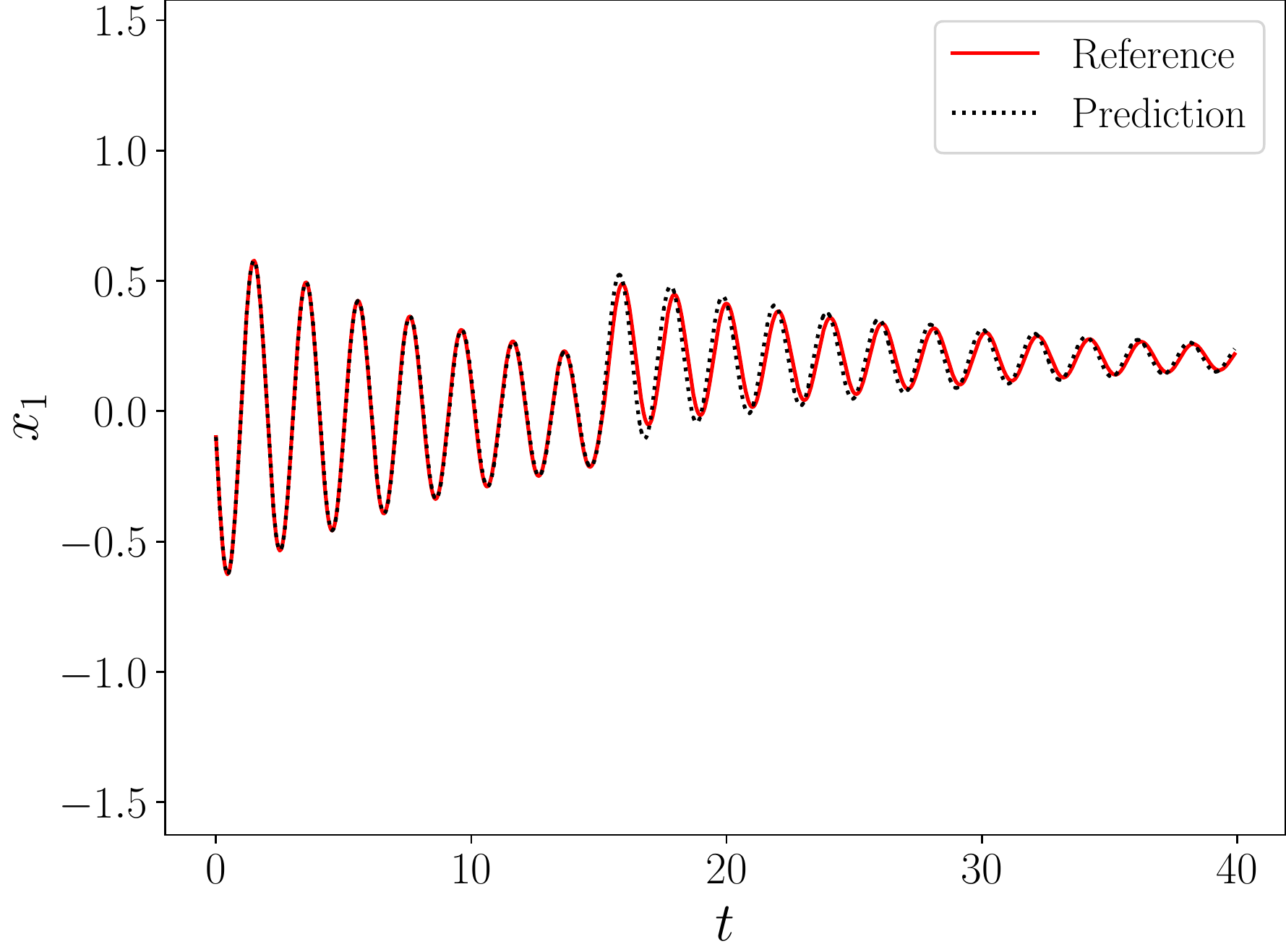}
      \label{fig:pendulum-x1-k5}
   }
   \subfigure[Iteration step 5, $x_2(t)$]{
      \includegraphics[scale=0.34]{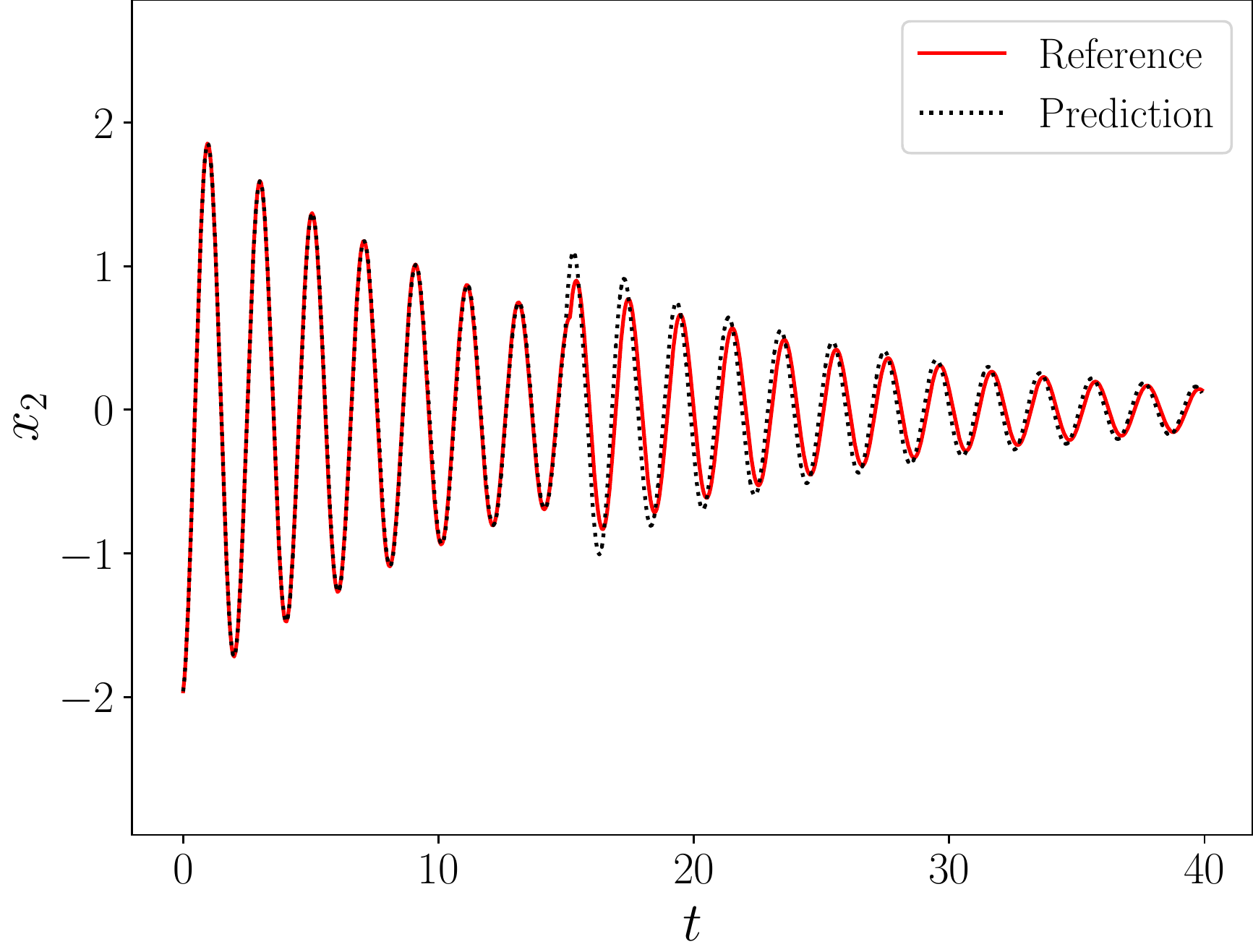}
      \label{fig:pendulum-x2-k5}
   }
  \caption{The neural network predictions and solutions of the LSODA solver, $\bx_0=[0,-2]^T$, forced damped pendulum.}
   \label{fig:force-damp-pendulum-plots}
\end{figure}

For comparison, the orbit with the initial state $\bx_0=[0,-2]^T$ is considered, which is not included in the observed dataset.  \figurename{\ref{fig:force-damp-pendulum-plots}} shows the reference solution (generated by the LSODA solver) and the predictions of the trained deep ResNets obtained through Algorithm \ref{alg:main-alg}, where the trajectories at adaptivity iteration step one, three, and five are presented. It is clear that, as the adaptivity iteration step increases, the predictions of the trained deep ResNets get closer to the reference results, and the predictions of the trained deep ResNets at adaptivity iteration step five are very close to the reference results.

\subsection{Heat equation}
In this test problem, we consider the following heat equation 
\begin{equation}
    \begin{aligned}
    &\frac{\partial u}{\partial t}  =\kappa\frac{\partial^2 u}{\partial x^2}+q(t,\,x)\,,\quad x\in (0,1) \textrm{ and } t\in (0,2],\\
    &u(0,\,x)=u_0(x),\\
    &u(t,\,0)=u(t,\,1)=0,\\
    \end{aligned}
    \label{eq:heat-eqn}
\end{equation}
where $\kappa>0$ is the thermal diffusivity of the medium, $u$ is the unknown temperature and $q(t,\,x)$ is a source function. 
The initial condition is set to 
\begin{eqnarray}
u_0(x) = a\cdot x(1-x), 
\label{eq:heat_a}
\end{eqnarray}
where $a\in \mathbb{R}$ is a parameter. 
To generate data and reference solutions, the spatial domain is discretized with a uniform grid with $I_{\text{grid}}=21$ nodes, and the time lag is set to $\Delta=0.01$ (then $J=201$). 
To result in a discretized problem, $\partial^2 u/\partial x^2$ in \eqref{eq:heat-eqn} is approximated by the second-order centered difference method, and the LSODA solver provided in SciPy \cite{2020SciPy-NMeth} is again applied to solve this problem. 
The thermal diffusivity of the medium $\kappa$ is set to $0.2$.  
The source function $q(t,\,x)$ in \eqref{eq:heat-eqn} is set to
\[
 q(t,x) = \begin{cases}
        20\exp\left(-\frac{(x-1)^2}{0.25}\right),\quad t\in (0,1.2]\\
        10\exp\left(-\frac{(x-1)^2}{0.25}\right),\quad t\in (1.2,2].
     \end{cases}
\] 
The signal function of this test problem is set to 
\[\sigma(t) = \begin{cases}
1,\quad t\in(0,1.2]\\
2,\quad t\in(1.2,2].
\end{cases}\]

To generate the observed datasets,  5000 different values of the parameter $a$ in \eqref{eq:heat_a} are generated using the uniform distribution with the range $[0,1]$, and the corresponding trajectories are computed using SciPy. For DNN-AL, each deep ResNet for this test problem contains 10 ResNet blocks, with two fully-connected layers per block. The two layers have 50 nodes and 21 nodes respectively.

\begin{figure}[htp!]
    \centering
    \includegraphics[scale=0.17]{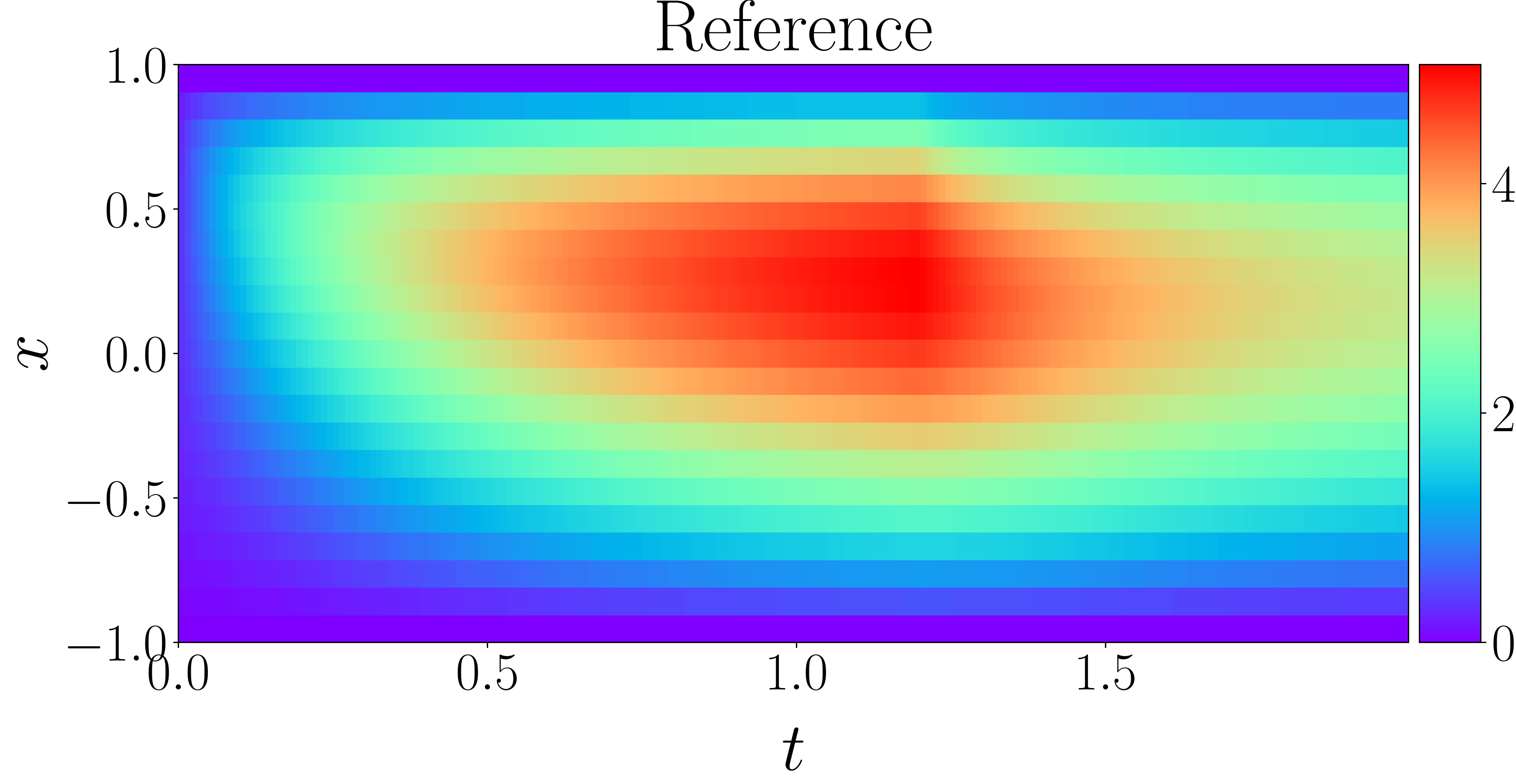}
    \caption{Reference solution, heat equation.}
    \label{fig:heat-eqn-reference}
\end{figure}

\begin{figure}[htp!]
   \centering
      \subfigure[Iteration step 1, $\hat{u}(t,x)$ and $|u(t,x)-\hat{u}(t,x)|$]{
      \includegraphics[scale=0.18]{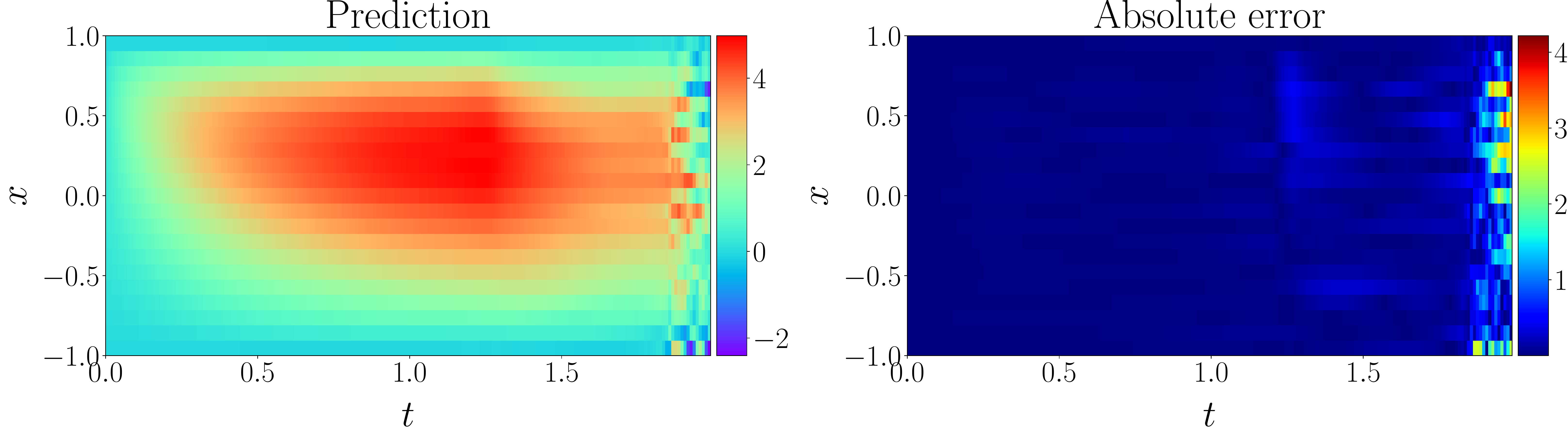}
      \label{fig:heat-k1}
      }
      \subfigure[Iteration step 3, $\hat{u}(t,x)$ and $|u(t,x)-\hat{u}(t,x)|$]{
      \includegraphics[scale=0.18]{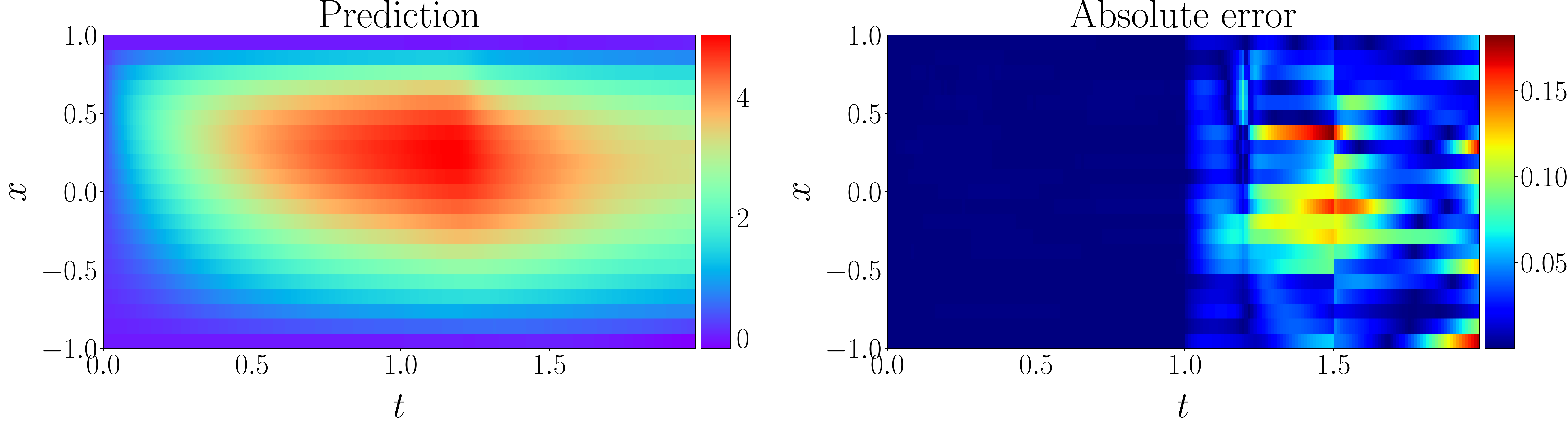}
      \label{fig:heat-k3}
      }
  \subfigure[Iteration step 6, $\hat{u}(t,x)$ and $|u(t,x)-\hat{u}(t,x)|$]{
      \includegraphics[scale=0.18]{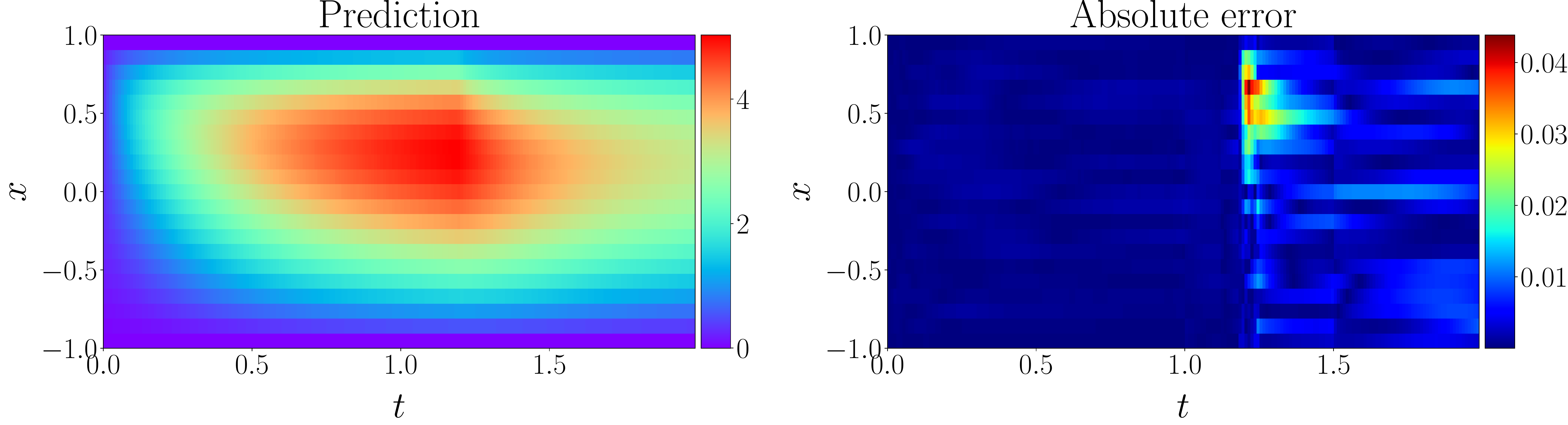}
      \label{fig:heat-k6}
      }
   \caption{
   DNN predictions and errors, heat equation.}
   \label{fig:heat-equation-plots}
\end{figure}

\begin{figure}[htp!]
   \centering
   \includegraphics[scale=0.2]{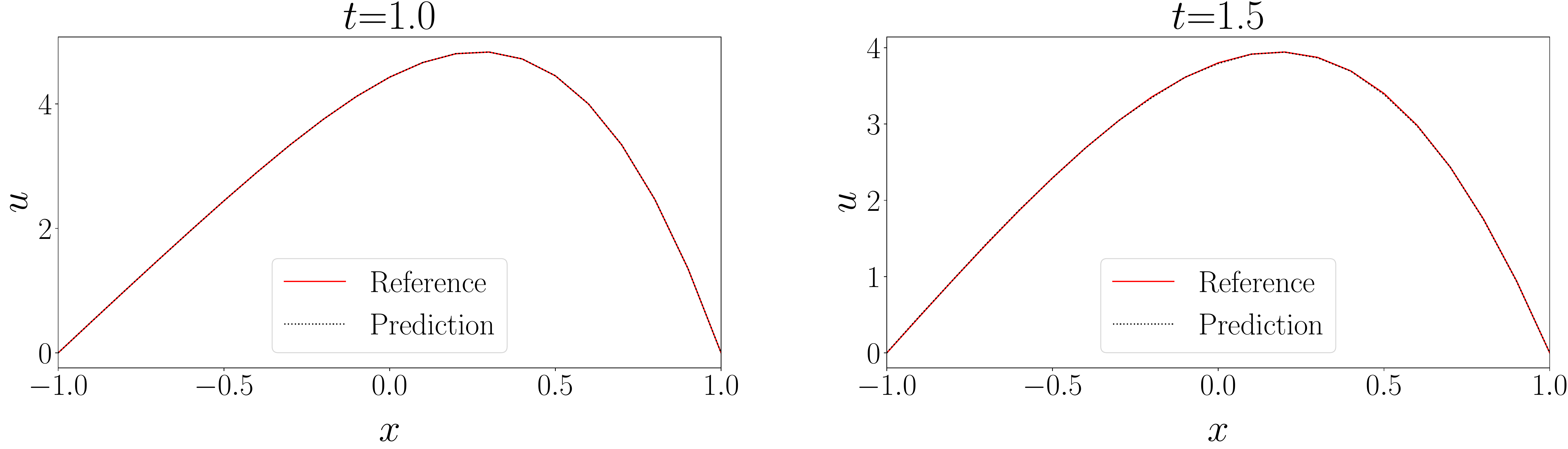}
   \caption{
   DNN predictions and reference solutions for $t=1$ and $t=1.5$, heat equation.}
   \label{fig:heat-eqn-sample}
\end{figure}

For comparison, a reference solution with $a=1$ is computed using SciPy, which is not included in the training datasets. \figurename{\ref{fig:heat-eqn-reference}} shows the reference solution. 
For each $x_i$ and $t_j$ ($i=1,\ldots,I_{\text{grid}}$ and $j=1,\ldots,J$), the prediction of the deep ResNets obtained by Algorithm \ref{alg:main-alg} is denoted by $\hat{u}(t_j,x_i)$, and the point-wise absolute error is defined as $|u(t_j,x_i)-\hat{u}(t_j,x_i)|$. 
\figurename{\ref{fig:heat-equation-plots}} shows the predictions of the trained DNNs and the corresponding errors. 
It can be seen that after three adaptivity iteration steps, the DNN predictions are very close to the reference solution, and the maximum error is only around $0.04$ for the DNNs obtained at the last iteration step (step six). 
Finally, the DNN predictions and the reference solutions at $t=1$ and $t=1.5$ are shown in \figurename{\ref{fig:heat-eqn-sample}}, where it can be seen that each DNN prediction and the corresponding reference solution are visually indistinguishable.


\section{Conclusions}
\label{sec:conclusions}
The divide and conquer principle is one of the fundamental concepts for learning dynamical systems with high complexity. With a focus on ResNet based learning methods, we have proposed a deep neural network based adaptive learning  (DNN-AL) approach. Through adaptively decomposing the datasets associated with large validation errors, deep neural networks (DNNs) are hierarchically constructed to efficiently learn local governing equations, and the unknown switching time instants are identified gradually. Especially, during the adaptive procedure, the network parameters at the previous iteration steps are reused as initial parameters for the current iteration step, which provides an efficient initialization strategy for training local DNNs. From our analysis, prediction error bounds are established for the DNNs obtained in DNN-AL. Several numerical examples demonstrate the effectiveness of DNN-AL. As this work is limited to switched systems under time-dependent switching, developing adaptive learning methods for systems with both time and state dependent switching will be the focus of our future work. 

\section*{Acknowledgments}
This work is supported by the National Natural Science Foundation of China (No. 12071291), the Science and Technology Commission of Shanghai Municipality (No. 20JC1414300) and the Natural Science Foundation of Shanghai (No. 20ZR1436200).



\bibliographystyle{elsarticle-num} 
\bibliography{references}

\begin{thebibliography}{10}
\expandafter\ifx\csname url\endcsname\relax
  \def\url#1{\texttt{#1}}\fi
\expandafter\ifx\csname urlprefix\endcsname\relax\def\urlprefix{URL }\fi
\expandafter\ifx\csname href\endcsname\relax
  \def\href#1#2{#2} \def\path#1{#1}\fi

\bibitem{juang1985eigensystem}
J.-N. Juang, R.~S. Pappa, \href{https://doi.org/10.2514/3.20031}{An eigensystem
  realization algorithm for modal parameter identification and model
  reduction}, Journal of Guidance, Control, and Dynamics 8~(5) (1985) 620--627.
\newblock \href {https://doi.org/10.2514/3.20031} {\path{doi:10.2514/3.20031}}.
\newline\urlprefix\url{https://doi.org/10.2514/3.20031}

\bibitem{juang1993identification}
J.-N. Juang, M.~Phan, L.~G. Horta, R.~W. Longman,
  \href{https://doi.org/10.2514/3.21006}{Identification of observer/{K}alman
  filter {M}arkov parameters - {T}heory and experiments}, Journal of Guidance,
  Control, and Dynamics 16~(2) (1993) 320--329.
\newblock \href {https://doi.org/10.2514/3.21006} {\path{doi:10.2514/3.21006}}.
\newline\urlprefix\url{https://doi.org/10.2514/3.21006}

\bibitem{brunton2016discovering}
S.~L. Brunton, J.~L. Proctor, J.~N. Kutz,
  \href{https://www.pnas.org/doi/abs/10.1073/pnas.1517384113}{Discovering
  governing equations from data by sparse identification of nonlinear dynamical
  systems}, Proceedings of the National Academy of Sciences 113~(15) (2016)
  3932--3937.
\newblock \href {https://doi.org/10.1073/pnas.1517384113}
  {\path{doi:10.1073/pnas.1517384113}}.
\newline\urlprefix\url{https://www.pnas.org/doi/abs/10.1073/pnas.1517384113}

\bibitem{schaeffer2017sparse}
H.~Schaeffer, S.~G. McCalla,
  \href{https://link.aps.org/doi/10.1103/PhysRevE.96.023302}{Sparse model
  selection via integral terms}, Phys. Rev. E 96 (2017) 023302.
\newblock \href {https://doi.org/10.1103/PhysRevE.96.023302}
  {\path{doi:10.1103/PhysRevE.96.023302}}.
\newline\urlprefix\url{https://link.aps.org/doi/10.1103/PhysRevE.96.023302}

\bibitem{zhang2018robust}
S.~Zhang, G.~Lin,
  \href{https://royalsocietypublishing.org/doi/abs/10.1098/rspa.2018.0305}{Robust
  data-driven discovery of governing physical laws with error bars},
  Proceedings of the Royal Society A: Mathematical, Physical and Engineering
  Sciences 474~(2217) (2018) 20180305.
\newblock \href {https://doi.org/10.1098/rspa.2018.0305}
  {\path{doi:10.1098/rspa.2018.0305}}.
\newline\urlprefix\url{https://royalsocietypublishing.org/doi/abs/10.1098/rspa.2018.0305}

\bibitem{tran2017exact}
G.~Tran, R.~Ward, \href{https://doi.org/10.1137/16M1086637}{Exact recovery of
  chaotic systems from highly corrupted data}, Multiscale Modeling \&
  Simulation 15~(3) (2017) 1108--1129.
\newblock \href {https://doi.org/10.1137/16M1086637}
  {\path{doi:10.1137/16M1086637}}.
\newline\urlprefix\url{https://doi.org/10.1137/16M1086637}

\bibitem{schaeffer2018extracting}
H.~Schaeffer, G.~Tran, R.~Ward,
  \href{https://doi.org/10.1137/18M116798X}{Extracting sparse high-dimensional
  dynamics from limited data}, SIAM Journal on Applied Mathematics 78~(6)
  (2018) 3279--3295.
\newblock \href {https://doi.org/10.1137/18M116798X}
  {\path{doi:10.1137/18M116798X}}.
\newline\urlprefix\url{https://doi.org/10.1137/18M116798X}

\bibitem{schmidt2009distilling}
M.~Schmidt, H.~Lipson,
  \href{https://www.science.org/doi/abs/10.1126/science.1165893}{Distilling
  free-form natural laws from experimental data}, Science 324~(5923) (2009)
  81--85.
\newblock \href {https://doi.org/10.1126/science.1165893}
  {\path{doi:10.1126/science.1165893}}.
\newline\urlprefix\url{https://www.science.org/doi/abs/10.1126/science.1165893}

\bibitem{kevrekidis2003equation}
C.~W. Gear, J.~M. Hyman, P.~G. Kevrekidid, I.~G. Kevrekidis, O.~Runborg,
  C.~Theodoropoulos,
  \href{https://doi.org/10.4310/CMS.2003.v1.n4.a5}{Equation-free,
  coarse-grained multiscale computation: Enabling mocroscopic simulators to
  perform system-level analysis}, Communications in Mathematical Sciences 1~(4)
  (2003) 715 -- 762.
\newblock \href {https://doi.org/10.4310/CMS.2003.v1.n4.a5}
  {\path{doi:10.4310/CMS.2003.v1.n4.a5}}.
\newline\urlprefix\url{https://doi.org/10.4310/CMS.2003.v1.n4.a5}

\bibitem{chen2018neural}
R.~T.~Q. Chen, Y.~Rubanova, J.~Bettencourt, D.~K. Duvenaud,
  \href{https://proceedings.neurips.cc/paper/2018/file/69386f6bb1dfed68692a24c8686939b9-Paper.pdf}{Neural
  ordinary differential equations}, in: S.~Bengio, H.~Wallach, H.~Larochelle,
  K.~Grauman, N.~Cesa-Bianchi, R.~Garnett (Eds.), Advances in Neural
  Information Processing Systems, Vol.~31, Curran Associates, Inc., 2018.
\newline\urlprefix\url{https://proceedings.neurips.cc/paper/2018/file/69386f6bb1dfed68692a24c8686939b9-Paper.pdf}

\bibitem{lu2018beyond}
Y.~Lu, A.~Zhong, Q.~Li, B.~Dong,
  \href{http://proceedings.mlr.press/v80/lu18d.html}{Beyond finite layer neural
  networks: Bridging deep architectures and numerical differential equations},
  in: J.~G. Dy, A.~Krause (Eds.), Proceedings of the 35th International
  Conference on Machine Learning, {ICML} 2018, Stockholmsm{\"{a}}ssan,
  Stockholm, Sweden, July 10-15, 2018, Vol.~80 of Proceedings of Machine
  Learning Research, {PMLR}, 2018, pp. 3282--3291.
\newline\urlprefix\url{http://proceedings.mlr.press/v80/lu18d.html}

\bibitem{weinan2017proposal}
W.~E, \href{https://doi.org/10.1007/s40304-017-0103-z}{A proposal on machine
  learning via dynamical systems}, Communications in Mathematics and Statistics
  5~(1) (2017) 1--11.
\newblock \href {https://doi.org/10.1007/s40304-017-0103-z}
  {\path{doi:10.1007/s40304-017-0103-z}}.
\newline\urlprefix\url{https://doi.org/10.1007/s40304-017-0103-z}

\bibitem{he2016deep}
K.~He, X.~Zhang, S.~Ren, J.~Sun, Deep residual learning for image recognition,
  in: 2016 IEEE Conference on Computer Vision and Pattern Recognition (CVPR),
  2016, pp. 770--778.
\newblock \href {https://doi.org/10.1109/CVPR.2016.90}
  {\path{doi:10.1109/CVPR.2016.90}}.

\bibitem{qin2019data}
T.~Qin, K.~Wu, D.~Xiu,
  \href{https://www.sciencedirect.com/science/article/pii/S0021999119304504}{Data
  driven governing equations approximation using deep neural networks}, Journal
  of Computational Physics 395 (2019) 620--635.
\newblock \href {https://doi.org/https://doi.org/10.1016/j.jcp.2019.06.042}
  {\path{doi:https://doi.org/10.1016/j.jcp.2019.06.042}}.
\newline\urlprefix\url{https://www.sciencedirect.com/science/article/pii/S0021999119304504}

\bibitem{qin2021data}
T.~Qin, Z.~Chen, J.~D. Jakeman, D.~Xiu,
  \href{https://doi.org/10.1137/20M1342859}{Data-driven learning of
  nonautonomous systems}, SIAM Journal on Scientific Computing 43~(3) (2021)
  A1607--A1624.
\newblock \href {https://doi.org/10.1137/20M1342859}
  {\path{doi:10.1137/20M1342859}}.
\newline\urlprefix\url{https://doi.org/10.1137/20M1342859}

\bibitem{fu2020learning}
X.~Fu, L.-B. Chang, D.~Xiu, \href{https://doi.org/10.1615/.2020034232}{Learning
  reduced systems via deep neural networks with memory}, Journal of Machine
  Learning for Modeling and Computing 1~(2) (2020) 97--118.
\newblock \href {https://doi.org/10.1615/.2020034232}
  {\path{doi:10.1615/.2020034232}}.
\newline\urlprefix\url{https://doi.org/10.1615/.2020034232}

\bibitem{shixiaojiang21}
J.~Harlim, S.~W. Jiang, S.~Liang, H.~Yang,
  \href{https://www.sciencedirect.com/science/article/pii/S0021999120306963}{Machine
  learning for prediction with missing dynamics}, Journal of Computational
  Physics 428 (2021) 109922.
\newblock \href {https://doi.org/https://doi.org/10.1016/j.jcp.2020.109922}
  {\path{doi:https://doi.org/10.1016/j.jcp.2020.109922}}.
\newline\urlprefix\url{https://www.sciencedirect.com/science/article/pii/S0021999120306963}

\bibitem{keeling2001seasonally}
M.~J. Keeling, P.~Rohani, B.~T. Grenfell,
  \href{https://www.sciencedirect.com/science/article/pii/S0167278900001871}{Seasonally
  forced disease dynamics explored as switching between attractors}, Physica D:
  Nonlinear Phenomena 148~(3) (2001) 317--335.
\newblock \href {https://doi.org/https://doi.org/10.1016/S0167-2789(00)00187-1}
  {\path{doi:https://doi.org/10.1016/S0167-2789(00)00187-1}}.
\newline\urlprefix\url{https://www.sciencedirect.com/science/article/pii/S0167278900001871}

\bibitem{holmes2006dynamics}
P.~Holmes, R.~J. Full, D.~Koditschek, J.~Guckenheimer,
  \href{https://doi.org/10.1137/S0036144504445133}{The dynamics of legged
  locomotion: Models, analyses, and challenges}, SIAM Review 48~(2) (2006)
  207--304.
\newblock \href {https://doi.org/10.1137/S0036144504445133}
  {\path{doi:10.1137/S0036144504445133}}.
\newline\urlprefix\url{https://doi.org/10.1137/S0036144504445133}

\bibitem{dobson2007complex}
I.~Dobson, B.~A. Carreras, V.~E. Lynch, D.~E. Newman, Complex systems analysis
  of series of blackouts: Cascading failure, critical points, and
  self-organization, Chaos: An Interdisciplinary Journal of Nonlinear Science
  17~(2) (2007) 026103.

\bibitem{sanfelice2016analysis}
R.~G. Sanfelice, Analysis and Design of Cyber-Physical Systems: A Hybrid
  Control Systems Approach, CRC Press, 2015, pp. 3--31.
\newblock \href {https://doi.org/10.1201/b19290-3}
  {\path{doi:10.1201/b19290-3}}.

\bibitem{van2000introduction}
A.~Schaft, S.~Hans, An Introduction to Hybrid Dynamical Systems, Springer,
  London, 2000.
\newblock \href {https://doi.org/10.1007/BFb0109998}
  {\path{doi:10.1007/BFb0109998}}.

\bibitem{cortes2008discontinuous}
J.~Cortes, Discontinuous dynamical systems, IEEE Control Systems Magazine
  28~(3) (2008) 36--73.
\newblock \href {https://doi.org/10.1109/MCS.2008.919306}
  {\path{doi:10.1109/MCS.2008.919306}}.

\bibitem{liberzon2003switching}
D.~Liberzon, Switching in Systems and Control, Birkhäuser, Boston, 2003.
\newblock \href {https://doi.org/10.1007/978-1-4612-0017-8}
  {\path{doi:10.1007/978-1-4612-0017-8}}.

\bibitem{mangan2019model}
N.~M. Mangan, T.~Askham, S.~L. Brunton, J.~N. Kutz, J.~L. Proctor,
  \href{https://royalsocietypublishing.org/doi/abs/10.1098/rspa.2018.0534}{Model
  selection for hybrid dynamical systems via sparse regression}, Proceedings of
  the Royal Society A: Mathematical, Physical and Engineering Sciences
  475~(2223) (2019) 20180534.
\newblock \href {https://doi.org/10.1098/rspa.2018.0534}
  {\path{doi:10.1098/rspa.2018.0534}}.
\newline\urlprefix\url{https://royalsocietypublishing.org/doi/abs/10.1098/rspa.2018.0534}

\bibitem{goodfellow2016deep}
I.~Goodfellow, Y.~Bengio, A.~Courville, Deep Learning, MIT Press, Cambridge,
  2016, \url{http://www.deeplearningbook.org}.

\bibitem{bottou2018optimization}
L.~Bottou, F.~E. Curtis, J.~Nocedal,
  \href{https://doi.org/10.1137/16M1080173}{Optimization methods for
  large-scale machine learning}, SIAM Review 60~(2) (2018) 223--311.
\newblock \href {https://doi.org/10.1137/16M1080173}
  {\path{doi:10.1137/16M1080173}}.
\newline\urlprefix\url{https://doi.org/10.1137/16M1080173}

\bibitem{lecun1989backpropagation}
Y.~LeCun, B.~Boser, J.~S. Denker, D.~Henderson, R.~E. Howard, W.~Hubbard, L.~D.
  Jackel, Backpropagation applied to handwritten zip code recognition, Neural
  Computation 1~(4) (1989) 541--551.
\newblock \href {https://doi.org/10.1162/neco.1989.1.4.541}
  {\path{doi:10.1162/neco.1989.1.4.541}}.

\bibitem{krizhevsky2012imagenet}
A.~Krizhevsky, I.~Sutskever, G.~E. Hinton,
  \href{https://proceedings.neurips.cc/paper/2012/file/c399862d3b9d6b76c8436e924a68c45b-Paper.pdf}{Imagenet
  classification with deep convolutional neural networks}, in: F.~Pereira,
  C.~Burges, L.~Bottou, K.~Weinberger (Eds.), Advances in Neural Information
  Processing Systems, Vol.~25, Curran Associates, Inc., 2012.
\newline\urlprefix\url{https://proceedings.neurips.cc/paper/2012/file/c399862d3b9d6b76c8436e924a68c45b-Paper.pdf}

\bibitem{teschl2012ordinary}
G.~Teschl, Ordinary Differential Equations and Dynamical Systems, American
  Mathematical Society, Providence, 2012.

\bibitem{pinkus1999approximation}
A.~Pinkus, Approximation theory of the {MLP} model in neural networks, Acta
  Numerica 8 (1999) 143–195.
\newblock \href {https://doi.org/10.1017/S0962492900002919}
  {\path{doi:10.1017/S0962492900002919}}.

\bibitem{paszke2019pytorch}
A.~Paszke, S.~Gross, F.~Massa, A.~Lerer, J.~Bradbury, G.~Chanan, T.~Killeen,
  Z.~Lin, N.~Gimelshein, L.~Antiga, A.~Desmaison, A.~Kopf, E.~Yang, Z.~DeVito,
  M.~Raison, A.~Tejani, S.~Chilamkurthy, B.~Steiner, L.~Fang, J.~Bai,
  S.~Chintala,
  \href{http://papers.neurips.cc/paper/9015-pytorch-an-imperative-style-high-performance-deep-learning-library.pdf}{Pytorch:
  An imperative style, high-performance deep learning library}, in: H.~Wallach,
  H.~Larochelle, A.~Beygelzimer, F.~d\textquotesingle Alch\'{e}-Buc, E.~Fox,
  R.~Garnett (Eds.), Advances in Neural Information Processing Systems 32,
  Curran Associates, Inc., 2019, pp. 8024--8035.
\newline\urlprefix\url{http://papers.neurips.cc/paper/9015-pytorch-an-imperative-style-high-performance-deep-learning-library.pdf}

\bibitem{loshchilov2017decoupled}
I.~Loshchilov, F.~Hutter,
  \href{https://openreview.net/forum?id=Bkg6RiCqY7}{Decoupled weight decay
  regularization}, in: 7th International Conference on Learning
  Representations, {ICLR} 2019, New Orleans, LA, USA, May 6-9, 2019,
  OpenReview.net, 2019.
\newline\urlprefix\url{https://openreview.net/forum?id=Bkg6RiCqY7}

\bibitem{loshchilov2016sgdr}
I.~Loshchilov, F.~Hutter,
  \href{https://openreview.net/forum?id=Skq89Scxx}{{SGDR:} stochastic gradient
  descent with warm restarts}, in: 5th International Conference on Learning
  Representations, {ICLR} 2017, Toulon, France, April 24-26, 2017, Conference
  Track Proceedings, OpenReview.net, 2017.
\newline\urlprefix\url{https://openreview.net/forum?id=Skq89Scxx}

\bibitem{he2015delving}
K.~He, X.~Zhang, S.~Ren, J.~Sun, Delving deep into rectifiers: Surpassing
  human-level performance on imagenet classification, in: 2015 IEEE
  International Conference on Computer Vision (ICCV), 2015, pp. 1026--1034.
\newblock \href {https://doi.org/10.1109/ICCV.2015.123}
  {\path{doi:10.1109/ICCV.2015.123}}.

\bibitem{petzold1983automatic}
L.~Petzold, \href{https://doi.org/10.1137/0904010}{Automatic selection of
  methods for solving stiff and nonstiff systems of ordinary differential
  equations}, SIAM Journal on Scientific and Statistical Computing 4~(1) (1983)
  136--148.
\newblock \href {https://doi.org/10.1137/0904010} {\path{doi:10.1137/0904010}}.
\newline\urlprefix\url{https://doi.org/10.1137/0904010}

\bibitem{2020SciPy-NMeth}
P.~Virtanen, R.~Gommers, T.~E. Oliphant, M.~Haberland, T.~Reddy, D.~Cournapeau,
  E.~Burovski, P.~Peterson, W.~Weckesser, J.~Bright, S.~J. {van der Walt},
  M.~Brett, J.~Wilson, K.~J. Millman, N.~Mayorov, A.~R.~J. Nelson, E.~Jones,
  R.~Kern, E.~Larson, C.~J. Carey, {\.I}.~Polat, Y.~Feng, E.~W. Moore,
  J.~{VanderPlas}, D.~Laxalde, J.~Perktold, R.~Cimrman, I.~Henriksen, E.~A.
  Quintero, C.~R. Harris, A.~M. Archibald, A.~H. Ribeiro, F.~Pedregosa, P.~{van
  Mulbregt}, {SciPy 1.0 Contributors}, {SciPy} 1.0: Fundamental algorithms for
  scientific computing in python, Nature Methods 17 (2020) 261--272.
\newblock \href {https://doi.org/10.1038/s41592-019-0686-2}
  {\path{doi:10.1038/s41592-019-0686-2}}.

\end{thebibliography}
\end{document}